\documentclass{article}

\usepackage[preprint]{corl_2022} 

\pdfoutput=1 

\input{macro}

\begin{document}

\title{Rethinking Optimization with Differentiable Simulation from a Global Perspective}

\author{
  Rika Antonova\thanks{Rika and Jingyun contributed equally. Contacts: {\small{\texttt{\{rika.antonova, jingyuny\}@stanford.edu}}}. \quad
  This project was supported in part by a research award \mbox{from Meta.} Rika was supported by the National Science Foundation grant No.2030859 to the Computing Research Association for the CIFellows Project.}\ \ $^1$, \ \ Jingyun Yang$^*$$^1$,  \ \ Krishna Murthy Jatavallabhula$^2$, \ \ Jeannette Bohg$^1$ \\
  \\
  $^1$Stanford University, \quad $^2$Massachusetts Institute of Technology
}

\pdfinfo{
   /Author (Rika Antonova, Jingyun Yang, Krishna Murthy Jatavallabhula, Jeannette Bohg)
   /Title  (Rethinking Optimization with Differentiable Simulation from a Global Perspective)
   /CreationDate (D:2022060120000)
   /Subject (Robotics)
   /Keywords (Differentiable simulation, Global optimization, Deformable objects)
}
\definecolor{mydarkblue}{rgb}{0,0.08,0.45}
\hypersetup{ %
    pdftitle={Rethinking Optimization with Differentiable Simulation from a Global Perspective},
    pdfauthor={Rika Antonova, Jingyun Yang, Krishna Murthy Jatavallabhula, Jeannette Bohg},
    pdfsubject={Robotics},
    pdfkeywords={Differentiable simulation, Global optimization, Deformable objects},
    pdfborder=0 0 0,
    pdfpagemode=UseNone,
    colorlinks=true,
    linkcolor=mydarkblue,
    citecolor=mydarkblue,
    filecolor=mydarkblue,
    urlcolor=mydarkblue,
    pdfview=FitH
}

\maketitle

\vspace{-15px}
\begin{abstract}
    Differentiable simulation is a promising toolkit for fast gradient-based policy optimization and system identification. However, existing approaches to differentiable simulation have largely tackled scenarios where obtaining smooth gradients has been relatively easy, such as systems with mostly smooth dynamics. In this work, we study the challenges that differentiable simulation presents when it is not feasible to expect that a single descent reaches a global optimum, which is often a problem in contact-rich scenarios. We analyze the optimization landscapes of diverse scenarios that contain both rigid bodies and deformable objects. In dynamic environments with highly deformable objects and fluids, differentiable simulators produce rugged landscapes with nonetheless useful gradients in some parts of the space. We propose a method that combines Bayesian optimization with semi-local `leaps' to obtain a global search method that can use gradients effectively, while also maintaining robust performance in regions with noisy gradients. We show that our approach outperforms several gradient-based and gradient-free baselines on an extensive set of experiments in simulation, and also validate the method using experiments with a real robot and deformables. Videos and supplementary materials are available at \href{https://tinyurl.com/globdiff}{{https://tinyurl.com/globdiff}}.
\end{abstract}

\vspace{10px}

\section{Introduction}

Physics simulation is indispensable for robot learning: it has been widely used to generate synthetic training data, explore learning of complex sensorimotor policies, and also help anticipate the performance of various learning methods before deploying on real robots. An increasing volume of recent work attempts to \emph{invert} physics engines: given simulation outputs (e.g. trajectories), infer the input parameters (e.g. physical properties of the scene, robot controls) that best explain the outputs \cite{krishna2021gradsim, huang2021plasticinelab, lin2022diffskill, sundaresan2022diffcloud, ma2022risp}. Differentiable physics engines aim to offer a direct and efficient way to invert simulation, and could enable fast gradient-based policy optimization and system identification. Works that we survey in the background section show a number of recent successes. However, to highlight the most promising prospects of differentiable simulation, works in this space focus primarily on scenarios where obtaining smooth gradients is relatively easy.

With more research efforts gearing to develop (and use) differentiable physics engines, it is now crucial that we analyze the limits of these systems thoroughly. Surprisingly, efforts to investigate the \emph{differentiability} of these simulators are far and few.
One prior work~\cite{sun2022better} has highlighted a few fundamental limitations of differentiable simulation in the presence of rigid contacts, in low-dimensional systems.
In this work, we first investigate the \emph{quality} of gradients by visualizing loss landscapes through differentiable simulators for several robotic manipulation tasks of interest.
For this analysis we create several new challenging environments, and also use environments from prior work.

Our main focus is to understand whether the limitations of rigid contacts are also prevalent across deformable object simulation. We show that in many cases obtaining well-behaved gradients is feasible, but challenge the assumption that differentiable simulators provide easy landscapes in sufficiently interesting scenarios. We analyze scenarios with flexible deformable objects (cloth), plastics (clay), and fluids.
Our visualizations of optimization landscapes uncover numerous local optima and plateaus, showing the need for extending gradient-based methods with global search. We propose a method that combines global search using Bayesian optimization with a semi-local search.
Our semi-local strategy allows to make progress on parts of the landscape that are intractable for gradient descent alone. In our experiments we show visual, quantitative and qualitative analysis of the optimization problems arising from simulations with deformables in a variety of scenarios. We also validate our proposed approach on a real robot  interacting with cloth, where our aim is to identify the properties of the cloth that make the motion of the simulated cloth match the real one.

\vspace{-5px}
\section{Background}
\vspace{-5px}

\textbf{Differentiable Simulation:} The earliest differentiable simulators~\cite{degrave2019diffphysics,diff_basic_neurips2018} focused solely on rigid-body dynamics, often operating only over a small number of predefined object primitives.
A number of approaches subsequently focused on arbitrary rigid body shapes~\cite{song20a-diffsim,song2020b-slide}, articulations~\cite{qiao2021efficient,lutter2021differentiable,mora2021pods}, accurate contacts~\cite{werling2021fast,geilinger2020add,howelllecleach2022}, scalability~\cite{qiao2020scalable}, speed~\cite{freeman2021brax,hu2019difftaichi}, and multiphysics~\cite{krishna2021gradsim}.
Differentiable simulation has also been explored in the context of deformable objects~\cite{hu2019chainqueen}, cloth~\cite{liang2019differentiable}, fluids~\cite{difffluid}, robotic cutting~\cite{heiden2021disect} and other scientific and engineering phenomena~\cite{wang2020differentiable,ma2021diffaqua,jaxmd2020}.
While several approaches sought to achieve physically accurate forward simulations and their gradients, they did not rigorously study the impact of their loss landscapes on inference.
Most modern simulators are plagued by two fundamental issues. First, they rely on gradient descent as an inference mechanism, which makes the optimization dependent on a \emph{good} initial guess.
Second, the loss landscape for contact-rich scenarios is laden with discontinuities and spurious local optima.
This is evident even in the most basic case of a single bouncing ball as shown in~\cite{hu2019difftaichi}. In this work, we also verify that the issue is present in the more recent frameworks, such as Nimble~\cite{werling2021fast} and Warp~\cite{warp2022}.

\textbf{Global Search Methods:} Global optimization tackles the problem of finding the global optimum in a given search space. In our case, this could constitute finding the optimal parameters for a controller (e.g. target position, velocity, force, torque) or physical parameters of a simulator to make the behavior of simulated objects match reality. Global optimization includes several broad families of methods: 1) space covering methods that systematically visit all parts of the space; 2) clustering methods that use cluster analysis to decide which areas of the space could be promising; 3) evolutionary methods that start from a broad set of candidates and evolve them towards exploring the promising regions; 4) Bayesian optimization (BO) that uses non-parametric approaches to keep track of a global model of the cost function and its uncertainty on the whole search space. The survey in~\cite{locatelli2021global} gives further details. Random search is also considered a global search method, and is one of the few methods that guarantees eventually finding a global optimum. While data efficiency can be a challenge, it is often a surprisingly robust baseline. Space covering and clustering methods face significant challenges when trying to scale to high dimensions. The recent focus in global optimization has been on evolutionary methods and BO, which can be successful on search spaces with thousands of dimensions. Hence, in this work we compare methods based on random search, evolutionary approaches and BO.

\textbf{CMA-ES:} One of the most versatile evolutionary methods is Covariance Matrix Adaptation - Evolution Strategies (CMA-ES)~\cite{hansen2001completely}. It samples a randomized `generation' of points at each iteration from a multivariate Gaussian distribution. To evolve this distribution, it computes the new mean from a subset of points with the lowest cost from the previous generation. It also uses these best-performing points to update the covariance matrix. The next generation of points is then sampled from the distribution with the updated mean and covariance. CMA-ES succeeds in a wide variety of applications~\cite{hansen2006cma}, and has competitive performance on global optimization benchmarks~\cite{hansen2010comparing}. This method does not make any restrictive assumptions about the search space and can be used `as-is' i.e. without tuning. However, this method is technically not fully global --- while it can overcome shallow local optima, it can get stuck in deeper local optima.

\textbf{Bayesian optimization (BO):} BO views the problem of global search as seeking a point $\pmb{x}^*$ to minimize a given cost function $f(\pmb{x})$:
$\displaystyle f(\pmb{x}^*) = \textstyle\min_{\pmb{x}} f(\pmb{x})$.
At each trial, BO optimizes an auxiliary acquisition function to select the next promising $\pmb{x}$ to evaluate. $f$ is frequently modeled with a Gaussian process (GP): $f(\pmb{x}) \sim \mathcal{GP}(m(\pmb{x}), k(\pmb{x}_i, \pmb{x}_j))$.
Modeling $f$ with a GP allows to compute the posterior mean $\bar{f}(\pmb{x})$ and uncertainty (variance) $Var[f(\pmb{x})]$ for each candidate point $\pmb{x}$. Hence, the acquisition function can select points to balance a high mean (exploitation) with high uncertainty (exploration).
The kernel function $k(\cdot, \cdot)$ encodes similarity between inputs. If $k(\pmb{x}_i, \pmb{x}_j)$ is large for inputs $\pmb{x}_i, \pmb{x}_j$, then $f(\pmb{x}_i$) strongly influences $f(\pmb{x}_j)$. One of the most widely used kernel functions is the Squared Exponential (SE):
$k_{SE}(\br \!\equiv\! |\pmb{x}_i \!-\! \pmb{x}_j|) = \sigma_k^2 \exp\big(\!-\!\tfrac{1}{2} \br^T \diag(\pmb{\ell})^{\!-\!2} \br \big)$, where $\sigma_k^2, \ \pmb{\ell}$ are signal variance and a vector of length scales respectively. $\sigma_k^2, \ \pmb{\ell}$ are hyperparameters and are optimized automatically by maximizing the marginal data likelihood. See~\cite{BOtutorial2016} for further details.

\section{BO-Leap: A Method for Global Search on Rugged Landscapes}

We propose an approach for global search that can benefit from gradient-based descents and employs a semi-local strategy to make progress on rough optimization landscapes.
To explore the loss landscape globally, we use Bayesian optimization (BO). BO models uncertainty over the loss and reduces uncertainty globally by exploring unseen regions. It also ensures to return to low-loss regions to further improve within the promising areas of the search space. BO uses an acquisition function to compute the most promising candidate to evaluate next. We treat each candidate as a starting point for a semi-local search. As we show in our experiments, the straightforward strategy of using gradient descent from each of these starting points does not ensure strong performance in scenarios with noisy gradients.
Hence, we propose a hybrid descent strategy that combines gradient-free search with gradient-based descents. For this, we collect a small population of local samples and compute a sampling distribution based on CMA-ES. Instead of directly using the resulting mean and covariance to sample the next population (as CMA-ES would), we use gradient descent to evolve the distribution mean, then use this updated mean when sampling the next population.

\begin{wrapfigure}{r}{0.5\textwidth}
\centering
\vspace{-10px}
\includegraphics[width=0.5\textwidth, height=0.34949\textwidth]{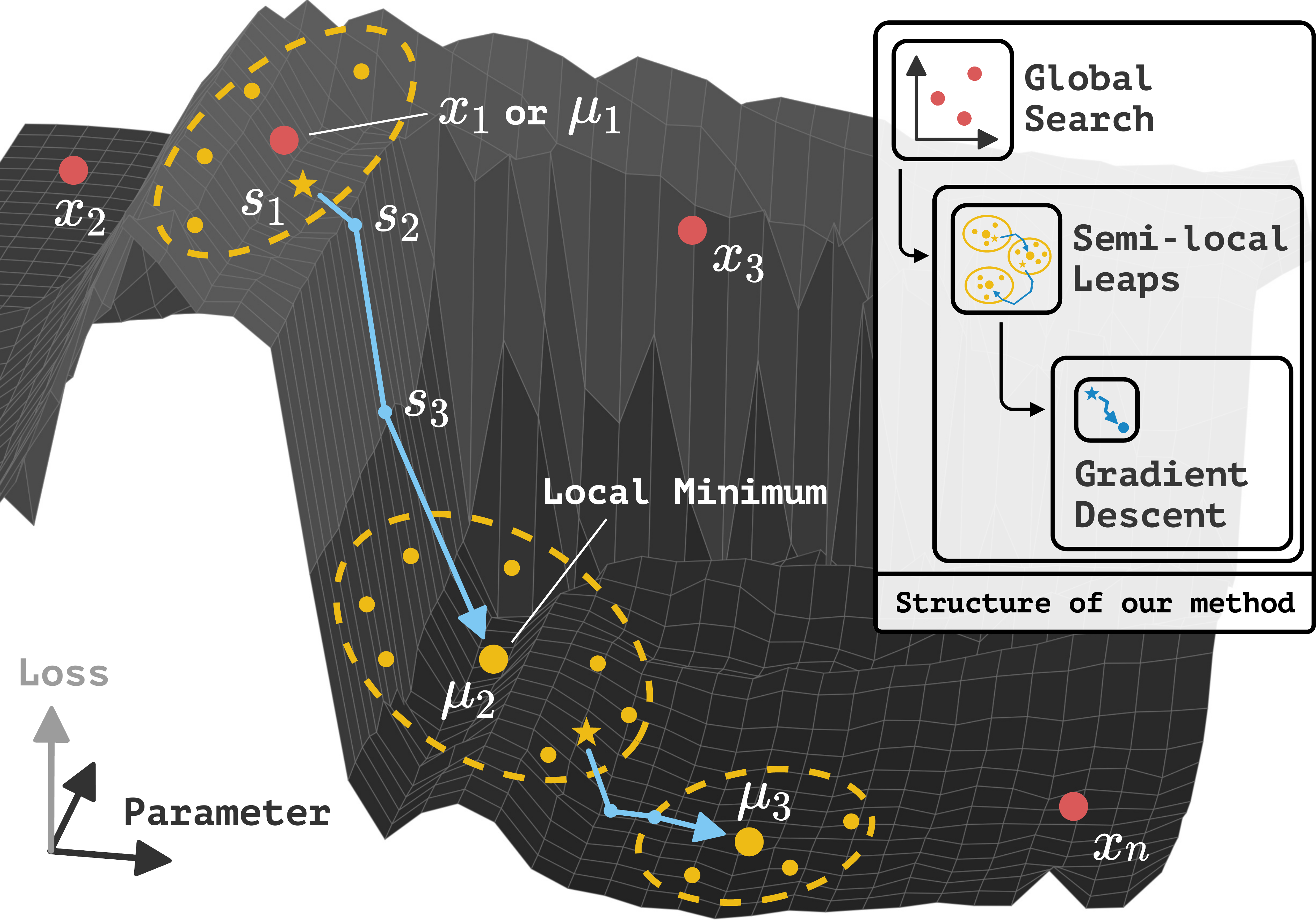}
\vspace{-10px}
\caption{A conceptual illustration of BO-Leap.}
\label{fig:leap}
\vspace{-10px}
\end{wrapfigure}
We outline our method, BO with semi-local leaps (\mbox{BO-Leap}), in Algorithm~\ref{alg:bol}. We also visualize the algorithm in Figure~\ref{fig:leap}. We start by initializing a global model for the loss with a Gaussian process (GP) prior. Using the BO acquisition function, we sample a vector $\pmb{x}_1$ of simulation or control parameters to evaluate. We then run a semi-local search from this starting point. For this, we initialize the population distribution $\mathcal{N}\big(\pmb{\mu}_{1}\!=\!\pmb{x}_1, \sigma^2_1\pmb{C}_{1}\big)$ and sample $K$ local candidates. Next, we update the distribution in a gradient-free way similar to the \mbox{CMA-ES} strategy. We then start a gradient descent from the updated mean $\pmb{s}_1$. The gradient descent runs for at most $J$ steps and is halted if the loss stagnates or increases for more than three steps. The gradient is clipped to avoid leaving the search space boundaries. When the semi-local search reaches a given number of steps (e.g. 100 in our experiments), we update the BO posterior and let BO pick a new starting point $\pmb{x}_2$ globally. We add all the points that the semi-local search encounters to the set $\pmb{S}_n$ that is used to compute the posterior for each BO trial.

Obtaining a noticeable improvement by incorporating gradients into a strategy based on CMA-ES is not trivial. One hybrid approach that seems conceptually sound proposes to shift the mean of each \mbox{CMA-ES} population by taking a step in the direction of the gradient~\cite{chen2009combining}. In our preliminary experiments, taking a single gradient step was insufficient to significantly improve the performance of \mbox{CMA-ES}: the method from~\cite{chen2009combining} performed worse than gradient-free \mbox{CMA-ES}. In contrast, our semi-local search strategy allows gradient-based descents to take large leaps on the parts of the landscape where \ gradients \ are relatively smooth. \ To prevent \ being \ mislead \ by \ unstable \ gradients

and avoid wasting computation when stuck on a plateau, our method monitors the quality of the gradient-based evolution of the mean and terminates unpromising descents early.
BO-Leap operates on three levels: global, semi-local and local (gradient descent), which allows it to tackle loss landscapes that are challenging due to various aspects: local optima, non-smooth losses, and noisy gradients. In the next section, we show that BO-Leap has strong empirical performance in contact-rich scenarios with highly deformable objects, plastic materials and liquid. 

\setlength{\textfloatsep}{6pt}
\IncMargin{1.2em}
\begin{algorithm}[t]
\DontPrintSemicolon
\caption{BO-Leap : Bayesian Optimization with Semi-local Leaps}
\label{alg:bol}
Initialize : $\pmb{S}_0 \lar \{\}$ \Comment*[r]{Set of points seen so far (initially empty)}
\ \ \quad \quad \quad \quad $\mathcal{L} \lar \mathcal{GP}\big(m(\pmb{x})\!=\!\pmb{0}, \!\ k(\pmb{x},\pmb{x'}) \ | \!\ \pmb{S}_0\big)$ \Comment*[r]{Gaussian process as a global loss model}
\ \ \quad \quad \quad \quad BO acquisition function $\lar$ LCB \Comment*[r]{Lower Confidence Bound}
\For{$n =1..max\_steps$}{
$\mathcal{GP}_{\pmb{S}_n}\!\!=\!\mathcal{GP}\big(m(\cdot),k(\cdot, \cdot)|\pmb{S}_n \big)$ \Comment*[r]{Compute GP posterior using Eq. 2.25-26 from~\cite{gpmlbook}}
$\pmb{x}_{n} \lar \argmin_{\pmb{x}} LCB\big(\pmb{x} | \ \mathcal{GP}_{\pmb{S}_n}\big)$ \Comment*[r]{Get next simulation (or control) parameter vector}
$\pmb{\mu}_{1} \lar \pmb{x}_n; \  \sigma_1 \lar 1.0 ; \ \pmb{C}_1 \lar \pmb{I}$ (identity)$; \ K\text{=}10 $ \Comment*[r]{Initialize local population distribution}
\For{$i = 1..local\_steps$} {
$\pmb{x}_{k} \sim \mathcal{N}\big(\pmb{\mu}_{i}, \sigma^2_i\pmb{C}_{i}\big), k=1..K$ \Comment*[r]{Sample $K$ population candidates}
$l_k \lar Sim(\pmb{x}_k), k=1..K$ \Comment*[r]{Compute losses $l_k$ with $\pmb{x}_k$ as sim/control parameters}
$K_{best} \lar sort(l_k)$ \Comment*[r]{Get candidates with the lowest loss}
$\pmb{s}_1 = \frac{1}{|K_{best}|} \!\!\sum_{k \in K_{best}}  \!\! \pmb{x}_k$ \ \Comment*[r]{Compute descent start point (CMA-ES population mean)}
\For{$j = 1..J$} {
$l_k, \nabla_{\!Sim}|_{\pmb{s}_j} \lar Sim(\pmb{s}_j)$ \Comment*[r]{Compute sim. loss $l_j$ and gradients $\nabla_{\!Sim}|_{\pmb{s}_j}$}
$\pmb{s}_j \lar \pmb{s}_{j-1} - \alpha \nabla_{\!Sim}|_{\pmb{s}_j}$ \Comment*[r]{Take a gradient step}
Break if $l_j$ stagnates for more than 3 steps
}  
$\pmb{\mu}_{i+1} \lar \pmb{s}_J ; \ \sigma_{i+1}, \pmb{C}_{i+1} \lar$ Eq.14-17 from~\cite{hansen2001completely} \Comment*[r]{Update local population distribution} 
$\pmb{S}_{n+1} = \pmb{S}_n \cup \{(\pmb{s}_j, l_j)\}_{j=1}^J \cup \{(\pmb{x}_k, l_k)\}_{k=1}^K$ \Comment*[r]{Update data for GP posterior}
}  
}  
\end{algorithm}
\DecMargin{1.2em}
\section{A Suite of Differentiable Simulation Scenarios}

\setlength{\tabcolsep}{3pt}
\begin{table}[H]
\centering
\small
\begin{tabular}{cccc}
\toprule
\textbf{Framework} & \textbf{Simulation Type} & \textbf{Supported Models} & \textbf{Environments} \\[1pt] \toprule
Nimble~\cite{werling2021fast} & Mesh-based               & Rigid                      & \makecell{Cartpole, 3-Link Cartpole*,\\ Pinball*}    \\\midrule
Warp~\cite{warp2022}  & Multiple                 & Rigid, deformable, fluid   & Bounce*                        \\\midrule
DiffTaichi ~\cite{hu2019difftaichi}   & Multiple                 & Rigid, deformable, fluid   & Fluid*, Swing*, Flip*    \\ \midrule
PlasticineLab~\cite{huang2021plasticinelab}  & Particle-based           & Rigid, deformable          & \makecell{Assembly, Pinch, RollingPin, Rope,\\ Table, Torus, TripleMove, Writer}   \\
\bottomrule
\end{tabular}
\vspace{5pt}
\caption{Summary of differentiable simulation frameworks and environments we use in this work. We used several existing environments and created new ones (marked with asterisks *). See supplement for more details. }
\vspace{-15pt}
\label{tab:summary_table}
\end{table}

In this work, we implement scenarios using several differentiable simulation frameworks. Table~\ref{tab:summary_table} summarizes their properties. Our main goal is to study scenarios with deformables, because prior works already explored a number of tasks limited to rigid-body motion (see previous section). That said, very few works considered contact-rich tasks, and one prior work highlighted the potential fundamental limitations of gradients for rigid contacts~\cite{sun2022better}. Hence, there is a need to further study the fundamentals of how loss landscapes and gradient fields are affected by rigid contacts. For that, we created several scenarios using \mbox{Nimble}~\cite{werling2021fast} and Warp~\cite{warp2022} frameworks (described below).
\vspace{3px}

\begin{wrapfigure}[4]{l}{45pt}
    \vspace{-13pt}
    \includegraphics[width=45pt]{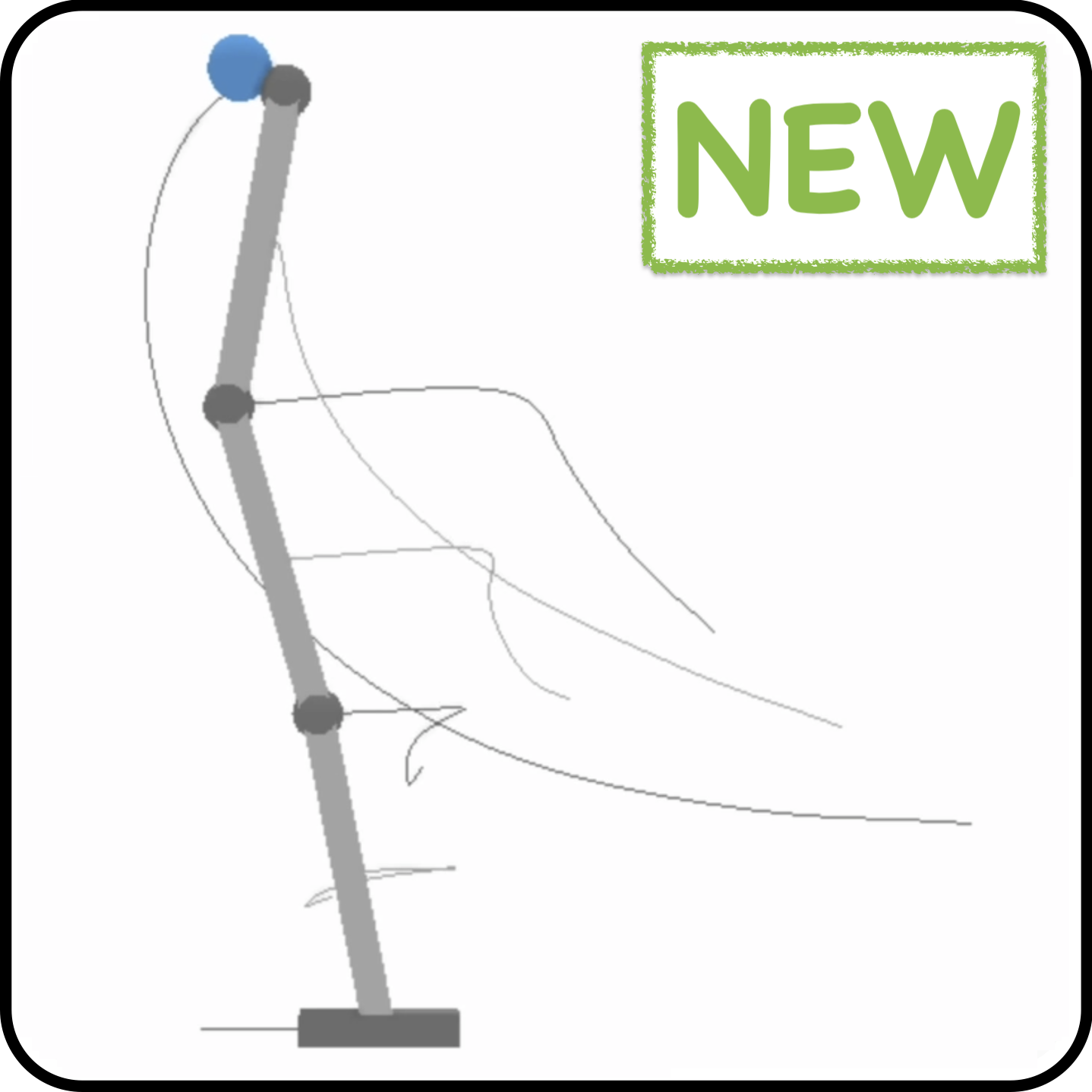}
\end{wrapfigure}
\textit{3-Link Cartpole}: an extension of the classic Cartpole to get more challenging dynamics. Here, a 3-link pole needs to reach the blue target with its tip. Cart velocity and joint torques are optimized for each of the 100 steps of the episode, yielding a 400-dimensional optimization problem.
\vspace{10pt}

\begin{wrapfigure}[4]{l}{45pt}
    \vspace{-1pt}
    \includegraphics[width=45pt]{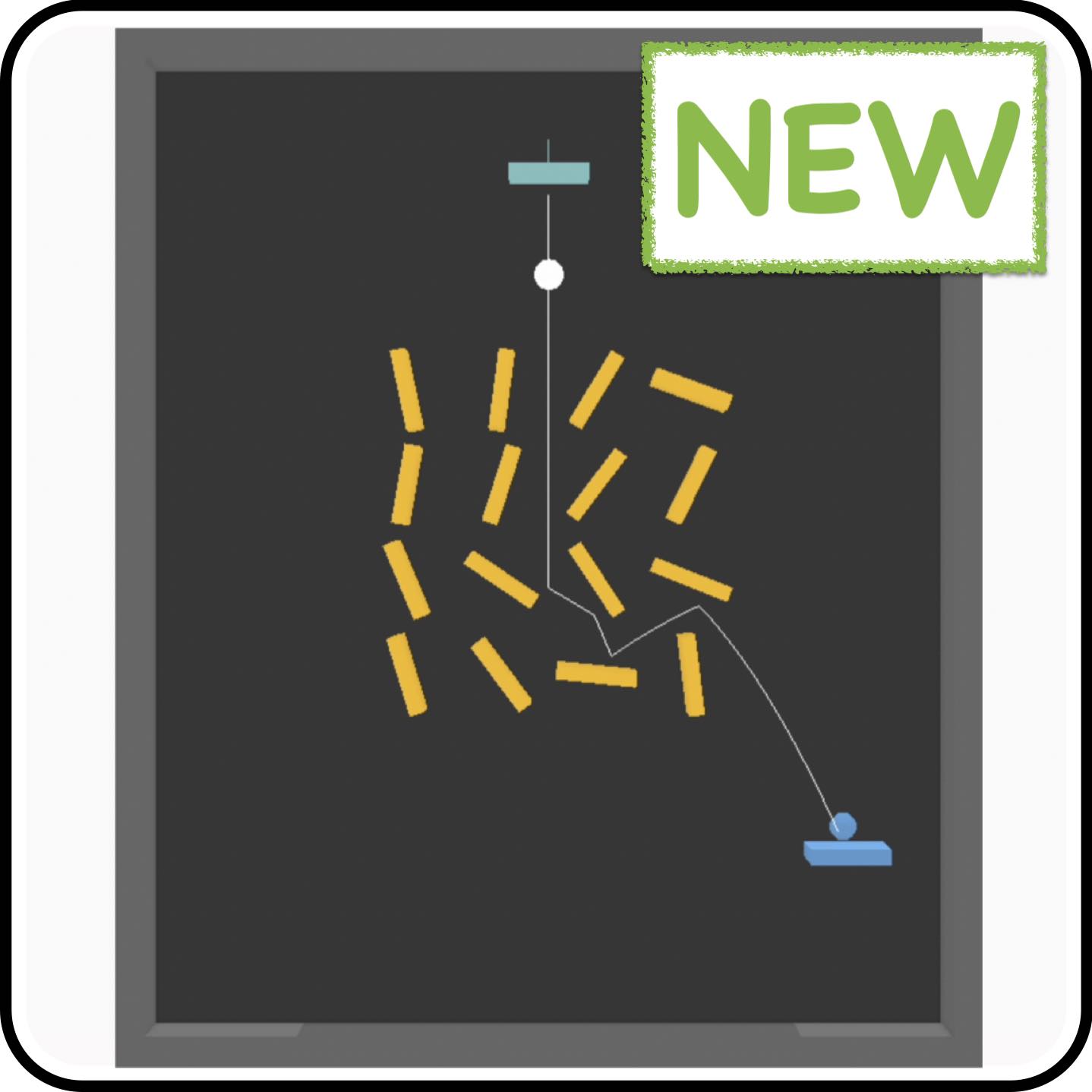}
\end{wrapfigure}
\textit{Pinball and Bounce}: a ball launches and bounces off colliders as in a pinball game. We optimize orientations of $n$ colliders to route the ball from the top to the blue target at the bottom. \textit{Pinball} helps analyze effects of increasing the number of contacts, implemented in Nimble~\cite{werling2021fast}. We also created a simplified \textit{Bounce} scenario to study effects of a single collision in Warp~\cite{warp2022}, which produced well-behaved gradients in prior work~\cite{heiden2021disect}. 
\vspace{-5px}

To go beyond rigid objects, mesh-based simulations can model highly deformable objects, such as cloth. Particle-based simulators can model interactions with granular matter and liquids, plastic deformation with objects permanently elongating, buckling, bending, or twisting under stress. We created several mesh-based and particle-based environments that involve deformables using DiffTaichi~\cite{hu2019difftaichi} and compared the quality of the loss landscapes and gradient fields they yield. We also included environments from the PlasticineLab~\cite{huang2021plasticinelab} that focus on plastic deformation. 

\begin{wrapfigure}[4]{l}{45pt}
    \vspace{-13pt}
    \includegraphics[width=45pt]{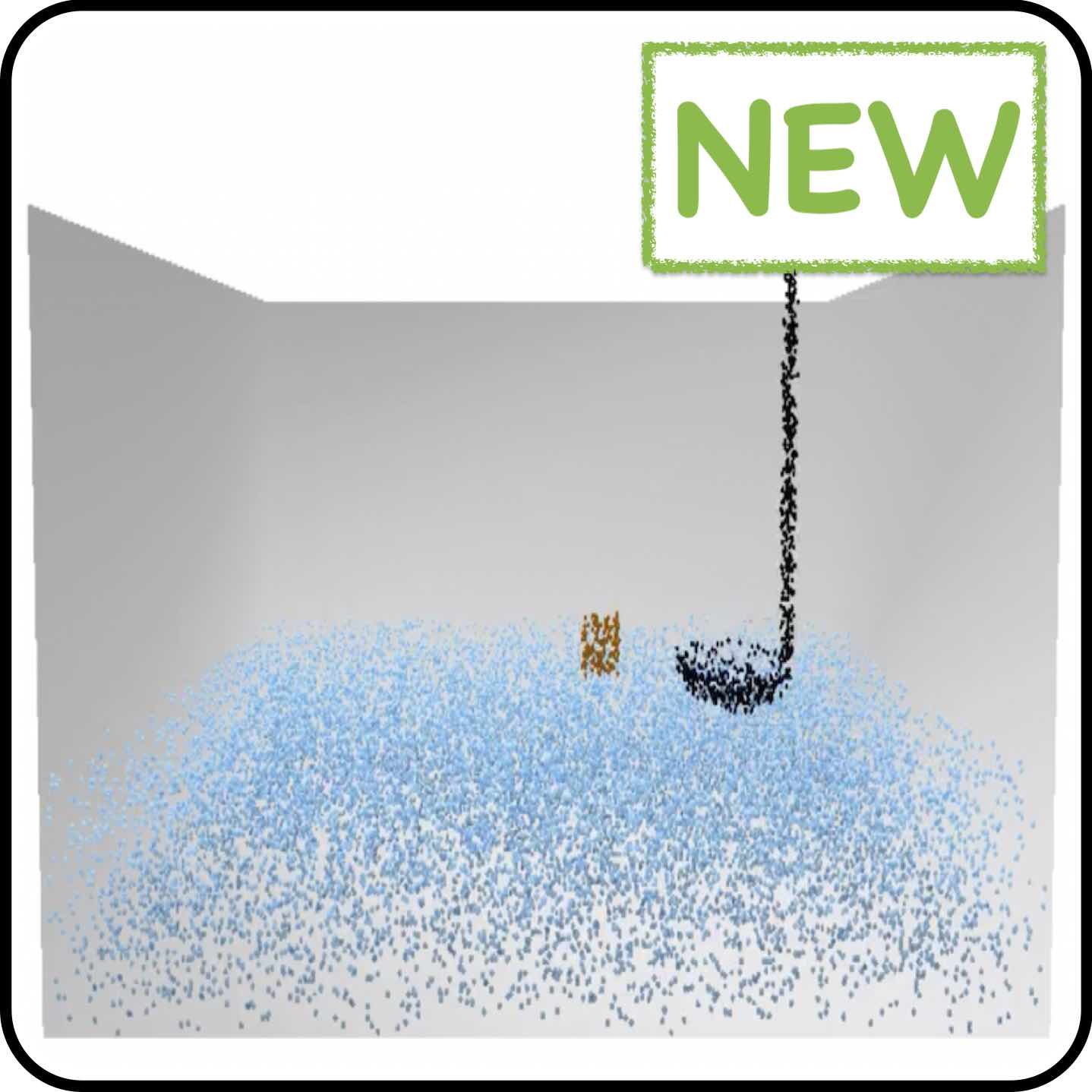}
\end{wrapfigure}
\textit{Fluid}: a particle-based fluid simulation that also involves two rigid objects (a spoon and a sugar cube) interacting in fluid. The objective is to scoop the sugar cube out of the liquid. We optimize forward-back and up-down velocity of the spoon, and let it change fives times per episode, yielding a 10-dimensional optimization problem.
\vspace{2px}

\begin{wrapfigure}[4]{l}{45pt}
    \vspace{-13pt}
    \includegraphics[width=45pt]{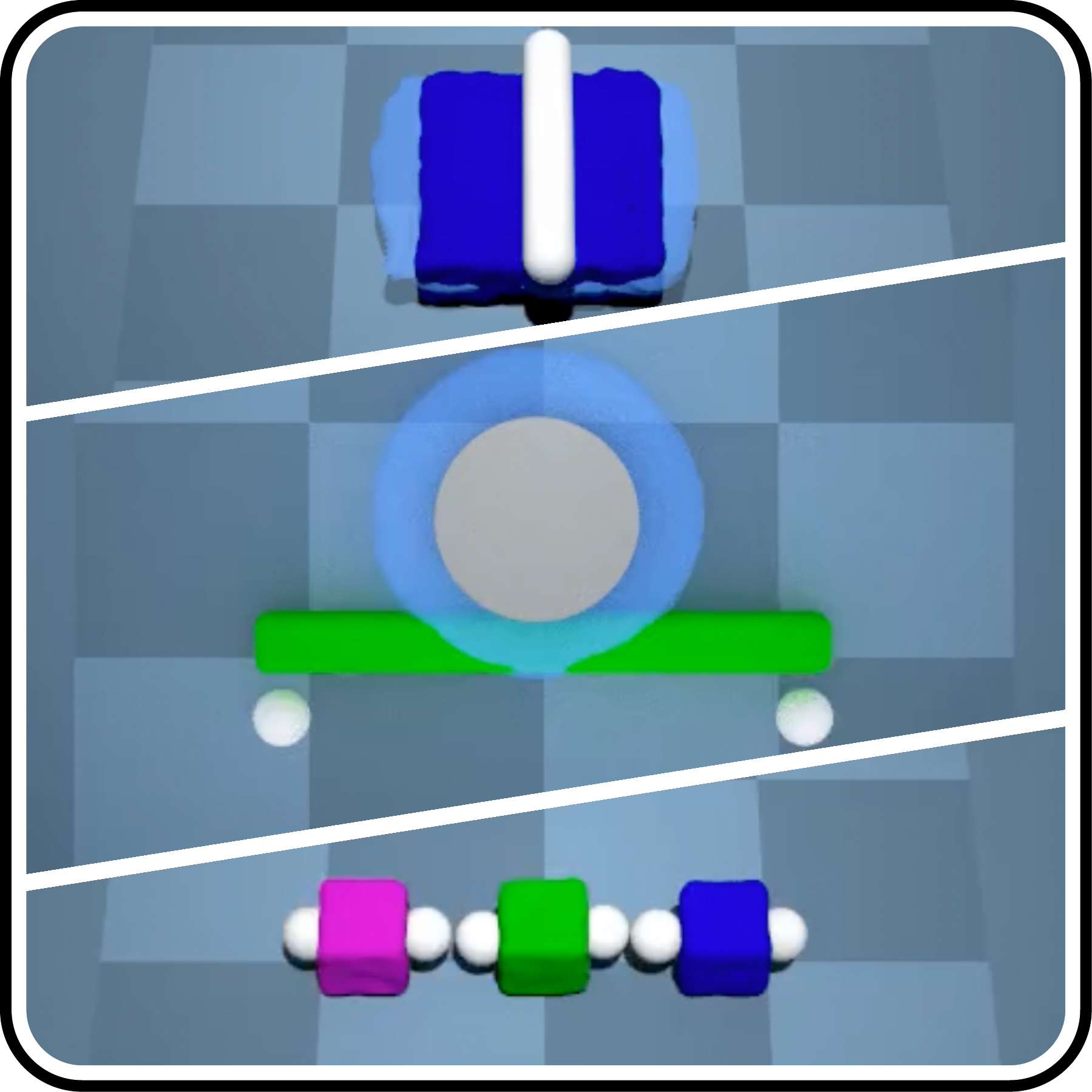}
\end{wrapfigure}
\textit{Assembly, Pinch, RollingPin, Rope, Table, Torus, TripleMove, Writer}: scenarios based on PlasticineLab~\cite{huang2021plasticinelab} with particle-based simulations of plastic deformation. We optimize 3D velocities of anchors or pins, allowing them to change five times per episode. This yields a 90D problem for \textit{TripleMove}, 30D for \textit{Assembly}, 15D for the rest.
\vspace{2px}

\begin{wrapfigure}[4]{l}{45pt}
    \vspace{-13pt}
    \includegraphics[width=45pt]{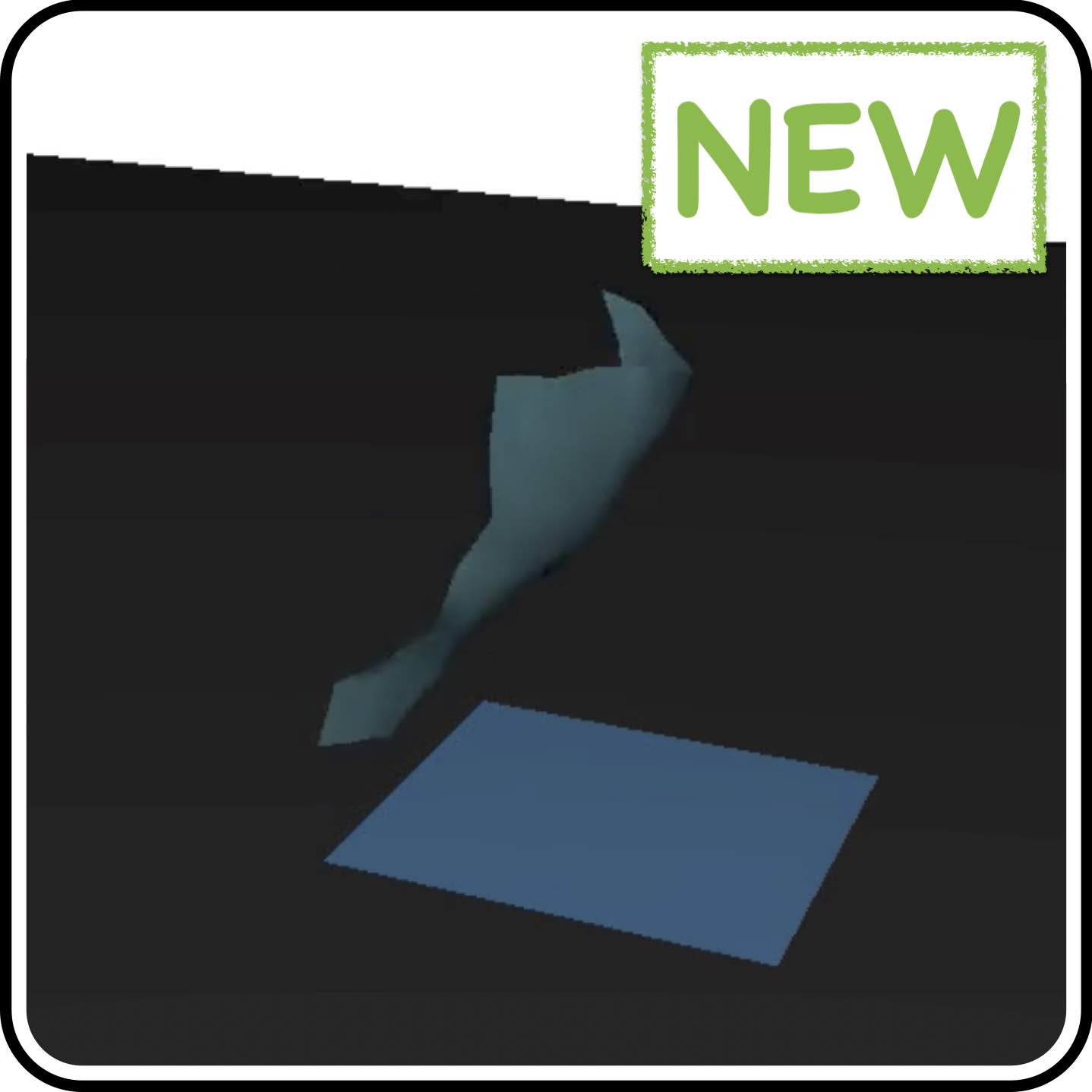}
\end{wrapfigure}
\textit{Swing}: a basic cloth swinging scenario to study dynamic tasks with deformables. We give options to optimize the speed of the anchor that swings the cloth, the cloth width \& length, and stiffness of cloth that is partitioned into $n \times m$ patches. Varying $n,m$ lets us experiment with effects of increasing dimensionality of the optimization problem.
\vspace{2px}

\begin{wrapfigure}[4]{l}{45pt}
    \vspace{-13pt}
    \includegraphics[width=45pt]{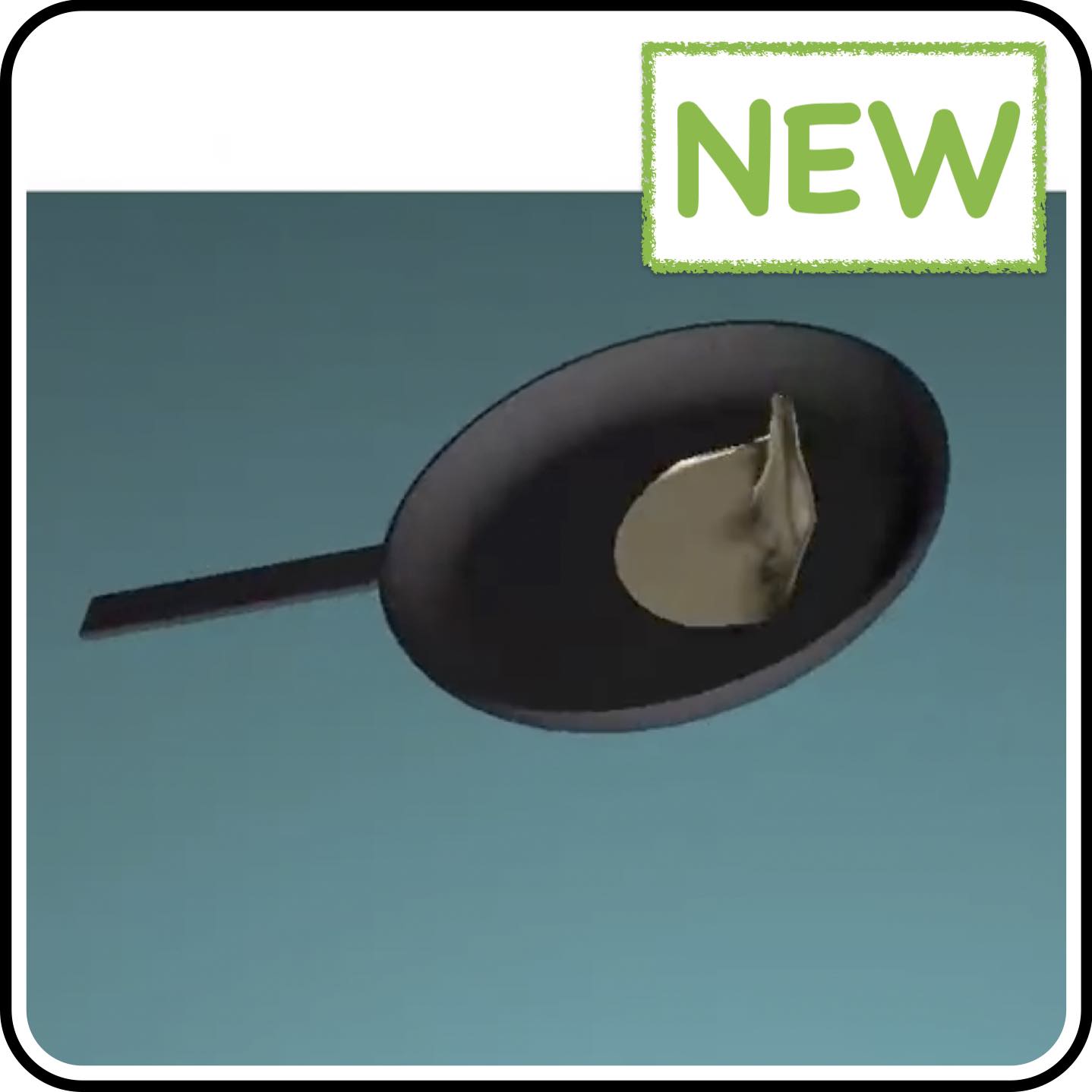}
\end{wrapfigure}
\textit{Flip}: a scenario with highly dynamic motion and rigid-deformable object collisions. The goal is to flip a pancake by moving the pan. We optimize pan motion: $n$ waypoints for left-right, up-down position and pan tilt. The pancake is modeled using a mass-spring model with small stiffness to avoid large forces from dynamic movement.
\vspace{2px}

We built this suite of environments to be representative of a wide range of possible robot manipulation scenarios, regardless of whether differentiable simulators produce sensible gradients in them. We found that some of these environments yield high-quality gradients, while others do not.
In Section~\ref{sec:sim_exp}, we show that \textit{Cartpole}, \textit{Fluid}, and the eight PlasticineLab environments produce well-behaved gradients and that our method can outperform competing baselines in these scenarios.
Then, in Section~\ref{limitations}, we show that differentiable simulators can produce incorrect gradients in highly dynamic and contact-rich environments like \textit{Pinball}, \textit{Swing}, and \textit{Flip}, which makes gradient-based methods (including our method) less effective in these environments.

\section{Experiments and Analysis of Optimization Landscapes}

In this section, we present visual analysis of optimization landscapes and gradients, as well as comparison experiments, in various environments. 
We use Rand (random search) and CMA-ES as our gradient-free baselines; we use RandDescents, an algorithm that runs multiple gradient descents with randomly sampled initial parameter value, and BO as our gradient-based baselines.

\subsection{Simulation Experiments and Analysis}
\label{sec:sim_exp}

\begin{wrapfigure}{r}{0.5\textwidth}
\vspace{-8px}
\centering
\includegraphics[width=0.5\textwidth]{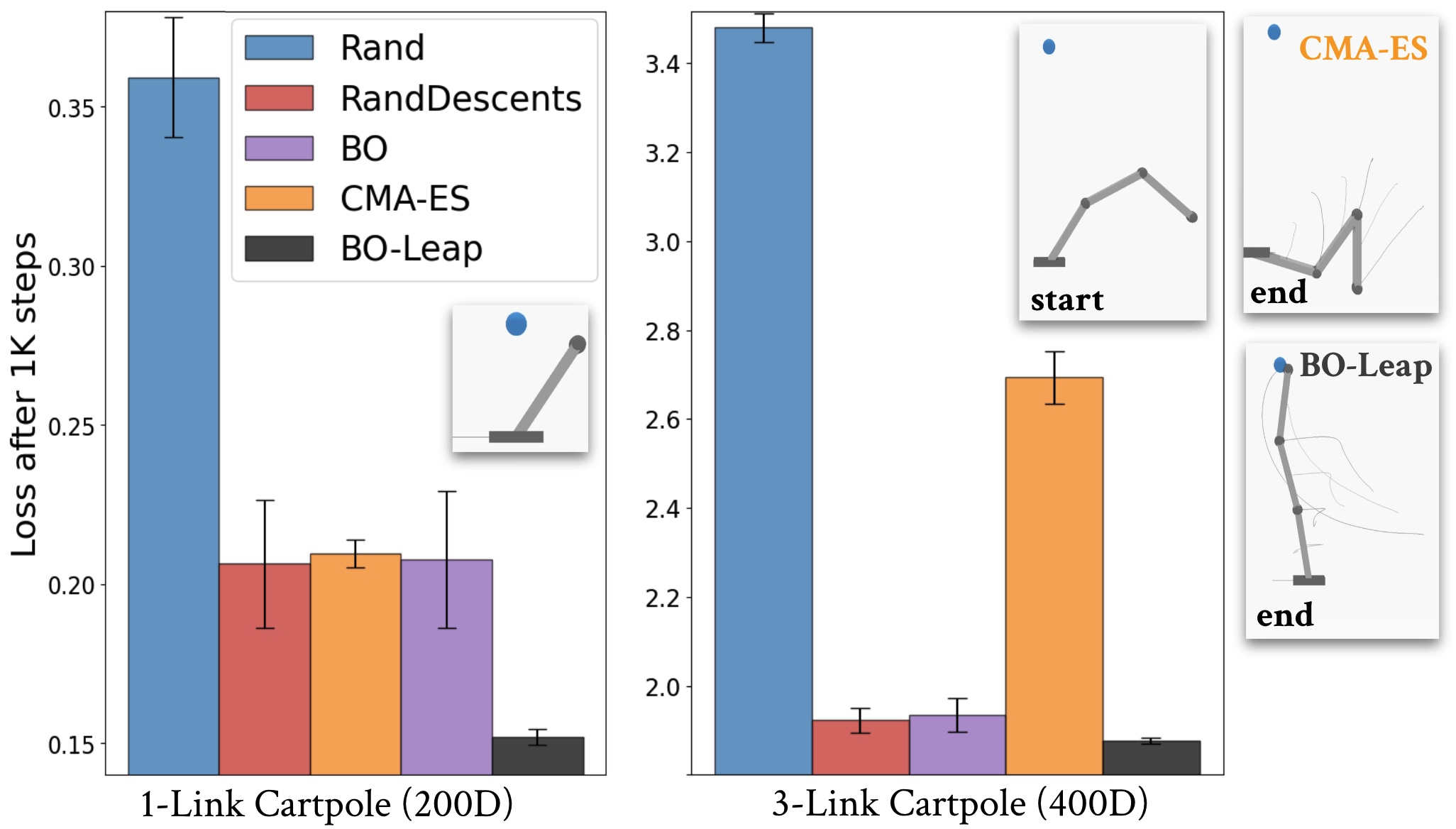}
\vspace{-15px}
\caption{Left: results for \textit{1-Link Cartpole}. Right: results for \textit{3-Link Cartpole}. For all bar charts in this paper, the vertical axis shows the mean and 95\% confidence interval of best loss value after 1,000 optimization steps. Each optimization step runs one simulation episode, computes the loss, and propagates the gradient wrt. optimization parameters if the algorithm needs it.}
\label{fig:cartpole}
\vspace{-10px}
\end{wrapfigure}

We first analyze using gradients with rigid objects. The left side of Figure~\ref{fig:cartpole} shows experiments on the \textit{1-Link Cartpole}, where gradient-free \mbox{CMA-ES} performs similar to gradient-based algorithms (RandDescents, BO), while BO-Leap obtains a significantly lower loss. The right side shows results for \textit{3-Link Cartpole} which has much more complex dynamics. Here, all gradient-based algorithms show a large improvement over gradient-free ones. CMA-ES does better than a completely random search (Rand), but fails to lift the pole tip to the target. 

On the top-right corner of Figure~\ref{fig:cartpole}, we show qualitative results from a CMA-ES run (top) and from a BO-Leap run (bottom), where BO-Leap brings the tip close to the target.


\begin{figure}[b]
\centering
\vspace{5px}
\includegraphics[width=1.0\textwidth]{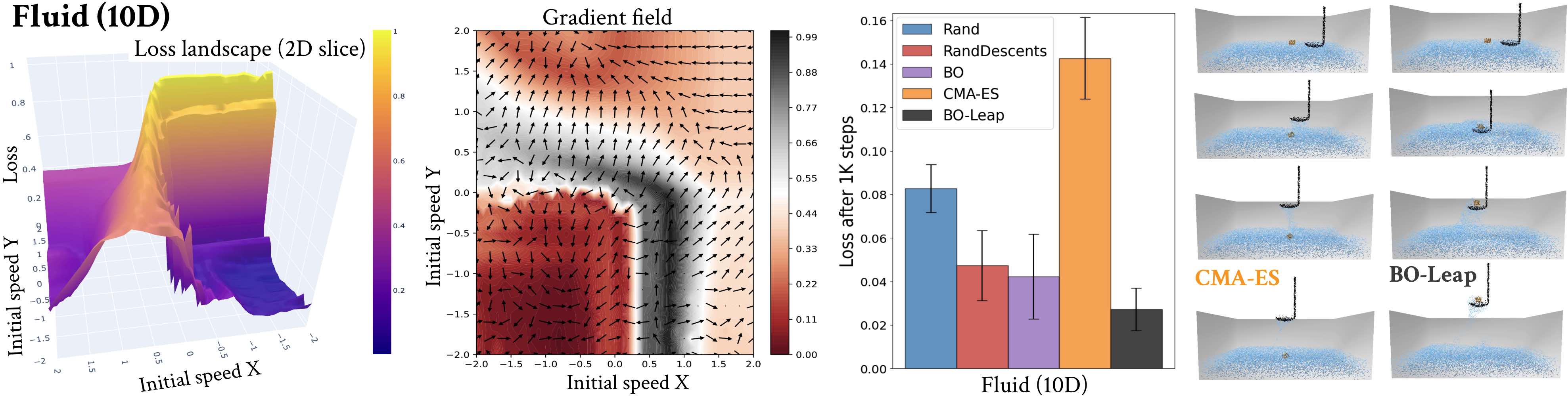}
\vspace{-10px}
\caption{A scenario with scooping up a sugar cube from fluid. The left side shows a 2D slice of the 10D optimization landscape and the corresponding gradients. To make gradient directions visible, we normalize the magnitude of gradients in all gradient plots in this paper; the arrows point towards the direction of negated gradients (i.e. the direction gradient descent updates take). The right plot shows quantitative evaluation of optimization methods. We visualize qualitative results for CMA-ES and BO-Leap on the right side.}
\label{fig:fluid}
\end{figure}

Next, we study a variety of scenarios that use particle-based simulation. Particle simulators have much larger computational and memory requirements than mesh-based options. Gradient computations further increase resource requirements. For the differentiability to be warranted, benefits of gradients have to be significant. We show that gradient-based methods can indeed have large benefits. Figure~\ref{fig:fluid} visualizes the \textit{Fluid} scenario: the left side shows a 2D slice of the loss landscape. It has many valleys with shallow local optima and appears more difficult than many test landscapes designed to challenge global optimization methods. The middle of Figure~\ref{fig:fluid} shows directions of gradients produced by the differentiable simulator. The right side shows results for gradient-based and gradient free methods. CMA-ES gets stuck in a local optimum: the spoon fails to lift the sugar cube in most runs as shown in the next-to-last column. In contrast, BO-Leap successfully lifts the sugar cube in most runs, outperforming all baselines. 
BO is designed to balance allocating trials to global exploration while still reserving enough trials to return to the well-explored regions that look promising. In this scenario such global optimization strategy proves to be beneficial. Furthermore, BO-Leap also handles the rough parts of the landscape more effectively than the baseline BO.

\begin{figure}[t]
\centering
\includegraphics[width=1.0\textwidth]{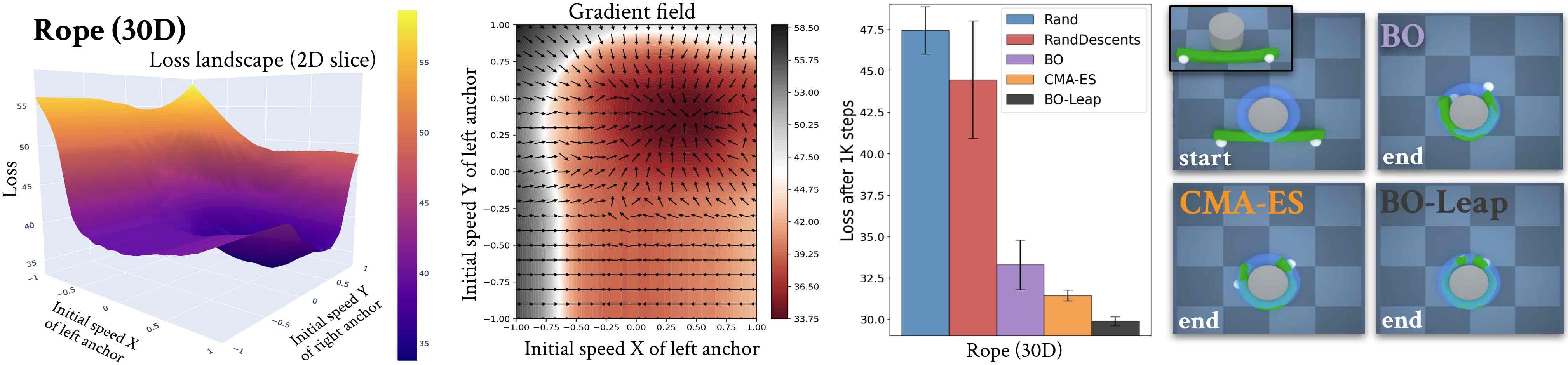}
\caption{Results for the \textit{Rope} environment. BO-Leap outperforms all other methods and successfully completes the task of wrapping the rope fully around the pole (light blue region shows the target shape). Gradient-free CMA-ES cannot reliably get the optimal behavior. While BO-Leap uses gradients effectively, this scenario shows that simply using gradients with random restarts is not sufficient: RandDescents has poor performance.}
\label{fig:plab_rope}
\vspace{-3pt}
\end{figure}
 
In the next set of experiments, we analyze eight environments from PlasticineLab~\cite{huang2021plasticinelab}. Figure~\ref{fig:plab_rope} shows a \textit{Rope} scenario, where the objective is to wrap it around a rigid cylindrical pole. 
The loss landscape is smooth in most dimensions (the left plot shows an example 2D slice). However, higher dimensionality makes the overall problem challenging. The middle plot confirms that gradients are correct in most parts, but also shows a large plateau where even the gradient-based approaches are likely to get stuck. The right side shows evaluation results: CMA-ES and BO fail to wrap the rope fully around the pole, while BO-Leap succeeds in pulling the ends on the back side of the pole correctly. This environment also shows that simply using gradients with random restarts is not sufficient even on smooth landscapes: RandDescents has poor performance that does not improve significantly over gradient-free random search.

\begin{figure}[t]
\centering
\includegraphics[width=1.0\textwidth]{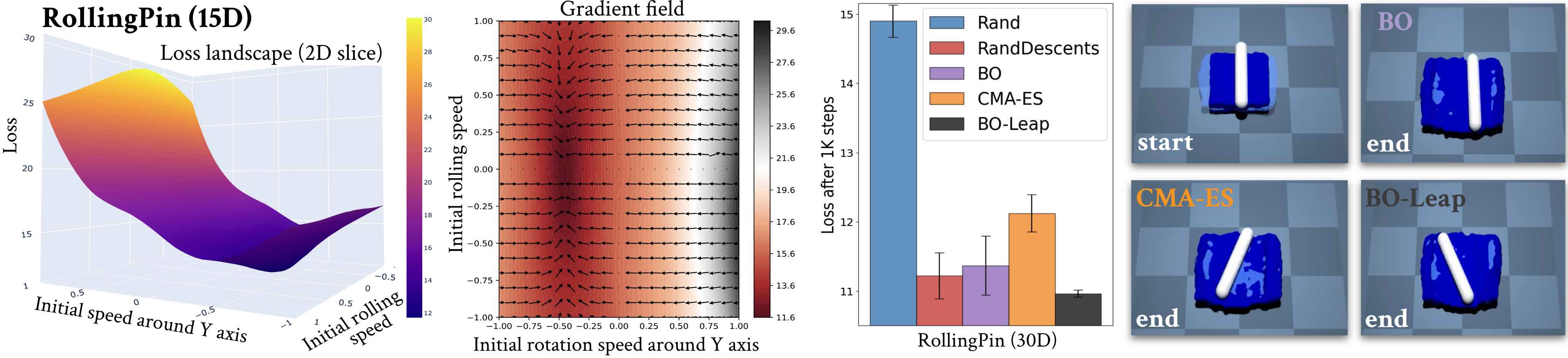}
\includegraphics[width=0.95\textwidth]{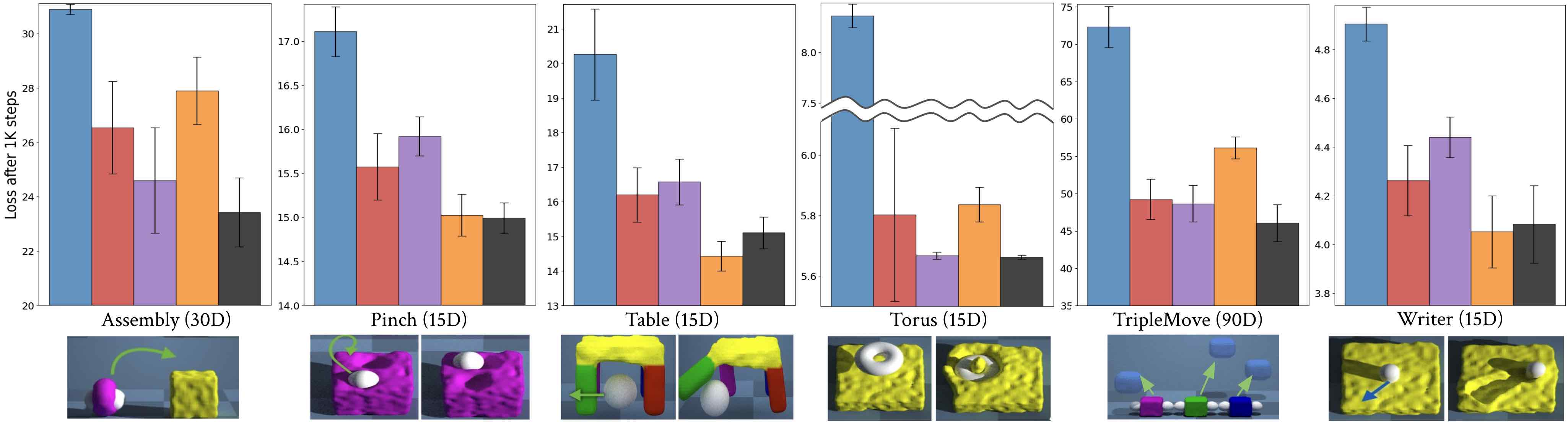}
\caption{Top row: analysis and results for \textit{RollingPin} task. Bottom rows: results for the other six PlasticineLab tasks. Plots show mean performance over 10 runs of each method per task (see supplemental for further details). Environment illustrations below the six bar charts at the bottom of this figure are borrowed from \cite{huang2021plasticinelab}.}
\label{fig:plab_pin}
\end{figure}

Figure~\ref{fig:plab_pin} shows experiments with the other seven PlasticineLab tasks. The top shows the \mbox{\textit{RollingPin}} task, where the objective is to spread the dark-blue dough using a thin rigid cylindrical white pin (the light-blue region shows the target shape). The loss landscape is smoother than that of the \textit{Rope} task, but has even larger plateaus and flat gradients in most regions, making the problem challenging despite a smooth loss. BO-Leap outperforms CMA-ES and gradient-based methods on this task as well. Example frames on the right show that CMA-ES thins out the dough too much (large gaps appear in the middle, revealing the light-blue target shape underneath). BO and BO-Leap keep the central portion of the dough more uniformly spread. The bottom plots show results for the additional six PlasticineLab tasks. BO-Leap outperforms gradient-based methods in all tasks, and achieves significantly lower loss than CMA-ES in \textit{Assembly}, \textit{Torus}, and \textit{TripleMove} tasks (see supplementary materials for further details).

\subsection{Validation in Real Robot Setup}
\label{sec:real_exp}

\begin{figure}[t]
\centering
\includegraphics[width=0.9\textwidth]{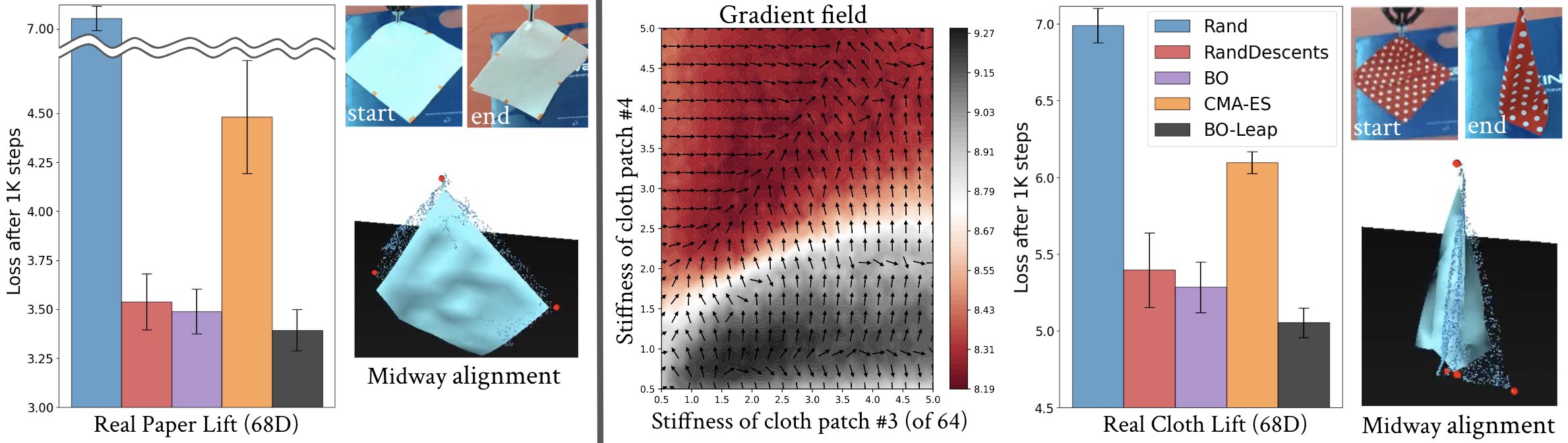}
\caption{Results for optimizing a simulated object to match the motion of a real paper (left) and cloth (right), which are lifted by a real robot. We visualize the midway alignment showing that our method finds the size, mass, friction and stiffness (for each of the 64 patches) to bring simulated object behavior close to the real one.}
\label{fig:real}
\end{figure}

To validate the proposed method using real data, we consider the task of identifying the properties of a simulated deformable object to make its motion match the real motion. In this scenario, a Gen3 (7DoF) Kinova robot manipulates a small deformable object by lifting it up from the table surface. The object is tracked by two RealSense D435 depth cameras, one placed overhead, the other on the side. 
The objective is to optimize the size (width \& length), mass, and friction of the deformable object, as well as stiffness of each of the $8 \!\times\! 8 \!=\!64$ patches of the object, which yields a 68D optimization problem. The loss penalizes the distance of the simulated corner vertices to the position of the corners marked on the real object. Figure~\ref{fig:real} shows results of experiments with two objects: a stiff but flexible paper (left) and a highly flexible cloth (right). In both cases, BO-Leap is able to find physical simulation parameters that produce the best alignment of the simulated and real object. This offers a validation of the method on real data, and shows that it can tackle `real2sim' problems to automatically bring the behavior of simulated objects closer to reality. This is valuable for highly deformable objects, since manual tuning is intractable for high dimensions.

\subsection{Limitations}
\label{limitations}

Modeling global loss posterior with BO can be computationally expensive. We use BoTorch~\cite{balandat2020botorch} for BO on GPU, which can scale to high dimensions. We focused on results within a budget of 1K optimization steps. If a much larger budget is allowed, then more tests would be needed to validate that BoTorch (or other frameworks) can scale well in terms of compute and memory resources.

\vspace{-3pt}
\begin{figure}[h]
\centering
\includegraphics[width=1.0\textwidth]{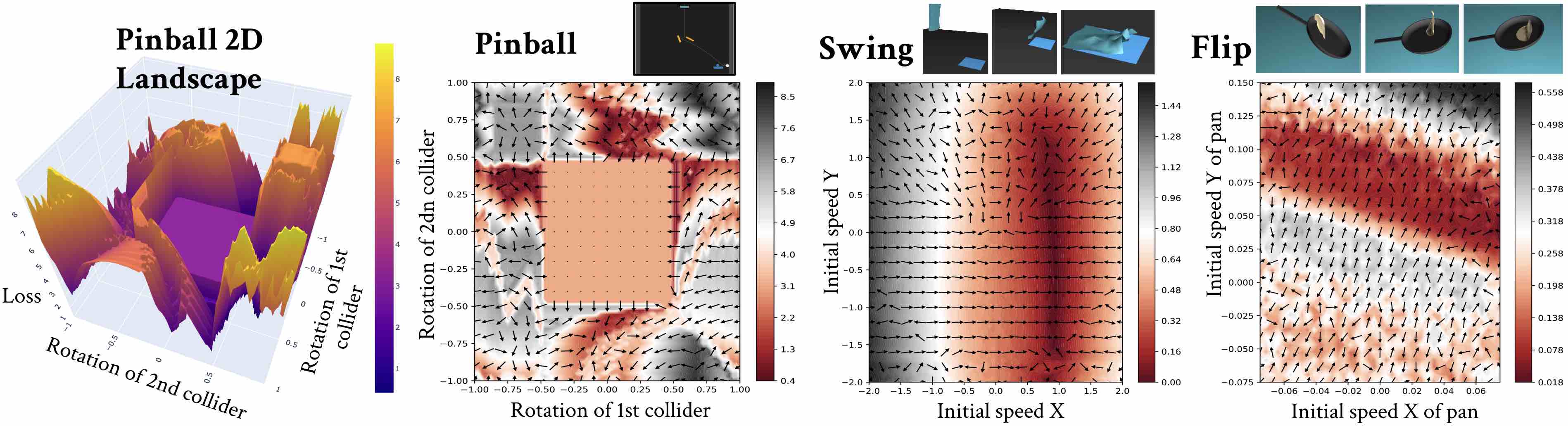}
\vspace{-10px}
\caption{Visual insights into the challenges of obtaining well-behaved gradients for cases with rigid contacts in \textit{Pinball}, and for deformables in the presence of contacts and highly dynamic tasks, such as \textit{Swing} \& \textit{Flip}. }
\vspace{-14pt}
\label{fig:pinball}
\end{figure}

The biggest challenges of optimization with differentiable simulators arise due to quality of the gradients, which can be insufficient to be beneficial for gradient-based algorithms, including our method.
In Figure \ref{fig:pinball}, we show three environments where gradients produced by differentiable simulators are of poor quality.
In the \textit{Pinball} environment, gradients with respect to collider orientations are computable only if a collision (with the pinball) has occurred to begin with. 
In addition to collision-induced discontinuities, the absence of gradients results in plateaus, affecting gradient-based optimizers.
Even in a state-of-the-art simulator Warp~\cite{warp2022} with relaxed contact models, a simple \textit{Bounce} task induces gradient discontinuities (see supplement for details).
In \textit{Swing} and \textit{Flip} tasks, while the dynamics appear realistic, differentiable simulators yield gradients with incorrect (often opposite) directions.
This is a common pitfall for practitioners who use differentiable simulators without assessing loss landscapes first.

\vspace{12px}

\textbf{Conclusion}\quad
Our analysis shows that differentiable simulation of contact-rich manipulation scenarios results in loss landscapes that are difficult for simple gradient-based optimizers.
To overcome this, we proposed a hybrid approach that combines local (gradient-based) optimization with global search, and demonstrated success on rugged loss landscapes, focusing on cases with deformables.
We believe our analyses and tools provide critical feedback to differentiable simulator designers and users alike, to take differentiable  simulators a step closer to real-world robot learning applications.

\appendix
\counterwithin{figure}{section}
\part*{\LARGE{Appendix}}
\startcontents[sections]

\renewcommand{\thesection}{\Alph{section}}
\setcounter{figure}{0}

\section{Additional Environment Descriptions and Details}

In this section, we list details of environments we covered in the paper. We describe the simulation framework, physics model, contact type, parameter information, loss configuration, as well as landscape and gradient characteristics for each environment.

\subsection{Rigid Body Environments}

\subsubsection{3-link Cartpole}

\begin{figure}[H]
    \vspace{-0.5em}
    \centering
    \includegraphics[width=\textwidth]{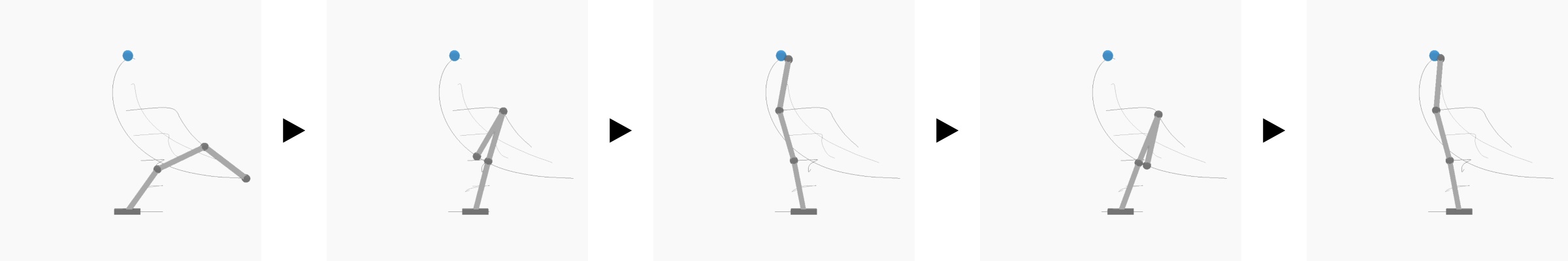}
    \caption{Illustration of the \textit{3-link Cartpole} environment.}
    \vspace{-1.7em}
\end{figure}

A cart carries a triple inverted pendulum where each link has length $1m$. Links farther away from the cart are lighter than the link attached to the cart (for easier control). The goal is to move the cart and actuate the joints so that the tip of the pendulum is as close as possible to a preset goal location.
\begin{itemize}[leftmargin=1.5em]
    \item \textbf{Simulation Framework or Physics Model:} Nimble~\cite{werling2021fast}
    \item \textbf{Types of Contacts:} none (there is no self-contact between different parts of the cartpole)
    \item \textbf{Parameter Dimensionality:} 400
    \item \textbf{Parameter Description:} at each of the 200 timesteps, the parameters specify cart velocity with 1 dimension and torques of 3 joints.
    \item \textbf{Loss:} L2 distance from final tip position to target position.
    \item \textbf{Landscape and Gradient Characteristics:} landscape is very smooth. Gradient quality is good.
\end{itemize}

\vspace{10px}
\subsubsection{Pinball}

\begin{figure}[H]
    \vspace{-0.5em}
    \centering
    \includegraphics[width=\textwidth]{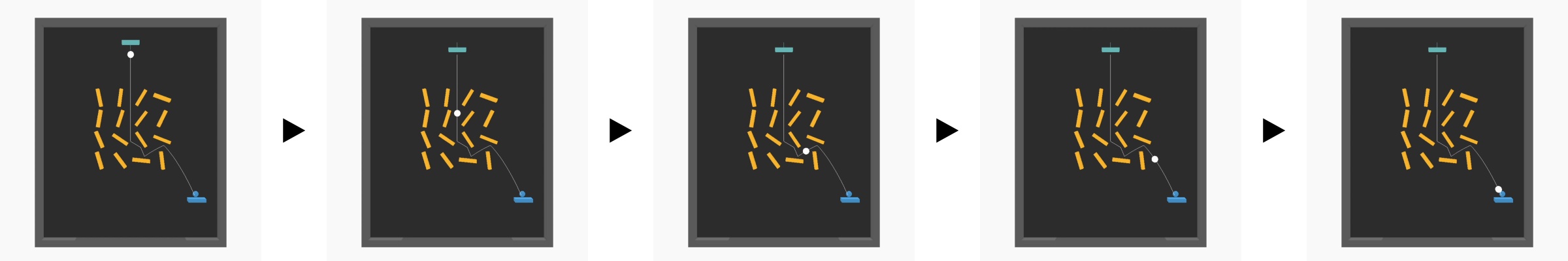}
    \caption{Illustration of the \textit{Pinball} environment with 16 colliders.}
\end{figure}

On a vertical platform of 8m wide and 10m tall, a ball is dropped on to a grid of $n_h \times n_w$ spinning colliders that have one revolute joint each attached to the platform. The goal is to guide the ball to a goal position at the end of the episode by adjusting the orientation of each collider.
\begin{itemize}[leftmargin=1.5em]
    \item \textbf{Simulation Framework or Physics Model:} Nimble~\cite{werling2021fast}
    \item \textbf{Types of Contacts:} rigid (pinball collides with colliders and walls)
    \item \textbf{Parameter Dimensionality:} $n_h \times n_w$ (in this work, we consider two setups with $n_h = 1, n_w = 2$ and $n_h = 4, n_w = 4$)
    \item \textbf{Parameter Description:} rotation angle of each spinning collider in the $n_h \times n_w$ grid.
    \item \textbf{Loss:} L2 distance from final pinball position to target pinball position near the bottom-right of the platform.
    \item \textbf{Landscape and Gradient Characteristics:} landscape has large flat regions as well as discontinuities. Gradients are zero in flat regions and not useful at locations of discontinuities.
\end{itemize}

\subsection{Deformable Object Environments}

\subsubsection{Fluid}

\begin{figure}[H]
    \vspace{-0.5em}
    \centering
    \includegraphics[width=\textwidth]{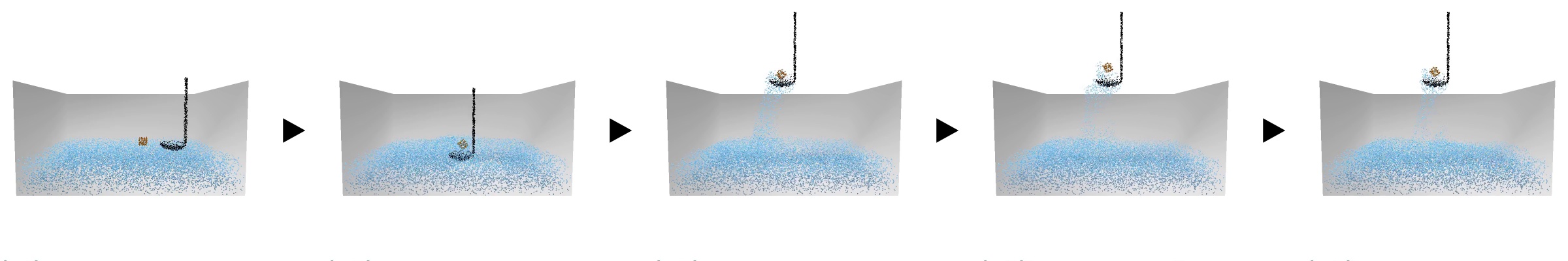}
    \includegraphics[width=\textwidth]{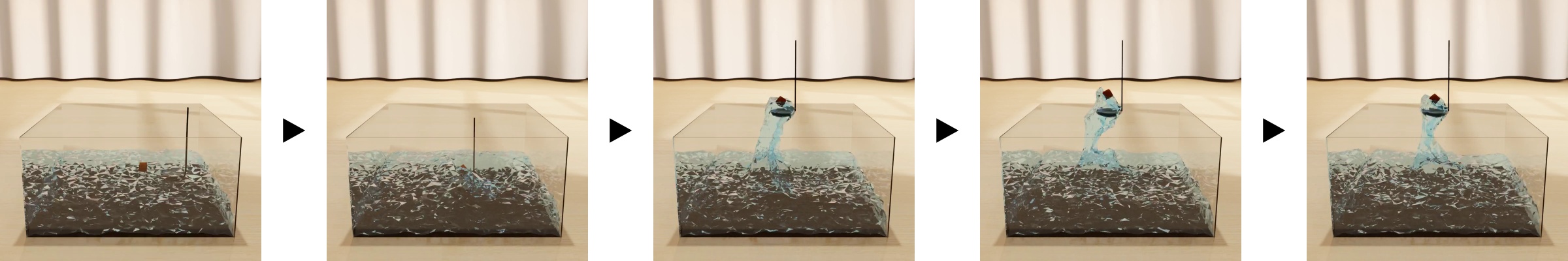}
    \caption{Illustration of the \textit{Fluid} environment. The upper and lower rows are renderings of the same episode in the same environment. The upper row uses the built-in realtime rendering engine in DiffTaichi; while the lower row uses Blender, which is slower but higher quality.}
    \vspace{-1.7em}
\end{figure}

A ladle with 2 DOF is manipulated to scoop a sugar cube from a tank of transparent syrup with width 0.4m, depth 0.4m, and height 0.2m. The goal is to scoop the cube to as high of a position as possible while being close to the ladle and the center vertical axis of the tank. This environment is made with DiffTaichi. The dynamics of syrup and the sugar cube in this environment are modeled with MLS-MPM~\cite{hu2018moving}, a state-of-the-art particle-based fluid simulation method.
\begin{itemize}[leftmargin=1.5em]
    \item \textbf{Simulation Framework or Physics Model:} DiffTaichi~\cite{hu2019difftaichi} with MLS-MPM~\cite{hu2018moving} as physics model for fluid and the sugar cube
    \item \textbf{Types of Contacts:} collision between rigid objects and fluid (such as fluid particles bouncing back off container walls)
    \item \textbf{Parameter Dimensionality:} 10
    \item \textbf{Parameter Description:} an episode splits into 5 equal-length segments. In each segment, 2 parameter values control the horizontal (forward-backward) and vertical (up-down) speed of the ladle. Note that although the ladle might not directly make contact with the cube, the ladle can push the liquid particles, which can then push the cube away from the ladle.
    \item \textbf{Loss:} $\ \text{max}(0, y - y_w) + 3 \cdot \sqrt{(x - x_s)^2 + (z - z_s)^2} + \sqrt{(x - x_c)^2 + (z - z_c)^2},$ where $(x, y, z)$ denotes the final sugar cube position, $(x_s, y_s, z_s)$ denotes final ladle body center position, and $y_w = 0.2$ denotes height of the container. In this and all following DiffTaichi environments, the $x$-axis points rightward, $y$-axis points upward, $z$-axis points to the front.
    \item \textbf{Landscape and Gradient Characteristics:} landscape is rugged and has many local minima. Gradient quality is good.
\end{itemize}

\subsubsection{PlasticineLab Environments}

\begin{figure}[H]
    \vspace{-0.6em}
    \centering
    \includegraphics[width=0.95\textwidth]{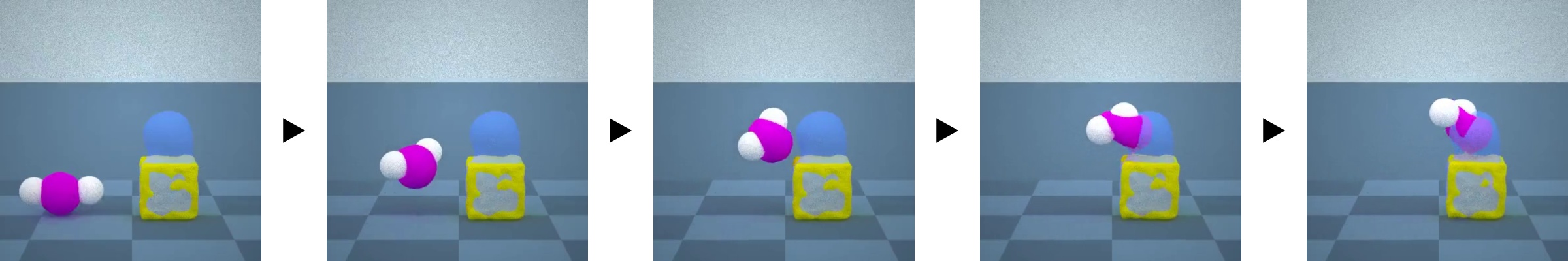}
    \includegraphics[width=0.95\textwidth]{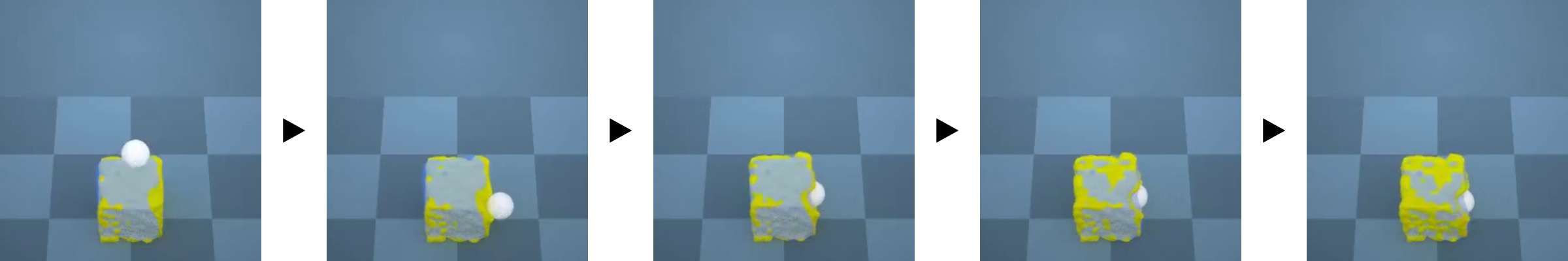}
    \includegraphics[width=0.95\textwidth]{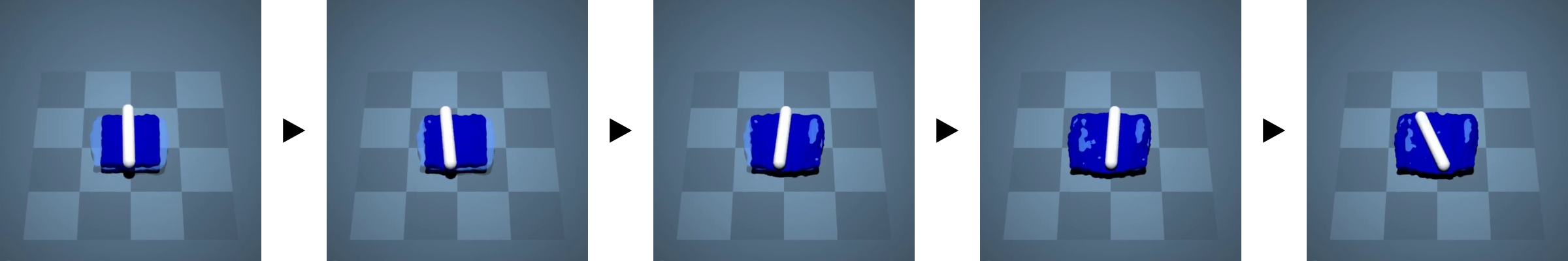}
    \includegraphics[width=0.95\textwidth]{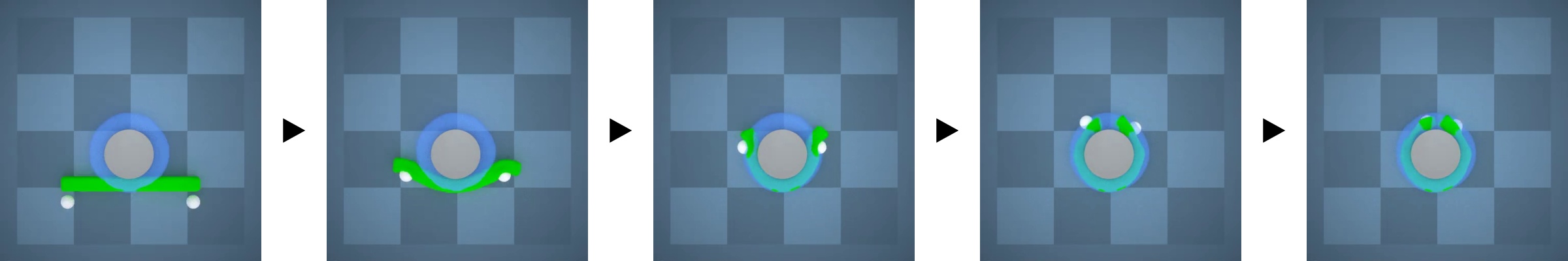}
    \includegraphics[width=0.95\textwidth]{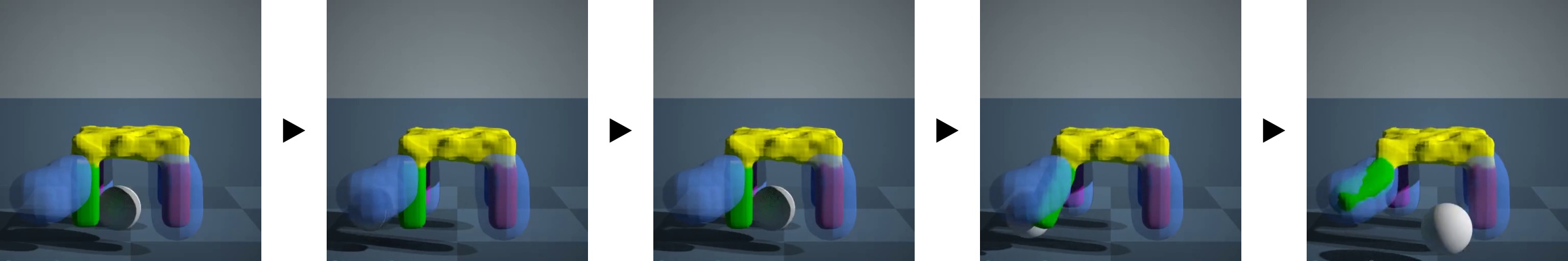}
    \includegraphics[width=0.95\textwidth]{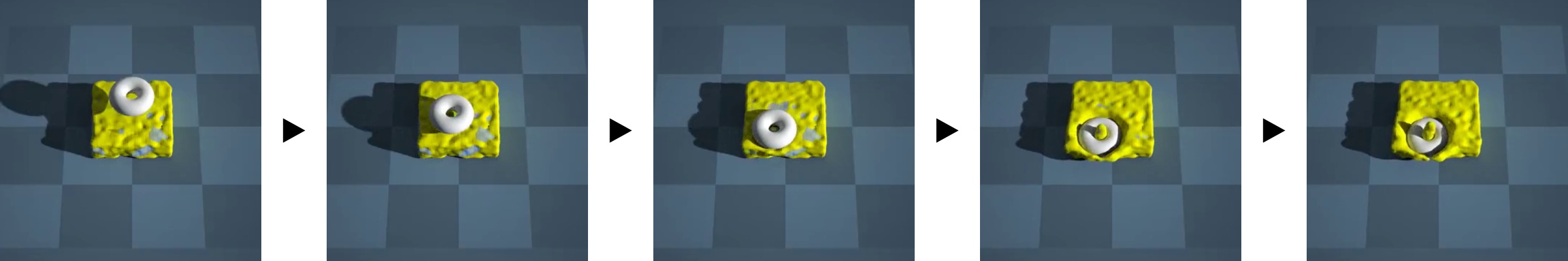}
    \includegraphics[width=0.95\textwidth]{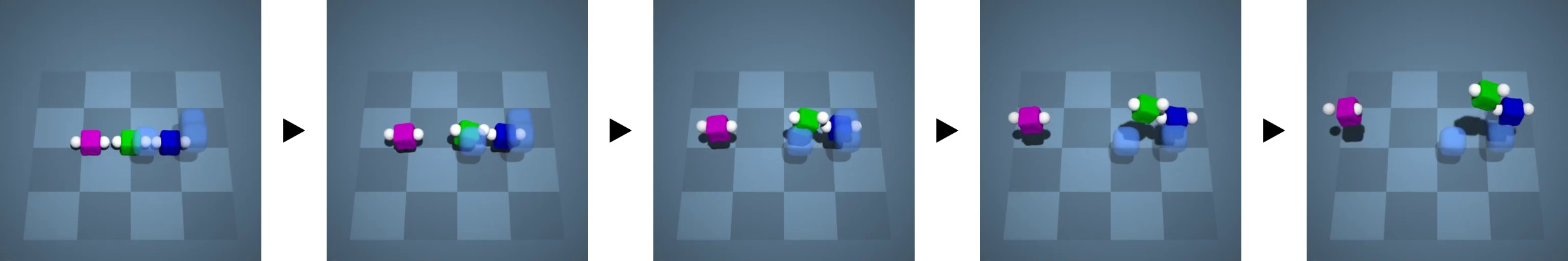}
    \includegraphics[width=0.95\textwidth]{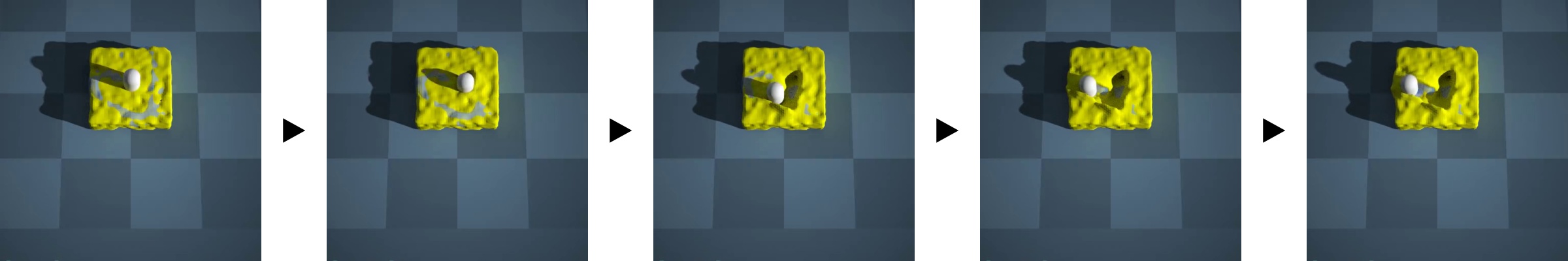}
    \caption{Illustration of environments derived from \textit{PlasticineLab}. From top to bottom, we show \textit{Assembly}, \textit{Pinch}, \textit{RollingPin}, \textit{Rope}, \textit{Table}, \textit{Torus}, \textit{TripleMove}, and \textit{Writer}.}
    \vspace{-1.5em}
\end{figure}

We consider 8 different environments derived from PlasticineLab~\cite{huang2021plasticinelab} --- Assembly, Pinch, RollingPin, Rope, Table, Torus, TripleMove, and Writer. These environments involve 1-3 anchors or a pin manipulating one or several pieces of deformable objects. The goal of all environments are to make the final deformable object configuration close to a target shape. When we adapt the environments for our purpose, we only modify the format of optimizable parameters and leave other aspects of the environments such as dynamics, episode length, and loss formulation unchanged. As we already described the high-level objectives of several PlasticineLab environments we analyzed in detail in the main paper, we direct readers to the original paper~\cite{huang2021plasticinelab} for additional details of each environment.
\begin{itemize}[leftmargin=1.5em]
    \item \textbf{Simulation Framework or Physics Model:} PlasticineLab is based on DiffTaichi~\cite{hu2019difftaichi}; its environments use MLS-MPM~\cite{hu2018moving} to model interactions between rigid and deformable objects; rigid bodies are modeled using signed distance fields (SDF)
    \item \textbf{Types of Contacts:} collision between rigid and deformable objects
    \item \textbf{Parameter Dimensionality:} 90 (TripleMove), 30 (Assembly, Rope), 15 (Pinch, RollingPin, Table, Writer)
    \item \textbf{Parameter Description:} an episode splits into 5 equal-length segments. In each segment, the 3D velocities of each anchor or pin are controlled by optimizable parameters. In TripleMove, there are six anchors to be controlled, so the optimizable parameter has a total of [5 segments $\times$ 6 anchors $\times$ 3 velocity values = 90] dimensions. In Assembly and Rope, there are two anchors to be controlled, so the parameter has [5 segments $\times$ 2 anchors $\times$ 3 velocity values = 30] dimensions. In Pinch, Table, and Writer, there is one anchor, so the parameter has [5 segments $\times$ 1 anchor $\times$ 3 velocity values = 15] dimensions. In RollingPin, instead of controlling 3D velocity of the pin in each segment, the environment uses 3 values to control the left-right, up-down, and top-down tilting angle of the pin, leading to [5 segments $\times$ 1 anchor $\times$ 3 control parameters = 15] dimensions.
    \item \textbf{Loss:} PlasticineLab uses a 3-part loss that encourages the anchors or pins to be closer to the deformable objects while penalizing the distance between the final deformable object shape and the target shape. The only modification we made to the original loss function is increasing the weight of the loss component that encourages the manipulators to get close to the target shape. For more details of the original loss function, please refer to Section 3.1 of the original paper~\cite{huang2021plasticinelab}).
    \item \textbf{Landscape and Gradient Characteristics:} the landscape is smooth with local minima, while the high dimensionality makes the problem challenging; gradient quality is good.
\end{itemize}

\vspace{10px}
\subsubsection{Swing}

\begin{figure}[H]
    \vspace{-0.5em}
    \centering
    \includegraphics[width=\textwidth]{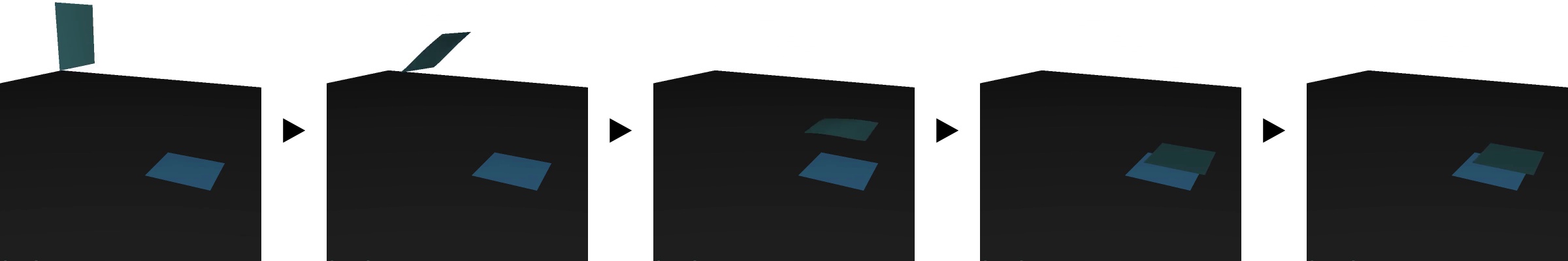}
    \includegraphics[width=\textwidth]{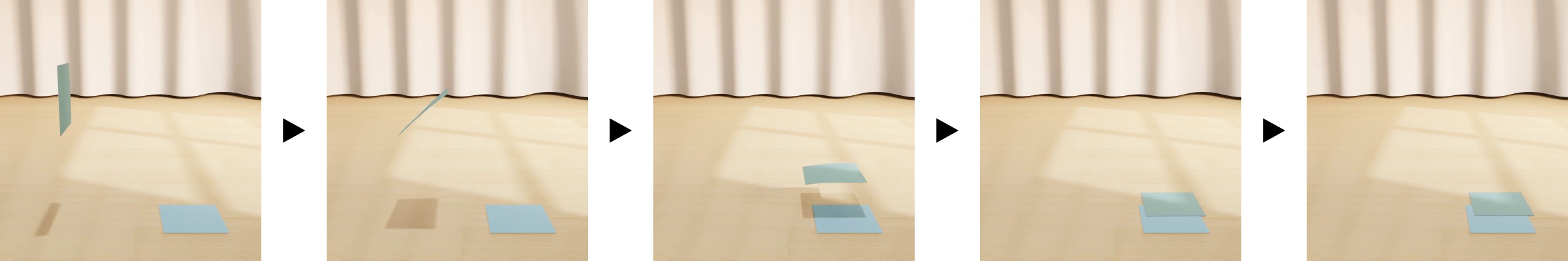}
    \caption{Illustration of the \textit{Swing} environment. The upper and lower rows are renderings of the same episode in the same environment. The upper row uses the built-in realtime rendering engine in DiffTaichi; while the lower row uses Blender, which is slower but has higher quality.}
    \vspace{-1.7em}
\end{figure}

Two anchors grasp the two corners of a $20 \times 20$cm piece of cloth and swing it onto the floor. The goal is to make the final cloth configuration as close as possible to a goal configuration.
\begin{itemize}[leftmargin=1.5em]
    \item \textbf{Simulation Framework or Physics Model:} DiffTaichi with mass-spring model as cloth simulation technique; to handle contact, we update the velocities of cloth vertices when contacts occur so that the updated speed is perpendicular to the normal of the contact surface
    \item \textbf{Types of Contacts:} collision between rigid (floor) and deformable objects (cloth)
    \item \textbf{Parameter Dimensionality:} $16$ if stiffness is optimized; $3$ if initial speed is optimized
    \item \textbf{Parameter Description:} we have two different parameter setups in this task. In the first setup, we split the cloth into $4 \times 4 = 16$ cloth patches and fix the swinging motion. The stiffness values of the 16 cloth patches are optimized. In the second setup, we fix the stiffness of the cloth and optimize the initial 3D velocity of the cloth.
    \item \textbf{Loss:} we have three different loss formulations for this task --- loss formulation `single' optimizes the distance between the center of mass of the cloth at the final frame to a goal position. Loss formulation `corner' optimizes the average distance between the four corners of the cloth to their corresponding goal positions. Loss formulation `mesh' optimizes the average distance between final vertex positions of the cloth to a target cloth mesh.
    \item \textbf{Landscape and Gradient Characteristics:} landscape is rugged. Gradients are noisy in large areas of the parameter space.
\end{itemize}

\subsubsection{Flip}

\begin{figure}[H]
    \vspace{-0.5em}
    \centering
    \includegraphics[width=\textwidth]{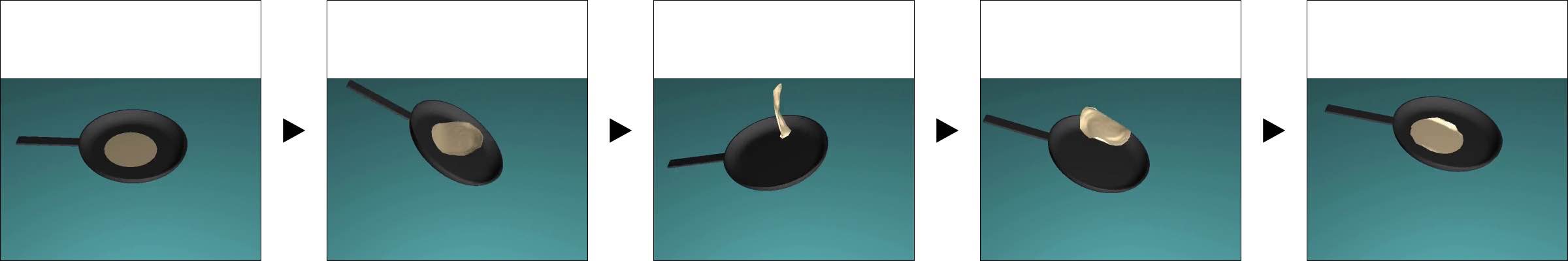}
    \includegraphics[width=\textwidth]{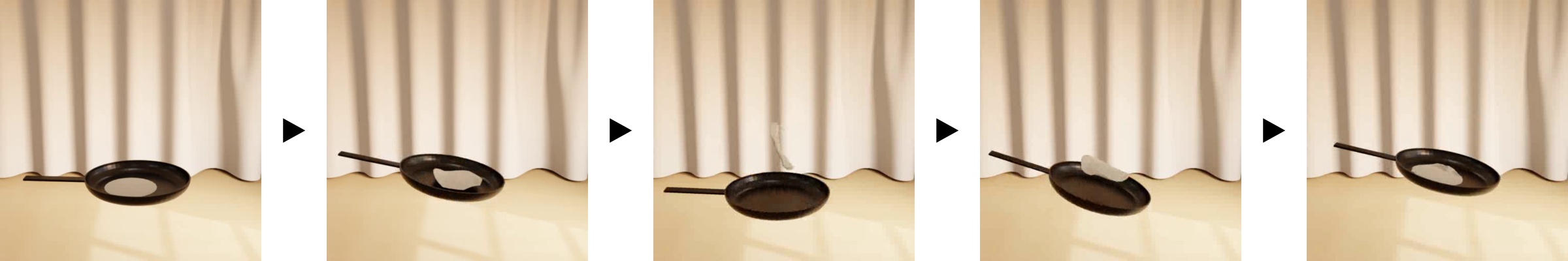}
    \caption{Illustration of the \textit{Flip} environment. The upper and lower rows are renderings of the same episode in the same environment. The upper row uses the built-in realtime rendering engine in DiffTaichi; while the lower row uses Blender, which is slower but has higher quality.}
    \vspace{-1.7em}
\end{figure}

A pancake is placed in a pan with 20cm radius and smooth edges that move and tilt with 3 DOF. The goal is to manipulate the pan so that the pancake is flipped at the end of the episode.
\begin{itemize}[leftmargin=1.5em]
    \item \textbf{Simulation Framework or Physics Model:} DiffTaichi with mass-spring model for simulating the pancake; compared to the Swing task, the stiffness value in this task is smaller to make sure collision forces are not too large during the dynamic movement of the pancake; to handle contact, we update the velocities of pancake vertices when contacts occur so that the updated speed is perpendicular to the normal of the contact surface
    \item \textbf{Types of Contacts:} collision between rigid (floor) and deformable objects (pancake)
    \item \textbf{Parameter Dimensionality:} $15$
    \item \textbf{Parameter Description:} an episode is split into 5 equal-length segments. The parameters set the $x$ (left-right) and $y$ (up-down) positions as well as the tilt of the pan at the end of each segment. This leads to [5 segments $\times$ 3 control parameters = 15] dimensions; position and angular velocity of the pan is linearly interpolated within each segment.
    \item \textbf{Loss:} average L2 distance between final position of the four pancake corners (the four pancake vertices with highest and lowest $x$ and $y$ values at the start of the episode) and four corresponding target positions.
    \item \textbf{Landscape and Gradient Characteristics:} landscape is extremely rugged. Gradients are noisy in large areas of the parameter space.
\end{itemize}

\pagebreak
\section{Additional Analysis}

\subsection{Analysis of Challenges with Gradients for Rigid Contacts}

In this section, we analyze the quality of gradients through a differentiable physics engine; observing the nature of discontinuities induced by contact.
We define the \textit{Bounce} task, where the goal is to steer a bouncing (red) ball with known friction and elasticity parameters to a target position (green). We seek a policy that imparts an initial 3D velocity $\mathbf{v}_{\textnormal{init}}$ to a ball such that, at the end of the simulation time $t_\textnormal{max}$, the center of mass of the ball achieves a pre-specified target position. Importantly, the policy must shoot the ball onto the ground plane, and upon a bounce, reach the target location (this is achieved by restricting the cone of velocities to contain a vertically downward component).
This enforces at least one discontinuity in the forward simulation. 

We use the relaxed contact model from Warp~\cite{warp2022}, and compute the gradients for a wide range of initial velocities ($-10$ to $10 \  m/s$ along both $X$ and $Y$ directions). Notice that, in the $X$-direction (horizontal speeds), the gradients are smoother, as changes to $X$ components of the velocity only push the ball further  (closer) to the goal, and have little impact on the discontinuities (bounces).
However, the $Y$-axis components of velocities (vertical speeds) tend to have a significant impact on the location and nature of the discontinuities, and therefore induce a larger number of local optima.

\begin{figure}[H]
\centering
\includegraphics[width=0.9\textwidth]{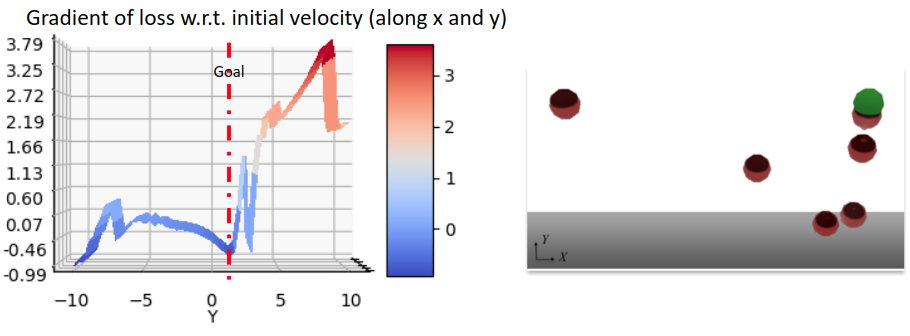}
\vspace{-8px}
\caption{A simple rigid body \textit{Bounce} task implemented using Warp~\cite{warp2022}. The goal is to impart an initial velocity to the red ball so that it reaches the target location (green) at the end of 2 seconds. The ball moves in 3D: $X$ left-right, $Y$ down-up, $Z$ in-out of the image plane. We restrict out attention the $Y$ direction and observe that discontinuities caused by the rigid contacts remain, even in Warp -- a recent framework with semi-implicit Euler integration and advanced relaxed-contact and stiffness models.}
\label{fig:bounce}
\end{figure}

While prior work (e.g., DiffTaichi~\cite{hu2019difftaichi}) has extensively analyzed gradients through similar contact scenarios, they leverage the conceptually simple (but numerically unstable) Euler integrator and perfectly elastic collisions.
Warp~\cite{warp2022}, on the other hand, uses both a symplectic time integrator and a contact model that includes both friction and elasticity parameters.
Our \textit{Bounce} experiments confirm the fact that significant challenges with computing gradients through rigid contacts remain, even in these more recent and advanced differentiable simulation frameworks.

\subsection{Landscape Gallery}

In this section, we present more landscape and gradient plots to provide more insight into the differentiable simulation environments we presented in the paper. Apart from this section, we also present animated landscape plots in the supplementary video.

\textbf{Pinball}\quad
Below, we show two landscape and gradient plots, one plotting the landscape of the \textit{Pinball} 2D environment with two colliders, and the other plotting a 2D slice for the \textit{Pinball} 16D environment with a grid of 4-by-4 colliders. In the right plot, the x and y axes correspond to the rotation of the center two colliders at the bottom of the collider grid. From the plots, we see that the rugged landscapes occur in different variations of the \textit{Pinball} task.

\begin{figure}[H]
    \centering
    \includegraphics[width=0.30\textwidth]{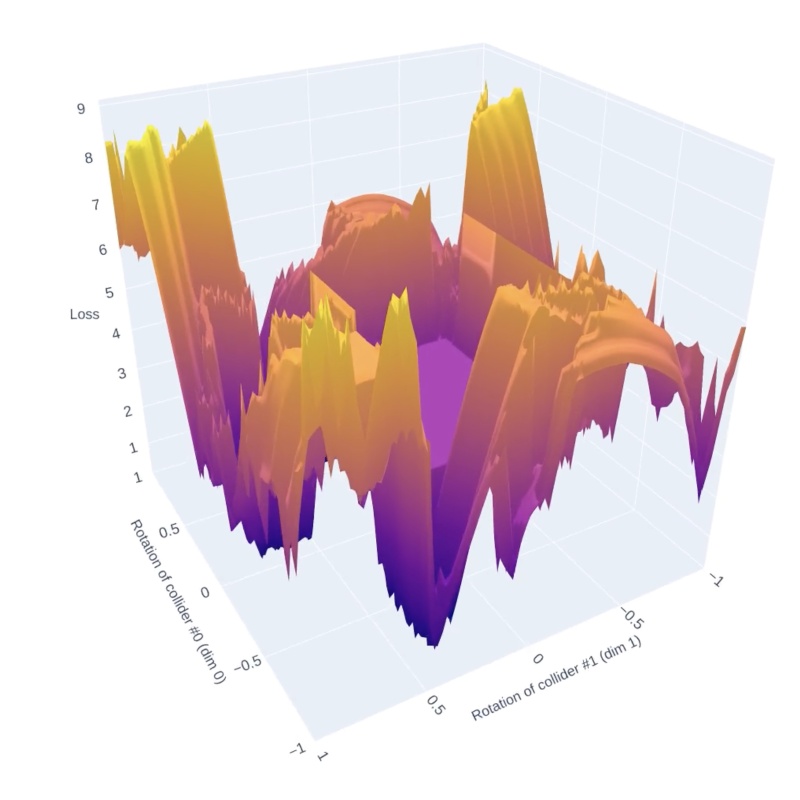}
    \includegraphics[width=0.30\textwidth]{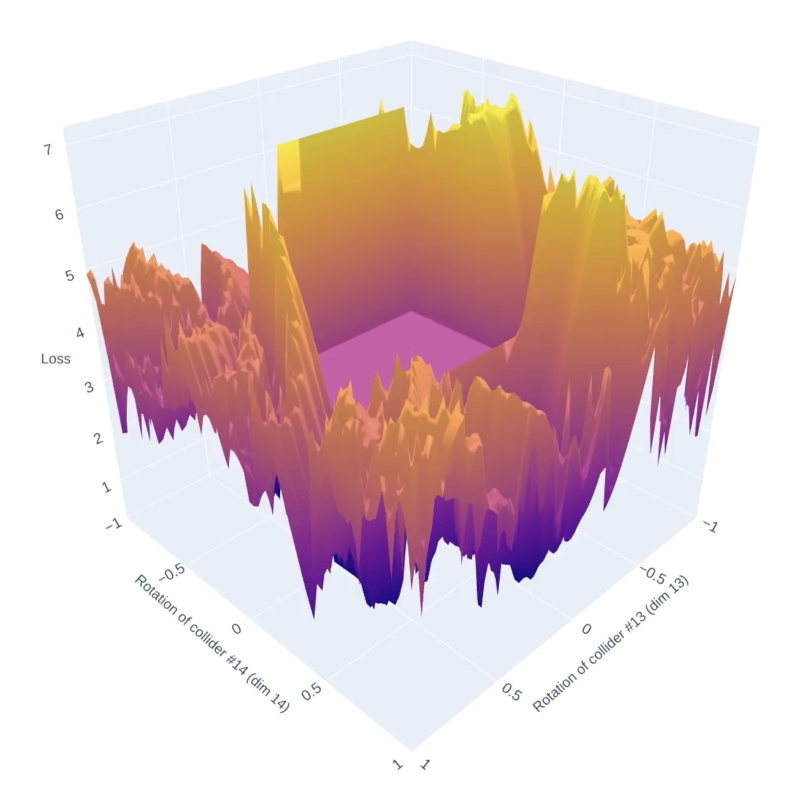} \\
    \includegraphics[width=0.30\textwidth]{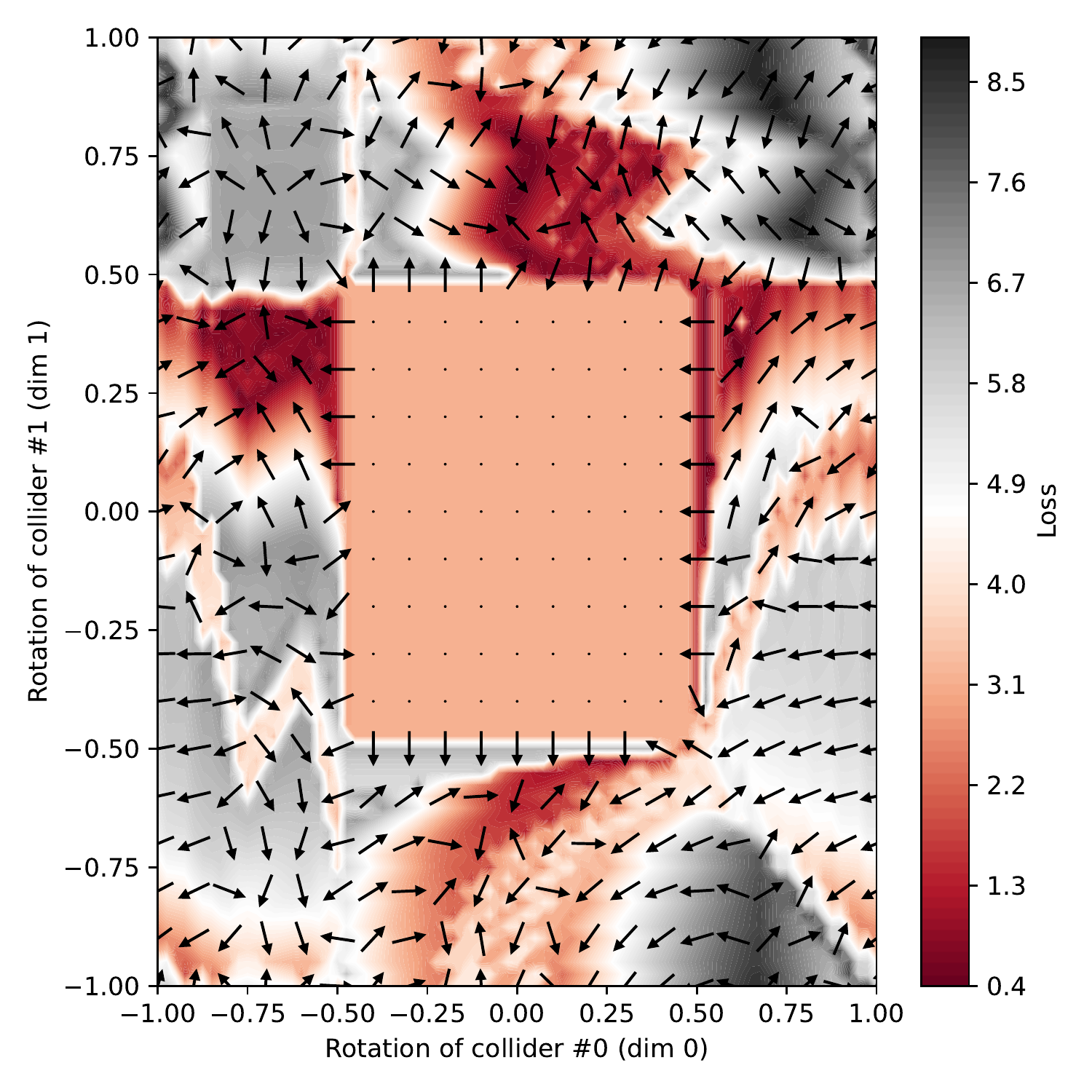}
    \includegraphics[width=0.30\textwidth]{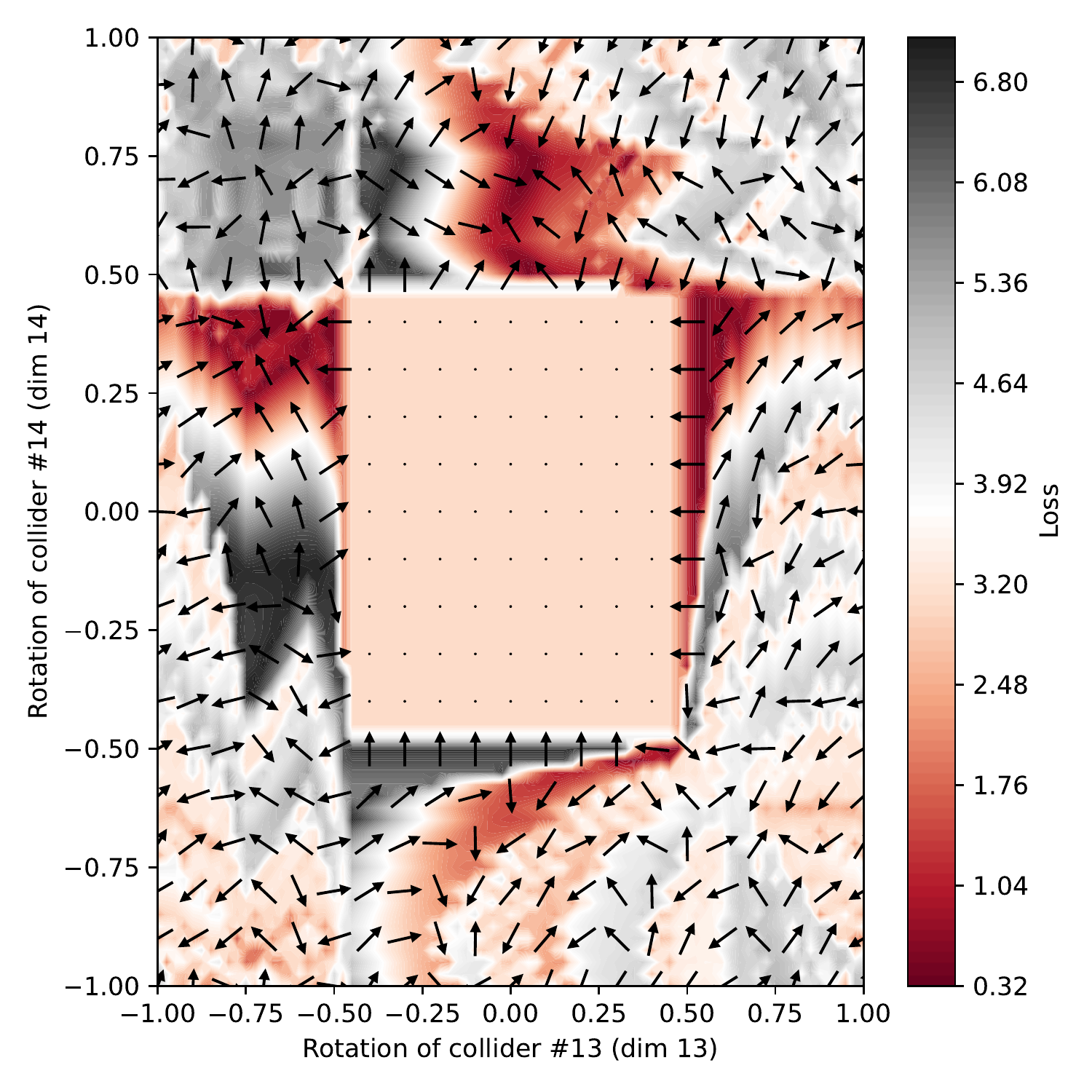}
    \caption{Landscape (top) and gradient (bottom) plots for Pinball environment. \textit{Left column} -- Pinball 2D dimensions 0 and 1. \textit{Right column} -- Pinball 16D dimensions 13 and 14.}
\end{figure}

\textbf{Fluid}\quad
In the main paper, we presented \textit{Fluid} as an environment where the landscape is rugged and showed one 2D slice of the loss landscape of the 10D environment. Here, we show two more slices of the landscape (the middle and right plots below).
In the plots, we see that the optimization landscape is similarly rugged in these dimensions.

\begin{figure}[H]
    \centering
    \vspace{-1em}
    \includegraphics[width=0.30\textwidth]{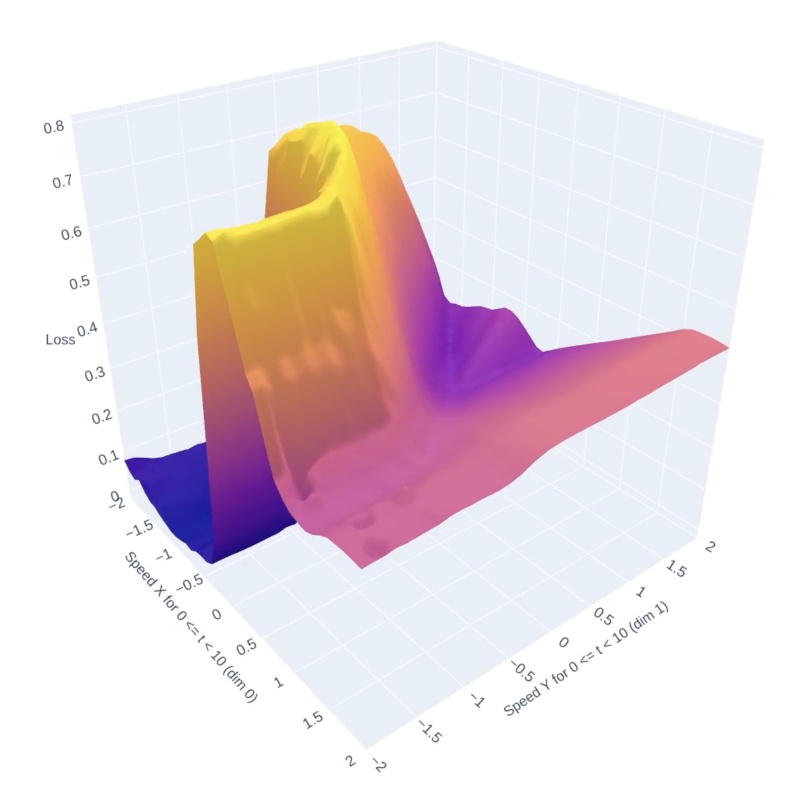}
    \includegraphics[width=0.30\textwidth]{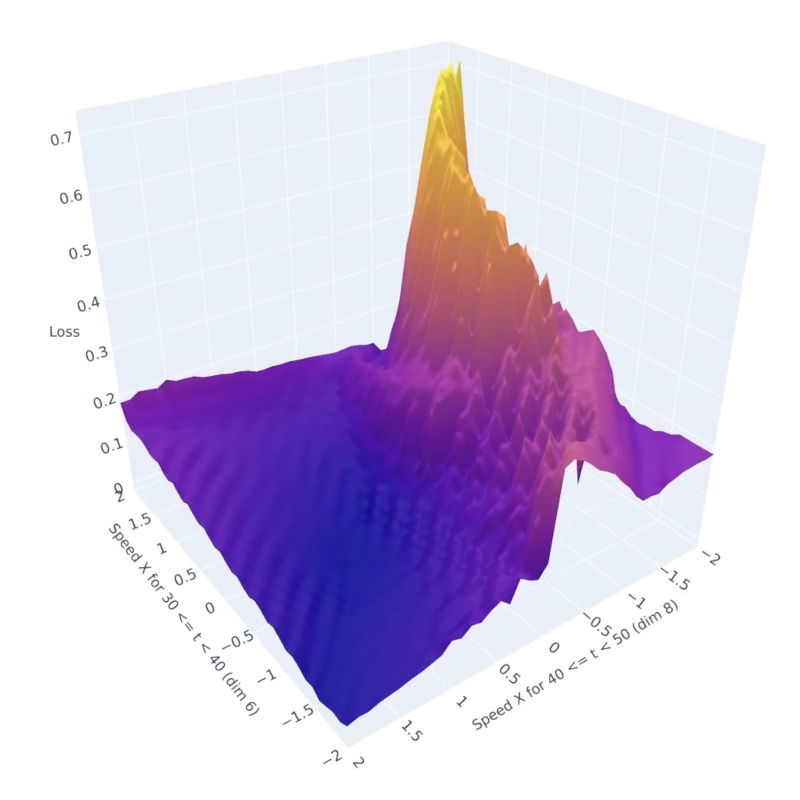}
    \includegraphics[width=0.30\textwidth]{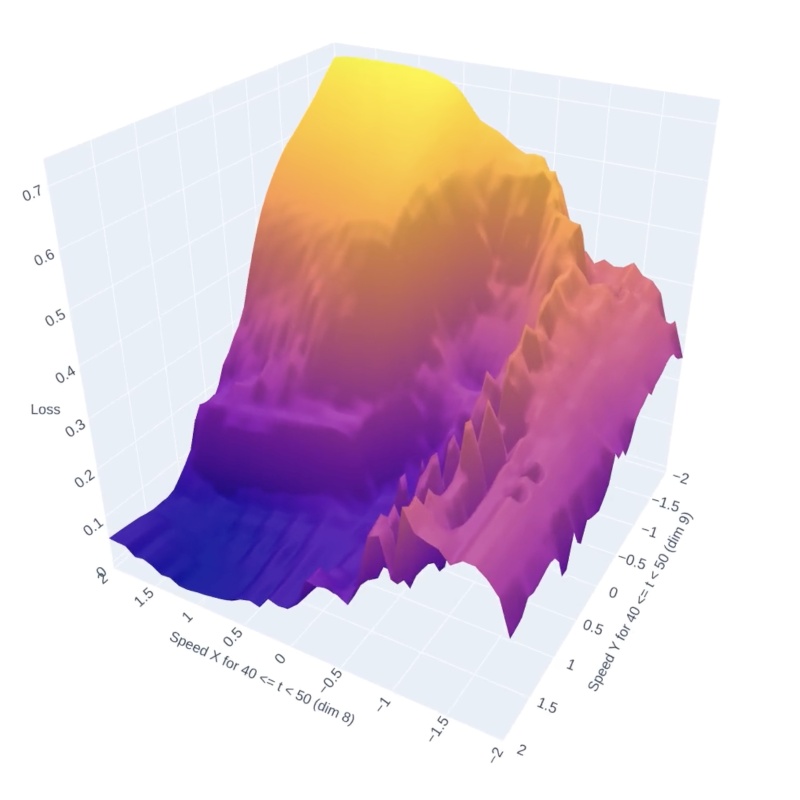}
    \includegraphics[width=0.30\textwidth]{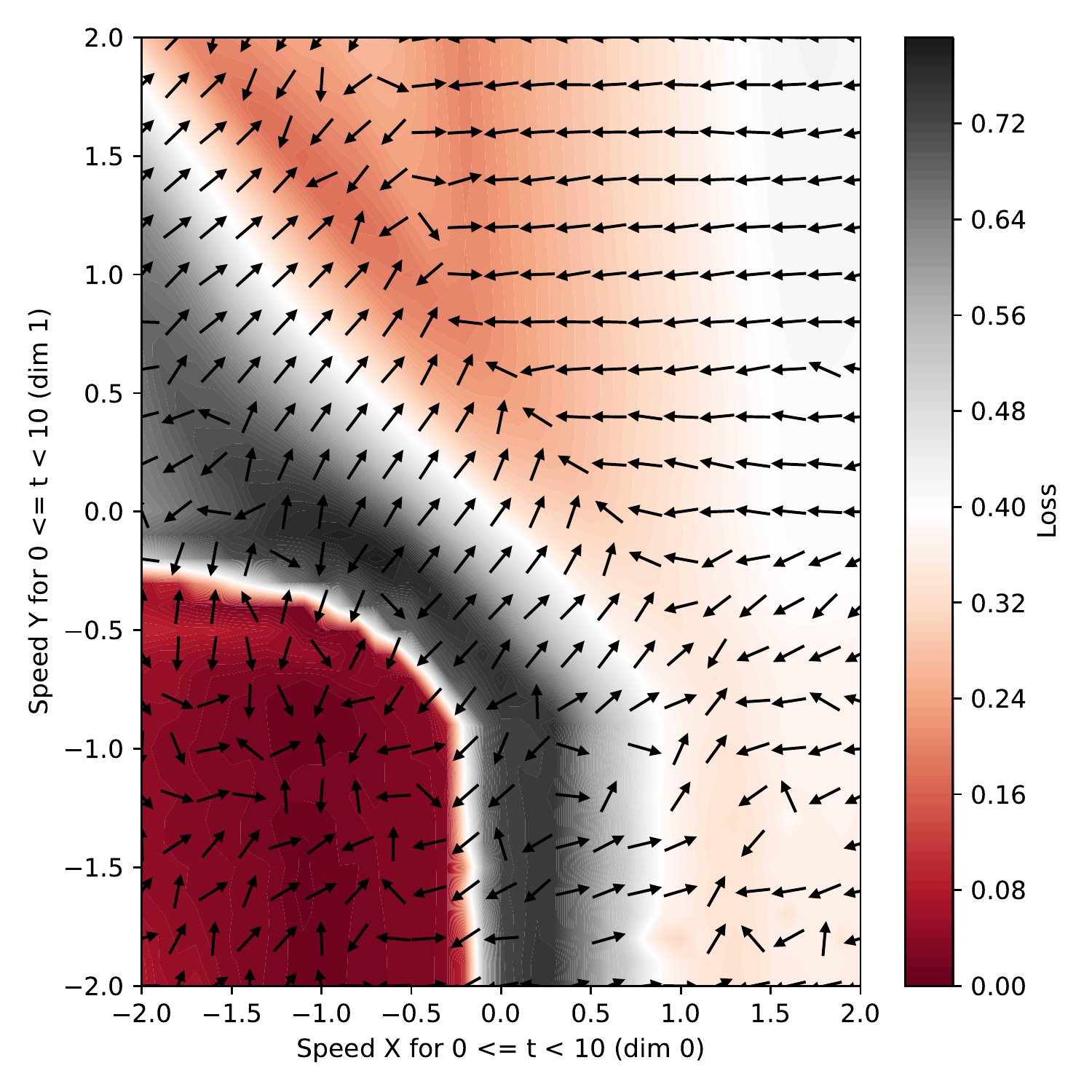}
    \includegraphics[width=0.30\textwidth]{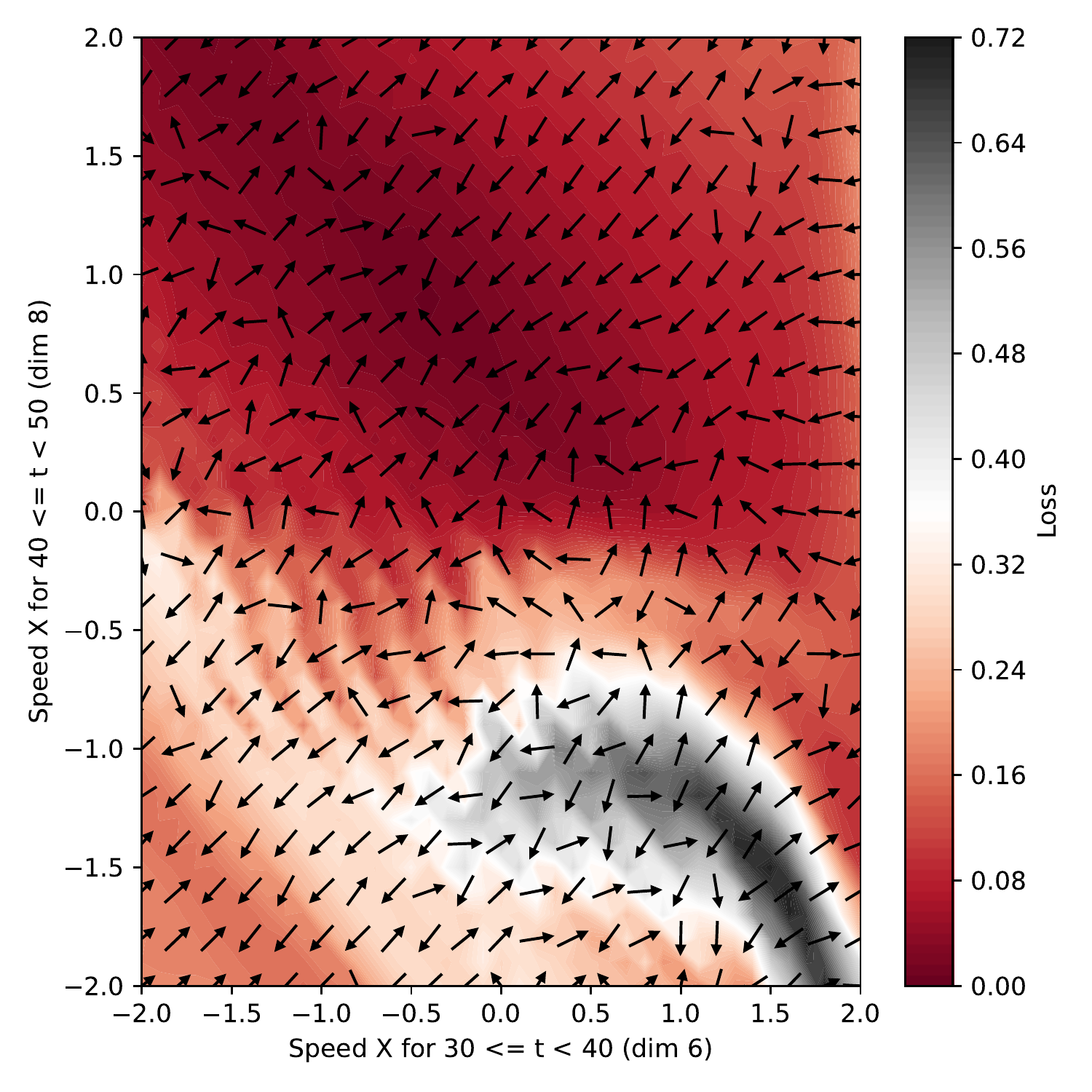}
    \includegraphics[width=0.30\textwidth]{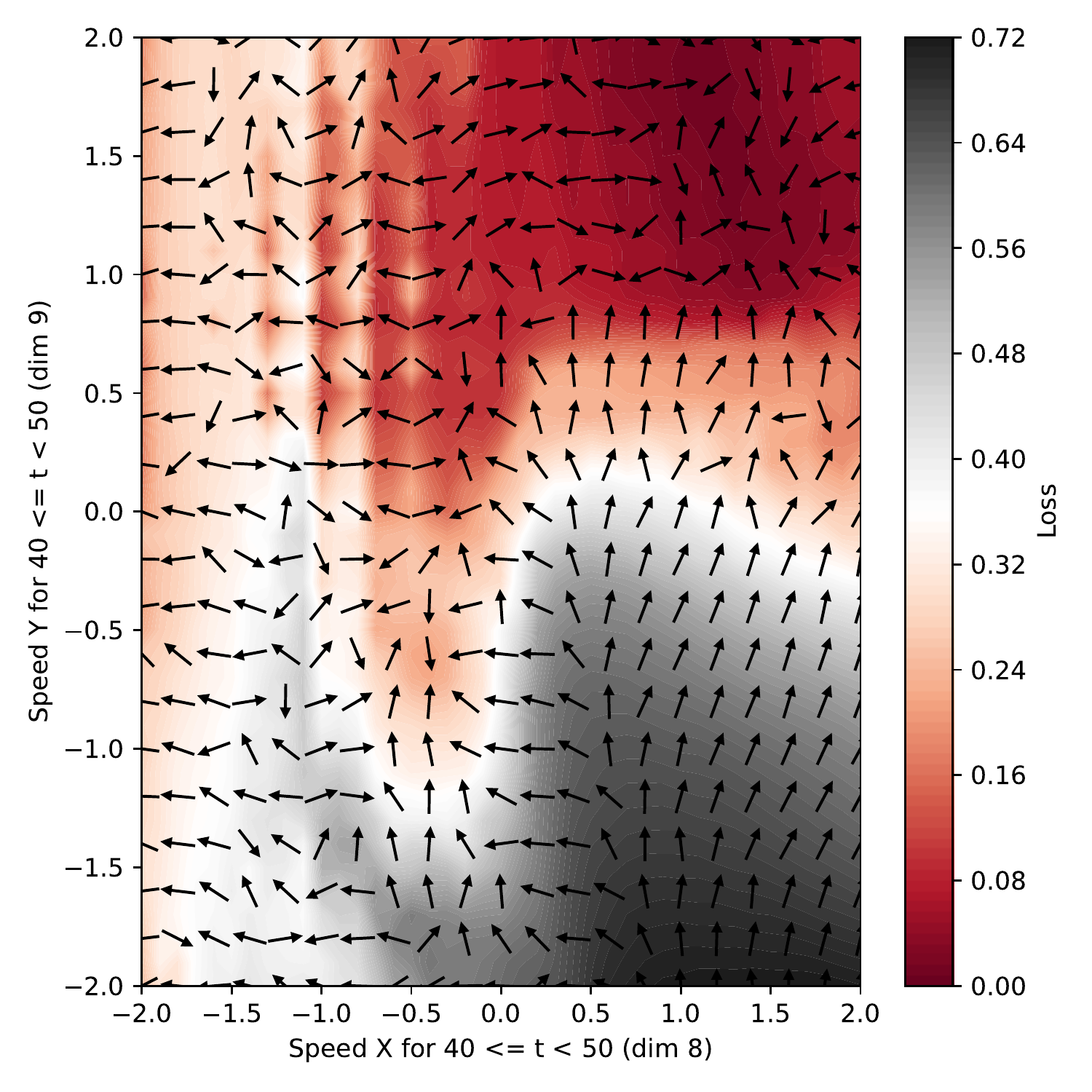}
    \caption{Landscape (top) and gradient (bottom) plots for Fluid environment with 10D parameters. \textit{Left column} -- dimensions 0 and 1. \textit{Middle column} -- dimensions 6 and 8. \textit{Right column} -- dimensions 8 and 9.}
\end{figure}

\textbf{Assembly}\quad
\textit{Assembly} is an environment in \textit{PlasticineLab} where two anchors need to pick up a soft purple ball on the left side of the scene and place it on a yellow stand on the right side. The landscape and gradients plotted below show that these environments have smooth landscapes with local minima and good quality gradients.

\begin{figure}[H]
    \centering
    \includegraphics[width=0.30\textwidth]{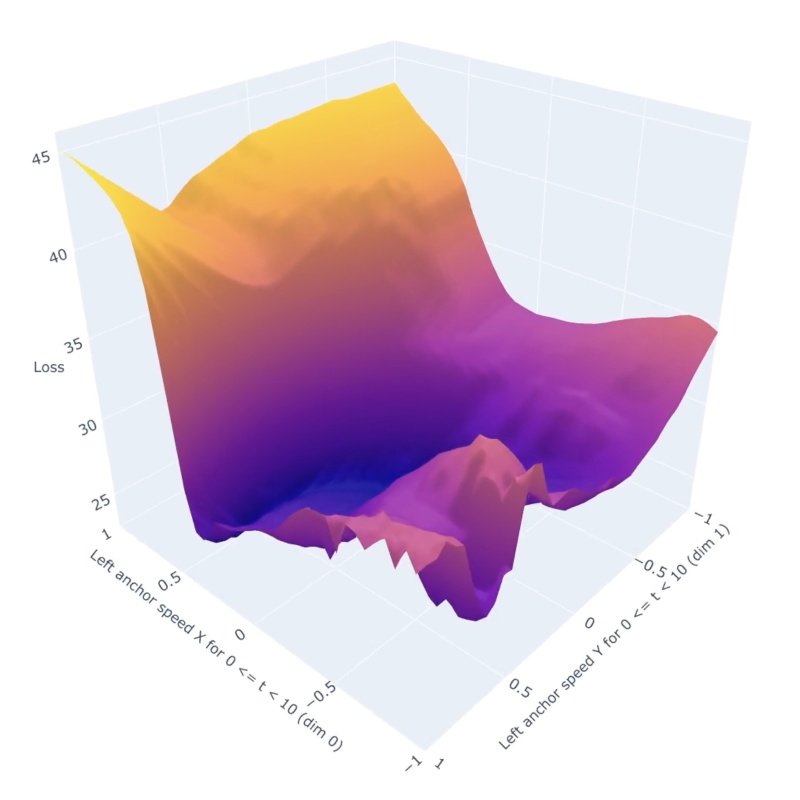}
    \includegraphics[width=0.30\textwidth]{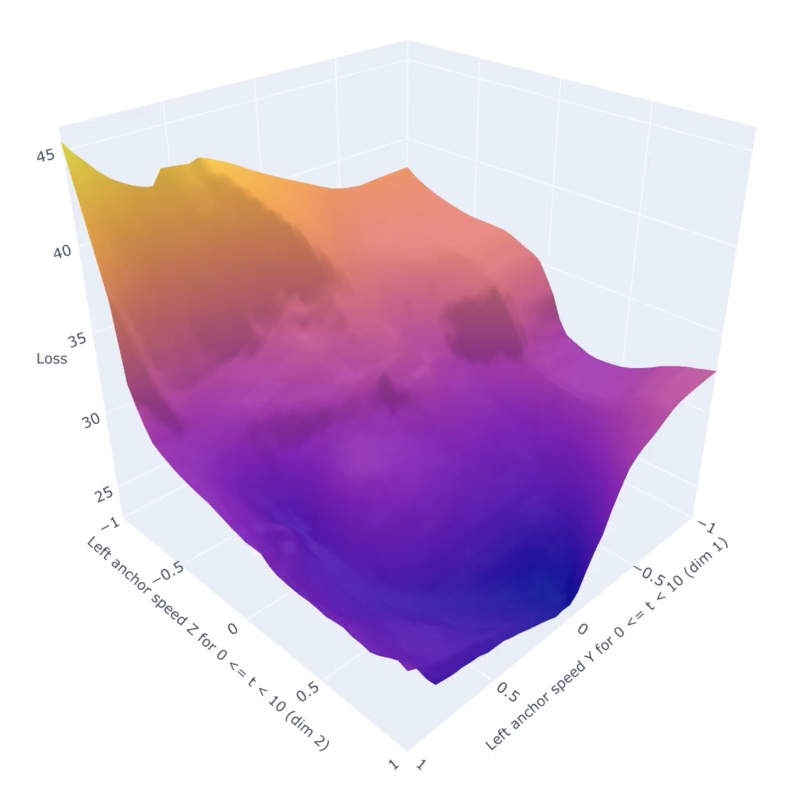}
    \includegraphics[width=0.30\textwidth]{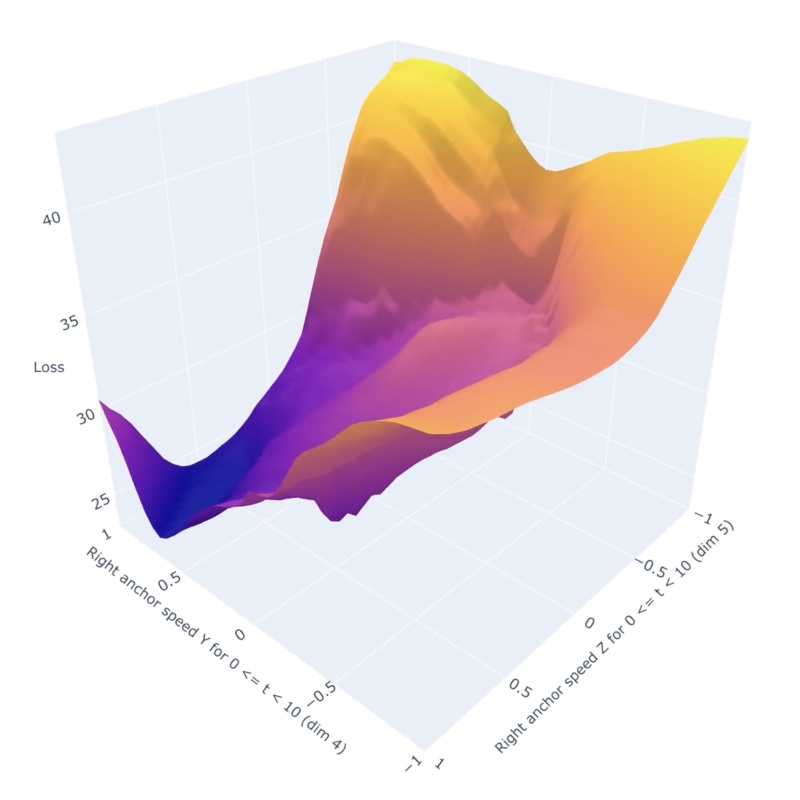}
    \includegraphics[width=0.30\textwidth]{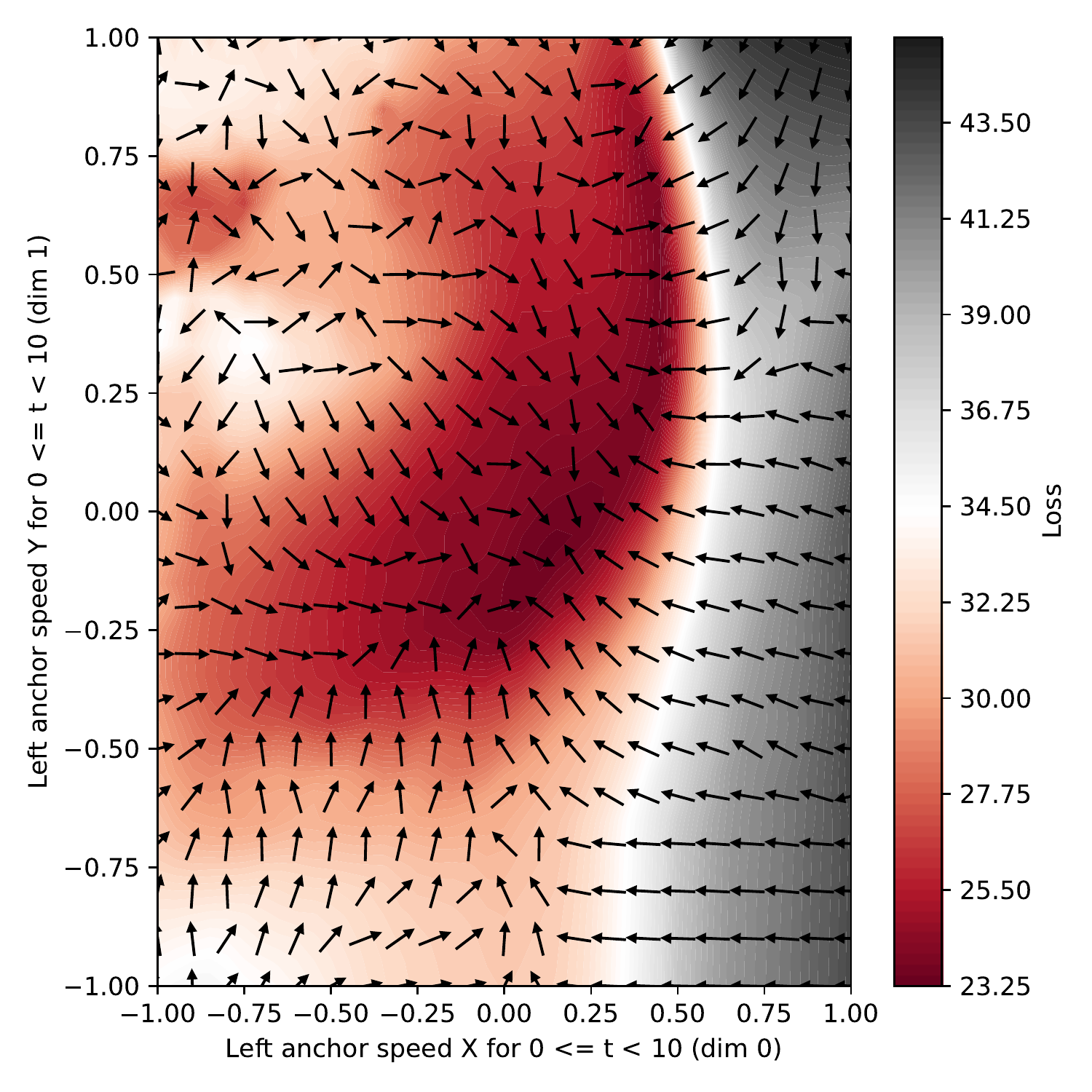}
    \includegraphics[width=0.30\textwidth]{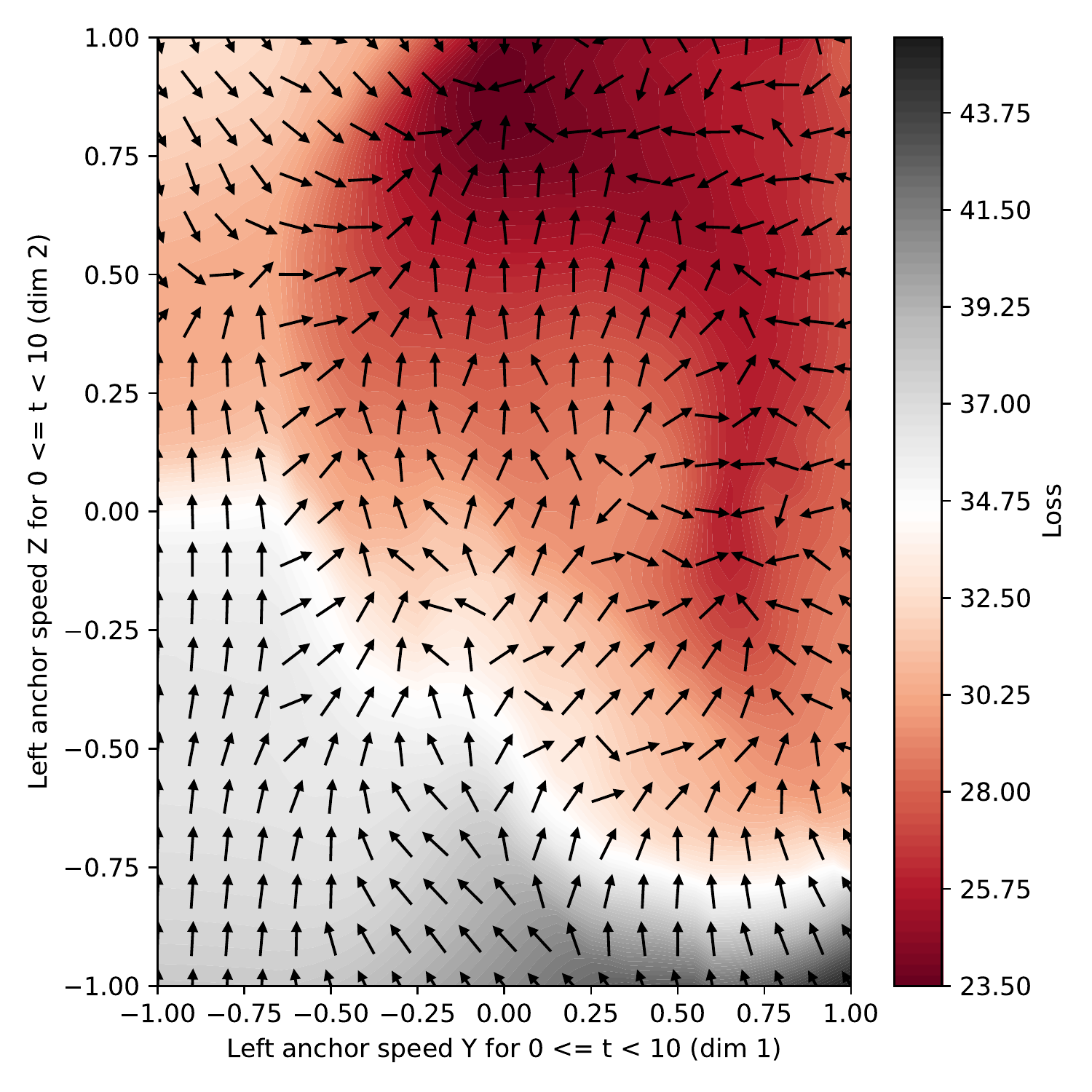}
    \includegraphics[width=0.30\textwidth]{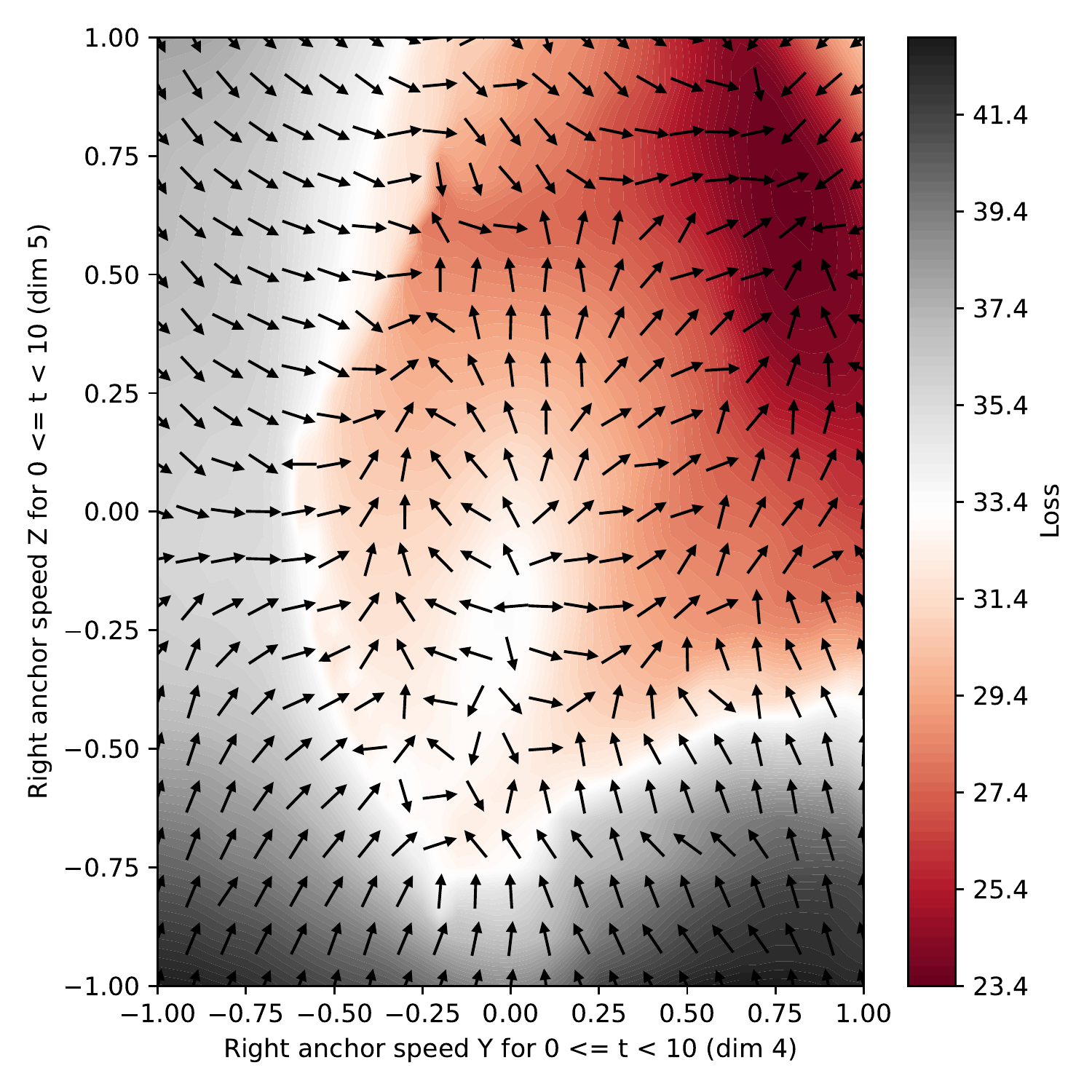}
    \caption{Landscape (top) and gradient (bottom) plots for Assembly environment with 30D parameters. \textit{Left column} -- dimensions 0 and 1. \textit{Middle column} -- dimensions 1 and 2. \textit{Right column} -- dimensions 4 and 5.}
\end{figure}

\textbf{Table}\quad
In the \textit{Table} environment, an anchor pushes one leg of a table so it points outward. The visualizations below reveal that small changes in the action can result in very different loss values.

\begin{figure}[H]
    \centering
    \includegraphics[width=0.30\textwidth]{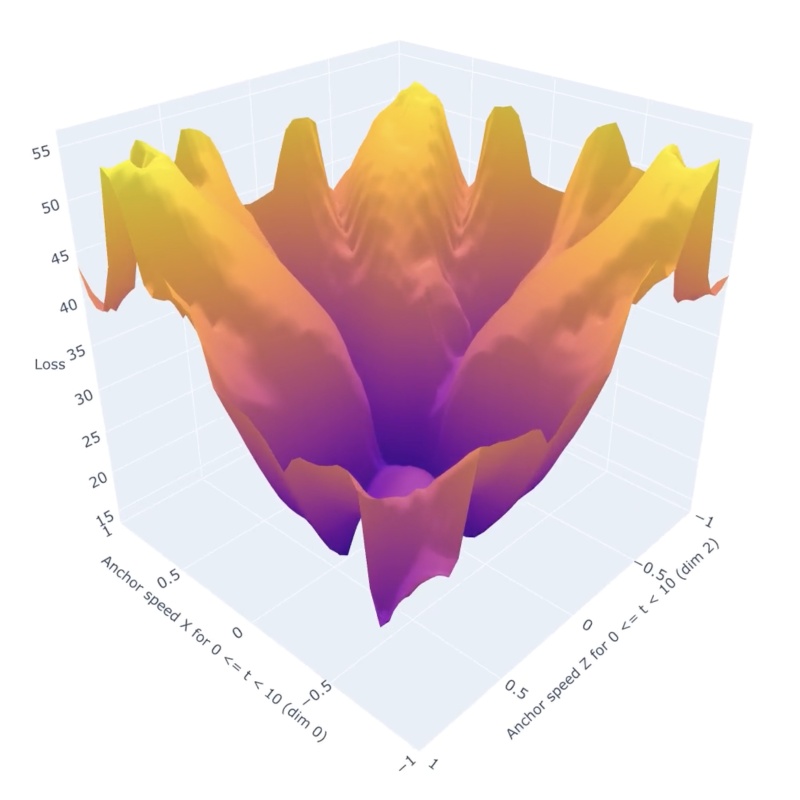}
    \includegraphics[width=0.30\textwidth]{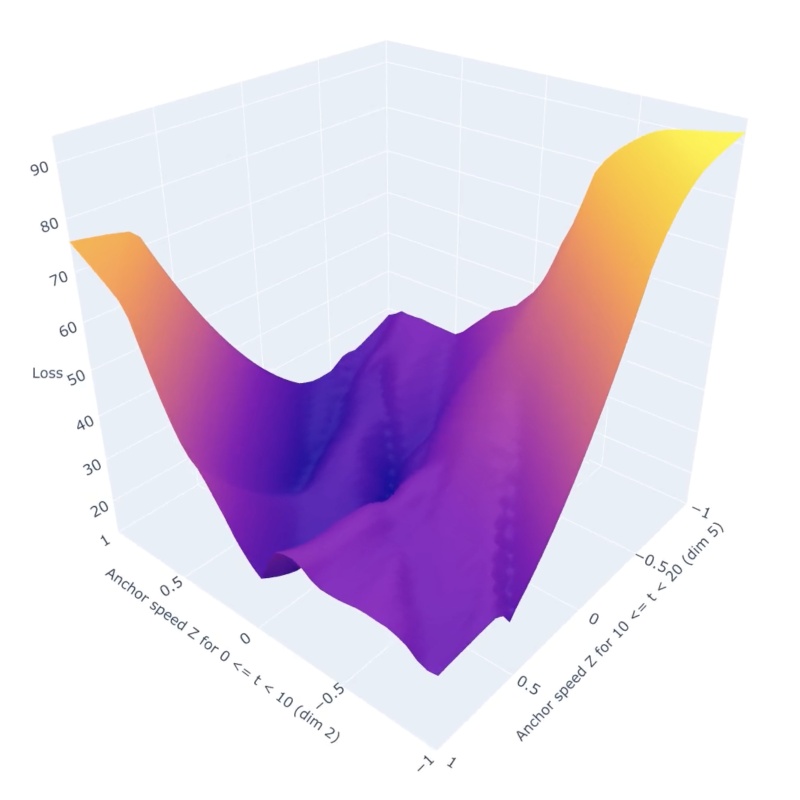}
    \includegraphics[width=0.30\textwidth]{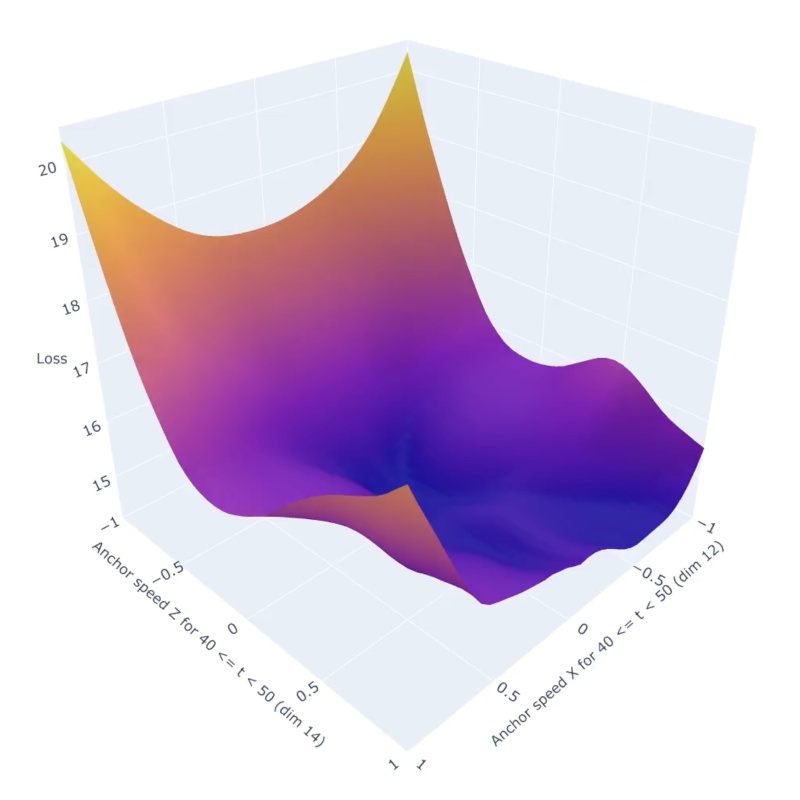}
    \includegraphics[width=0.30\textwidth]{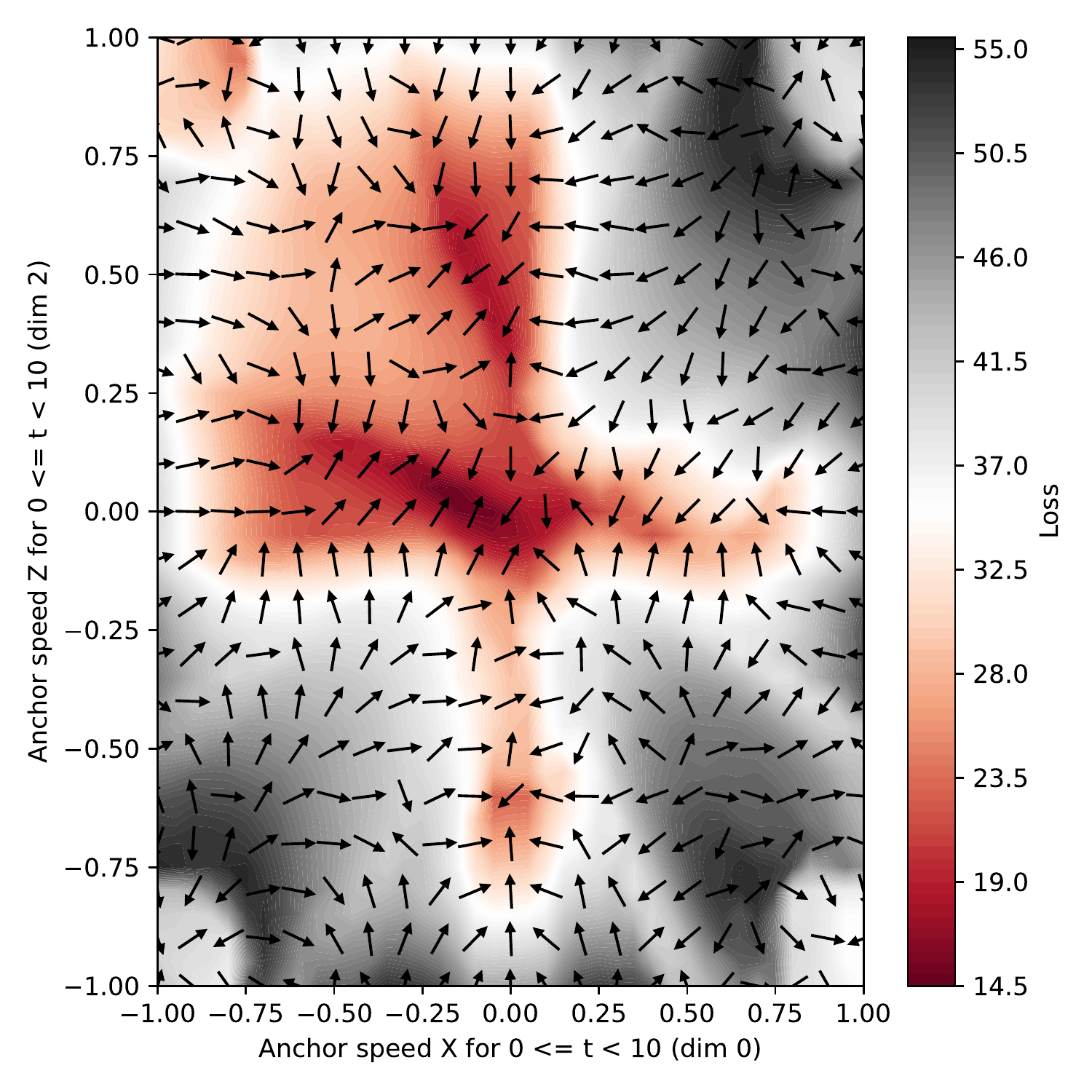}
    \includegraphics[width=0.30\textwidth]{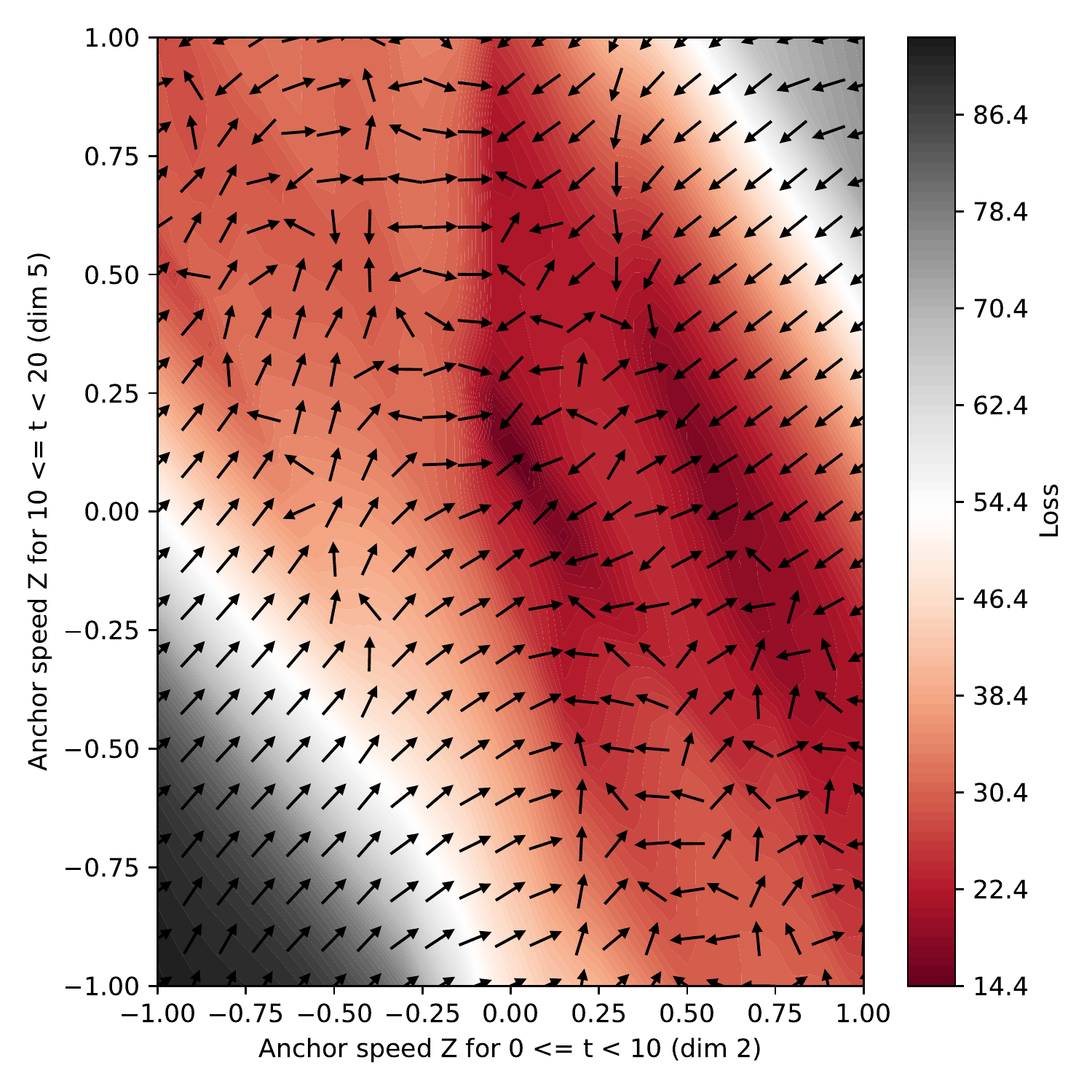}
    \includegraphics[width=0.30\textwidth]{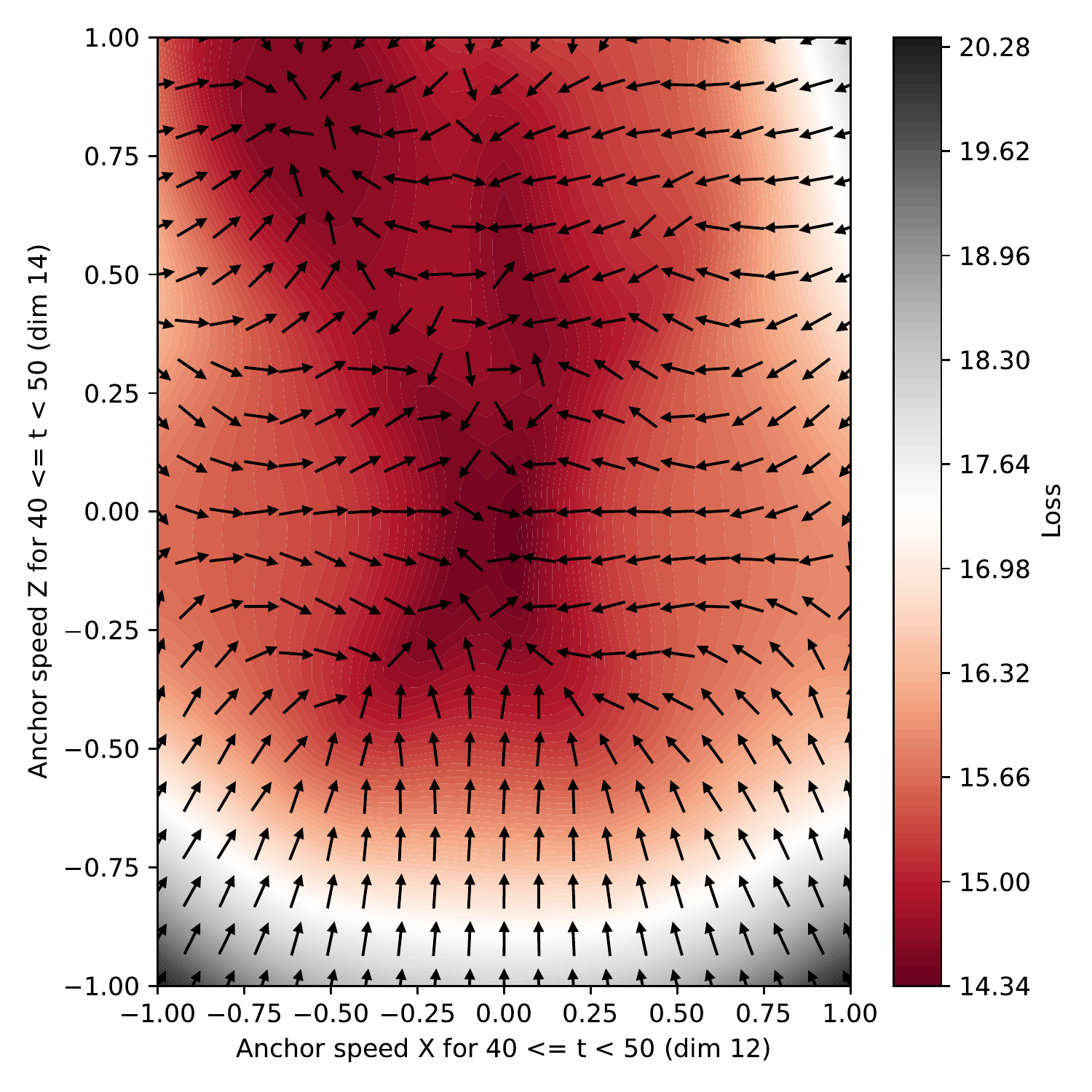}
    \caption{Landscape (top) and gradient (bottom) plots for Table environment with 30D parameters. \textit{Left column} -- dimensions 0 and 2. \textit{Middle column} -- dimensions 2 and 5. \textit{Right column} -- dimensions 12 and 13.}
\end{figure}

\textbf{TripleMove}\quad
In the \textit{TripleMove} environment, six anchors manipulate three blocks to move them to three corresponding goal positions. It can be seen in the plots below that the landscape generated by this environment is pretty rugged.

\begin{figure}[H]
    \centering
    \includegraphics[width=0.30\textwidth]{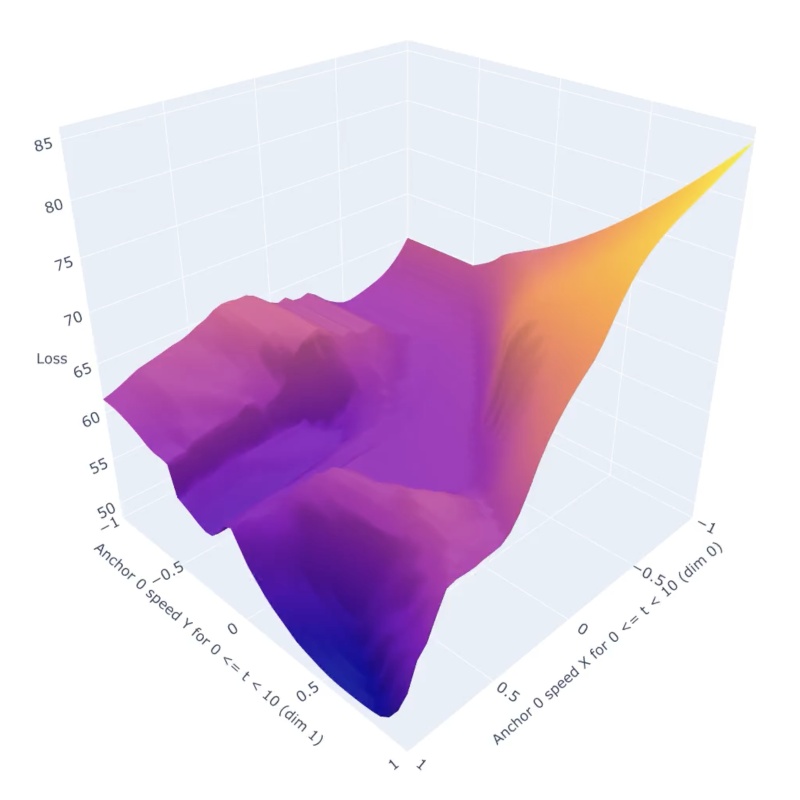}
    \includegraphics[width=0.30\textwidth]{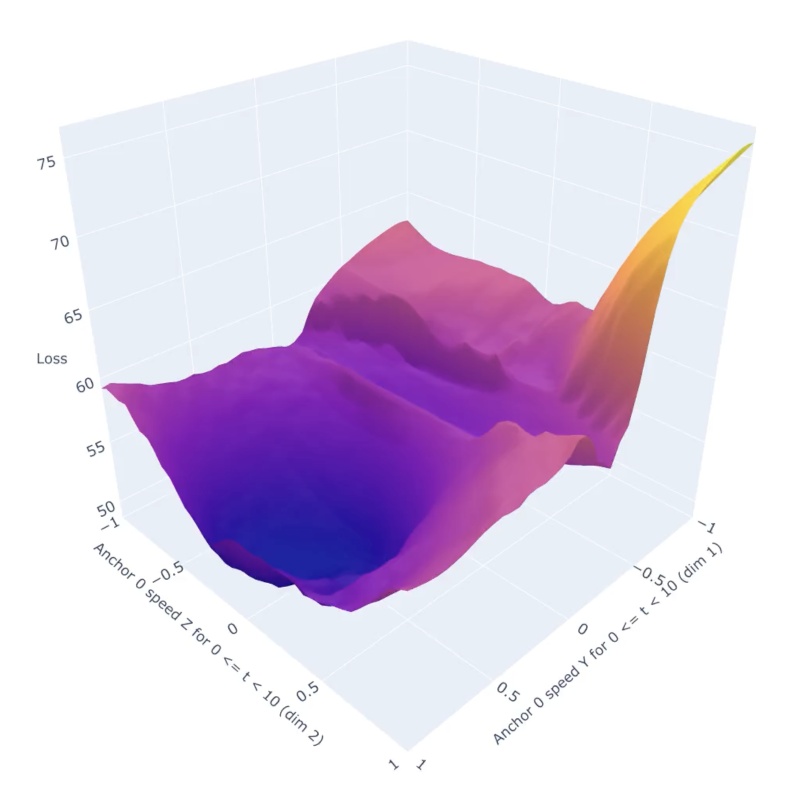}
    \includegraphics[width=0.30\textwidth]{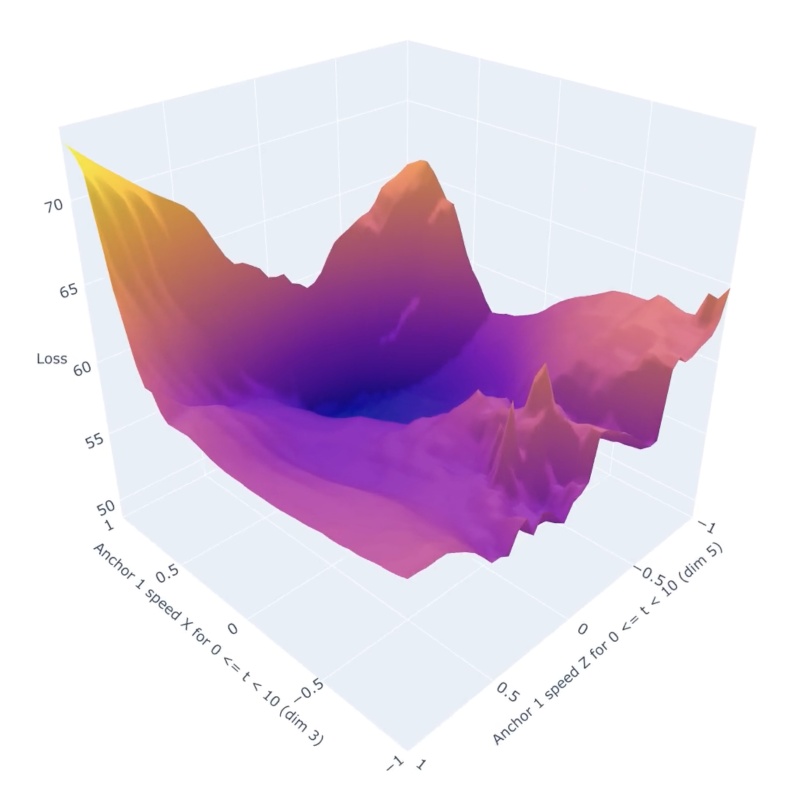}
    \includegraphics[width=0.30\textwidth]{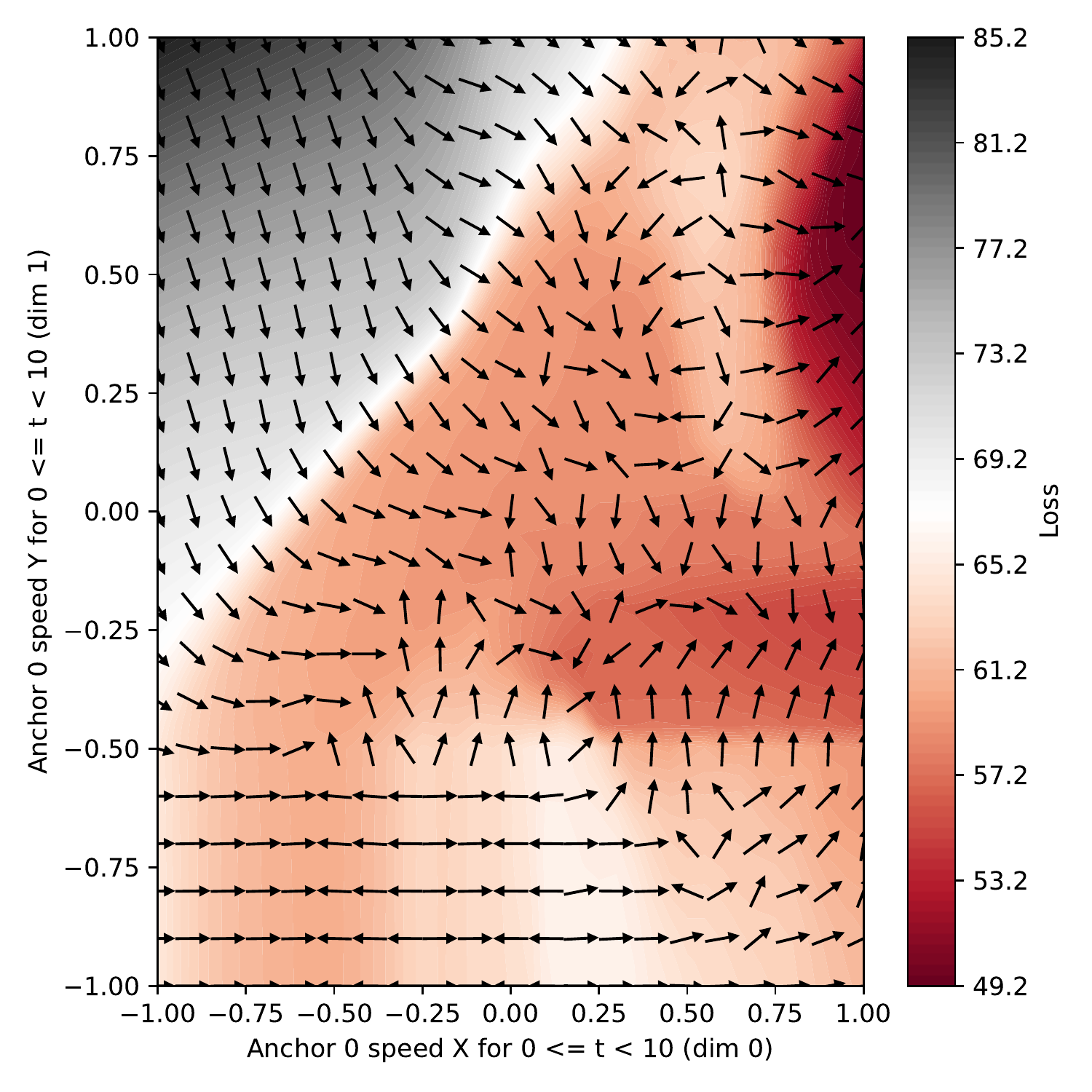}
    \includegraphics[width=0.30\textwidth]{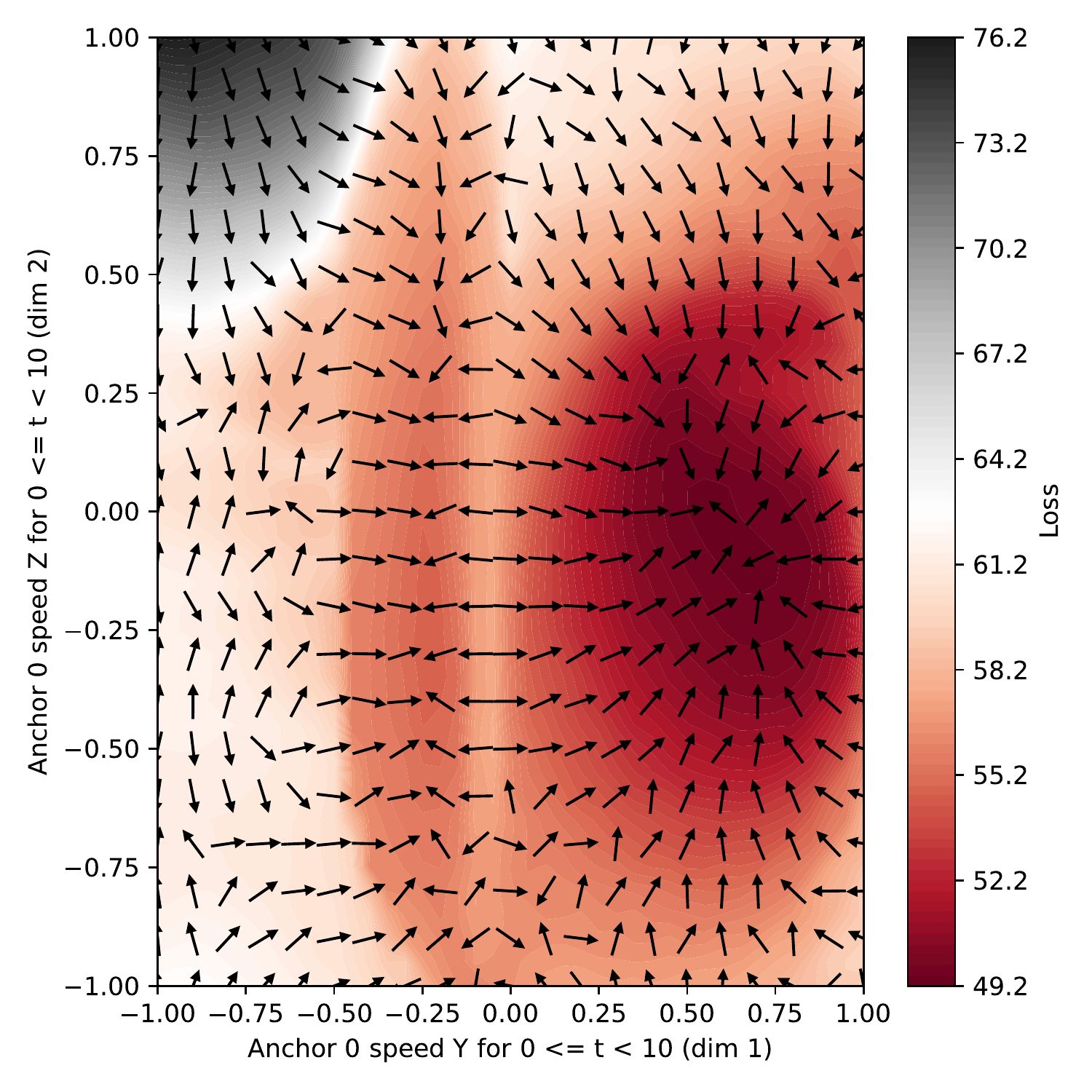}
    \includegraphics[width=0.30\textwidth]{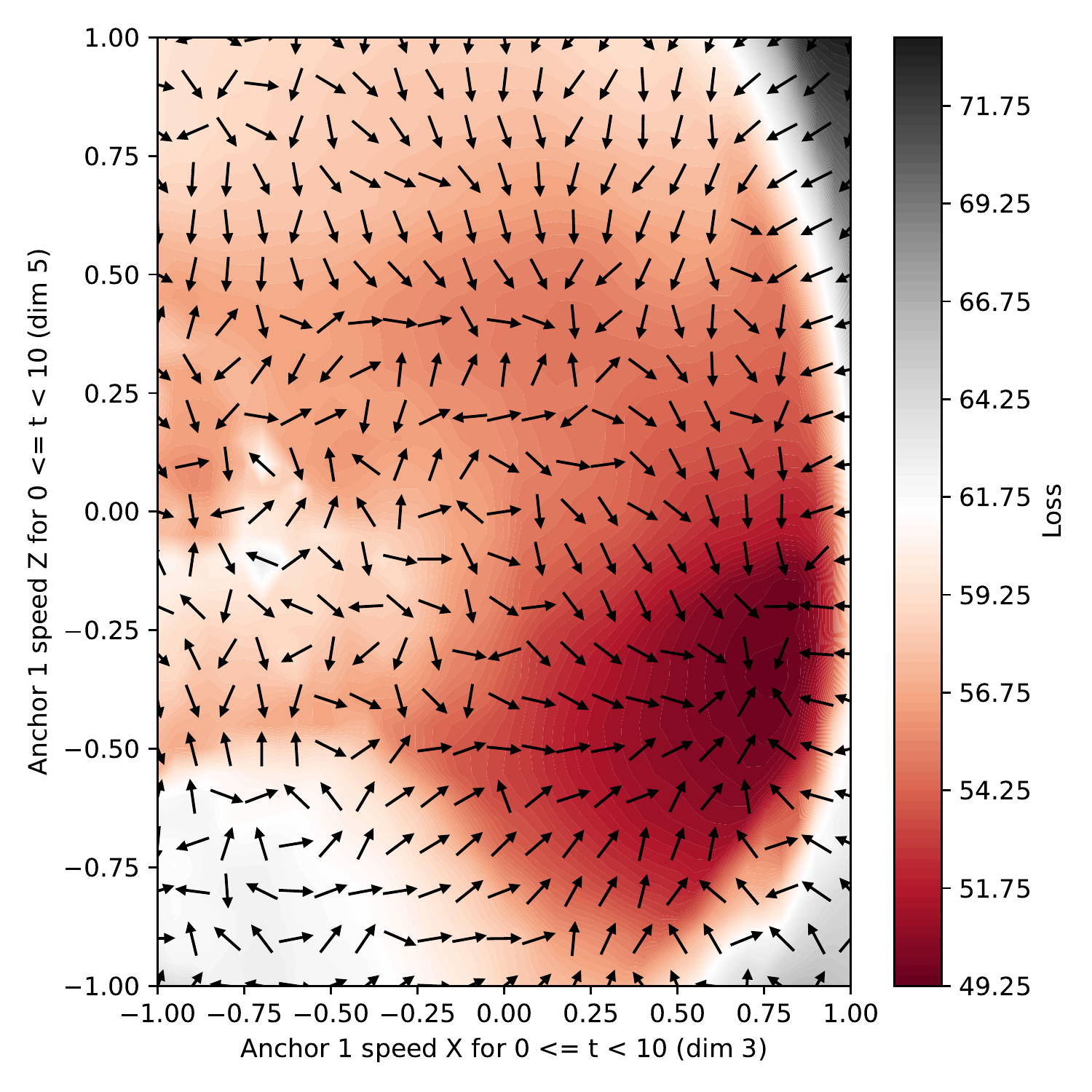}
    \caption{Landscape (top) and gradient (bottom) plots for TripleMove environment with 90D parameters. \textit{Left column} -- dimensions 0 and 1. \textit{Middle column} -- dimensions 1 and 2. \textit{Right column} -- dimensions 3 and 5.}
\end{figure}

\textbf{Writer}\quad
Below we show landscape and gradient plots of the \textit{Writer} environment. We observe that in this environment, the landscape is smoother than that of the previous environments, but there are still a number of local minima at different areas of the landscape (see the left plot and the right plot).

\begin{figure}[H]
    \centering
    \includegraphics[width=0.30\textwidth]{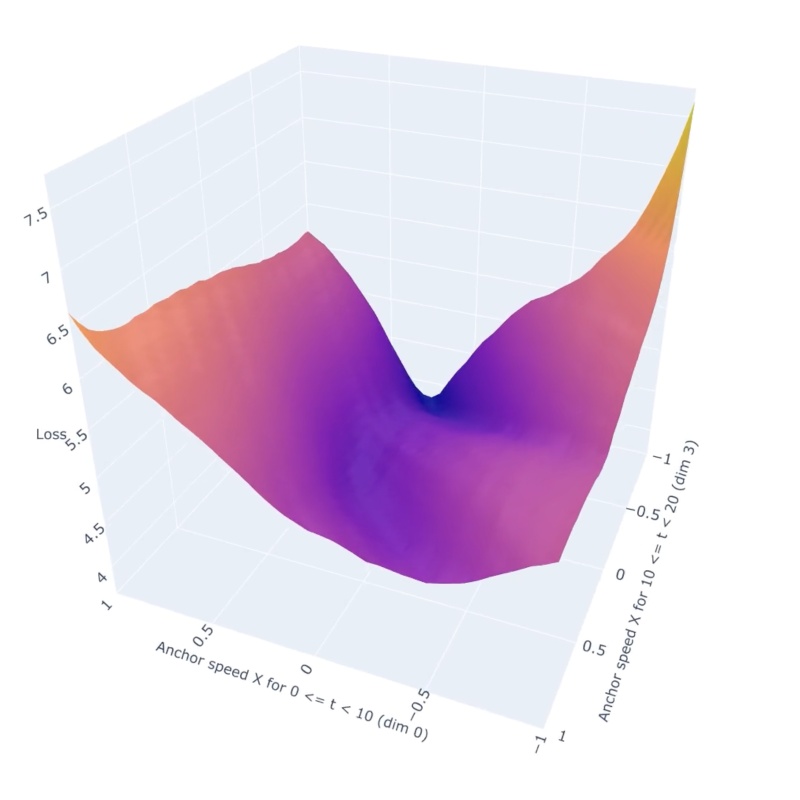}
    \includegraphics[width=0.30\textwidth]{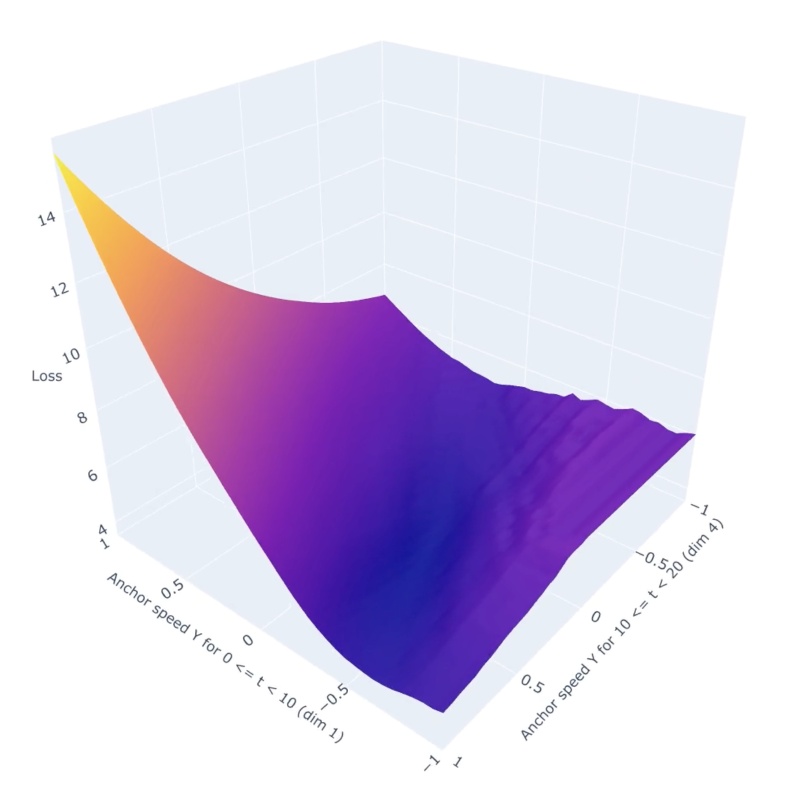}
    \includegraphics[width=0.30\textwidth]{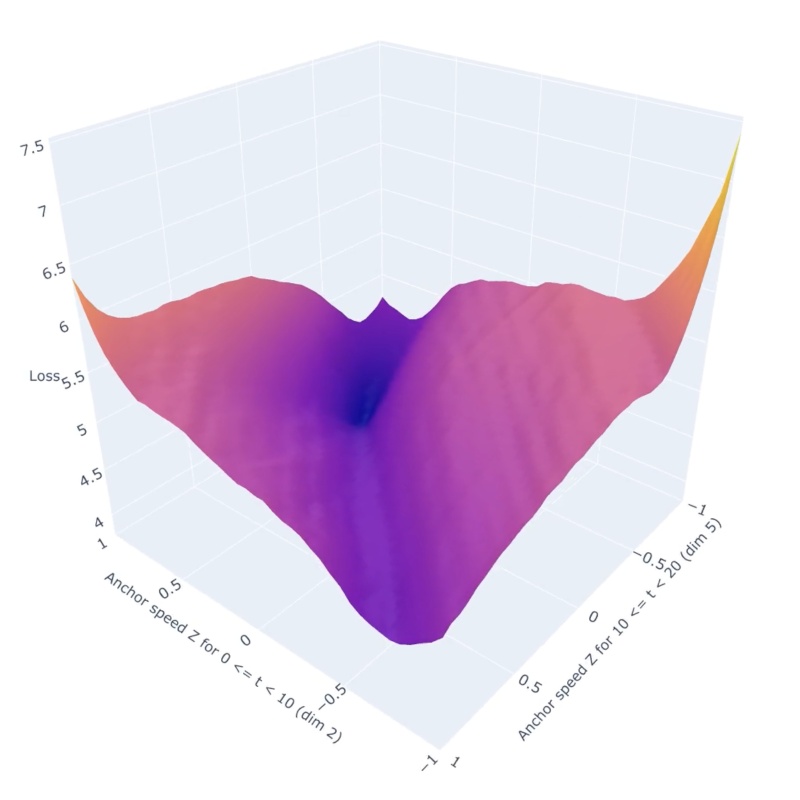}
    \includegraphics[width=0.30\textwidth]{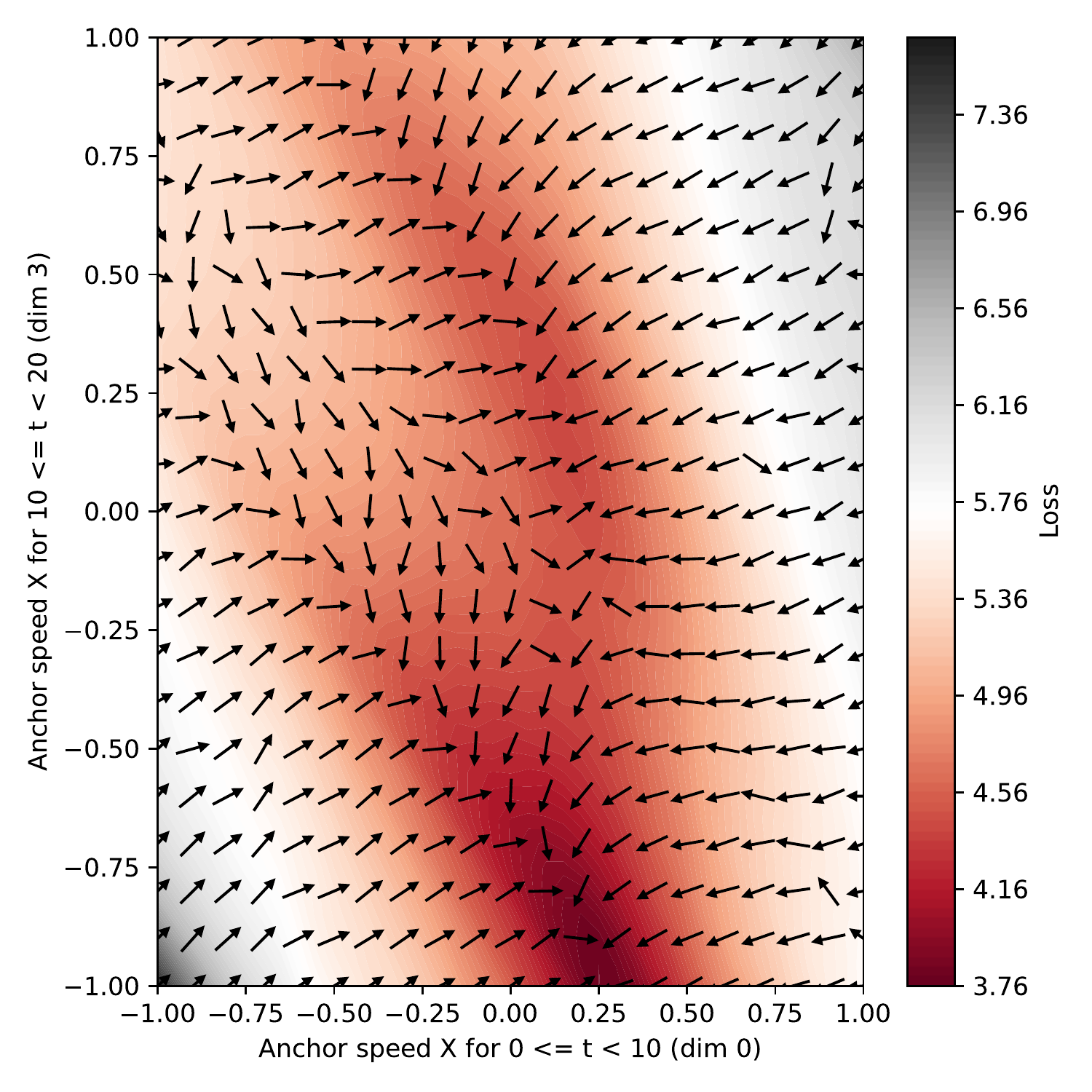}
    \includegraphics[width=0.30\textwidth]{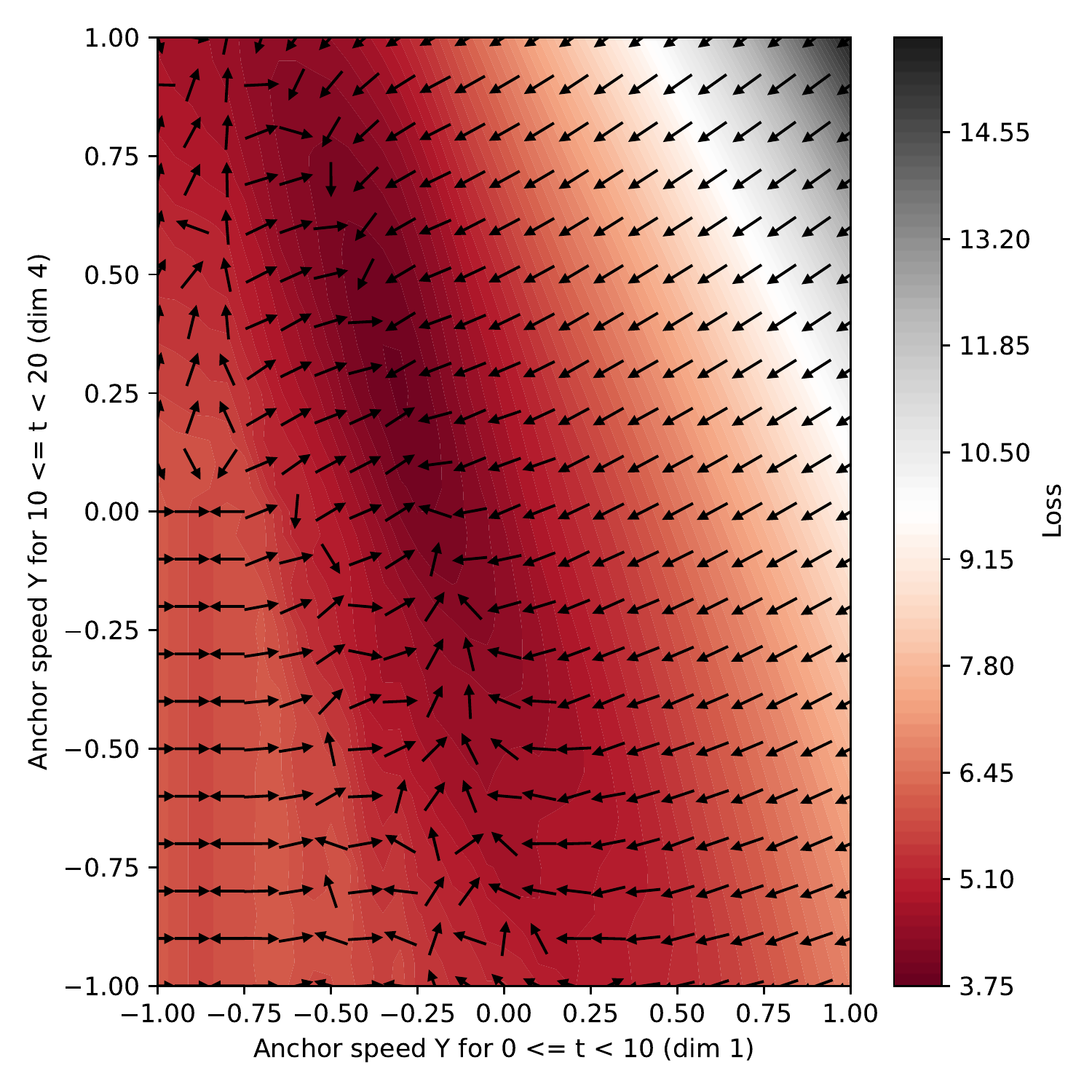}
    \includegraphics[width=0.30\textwidth]{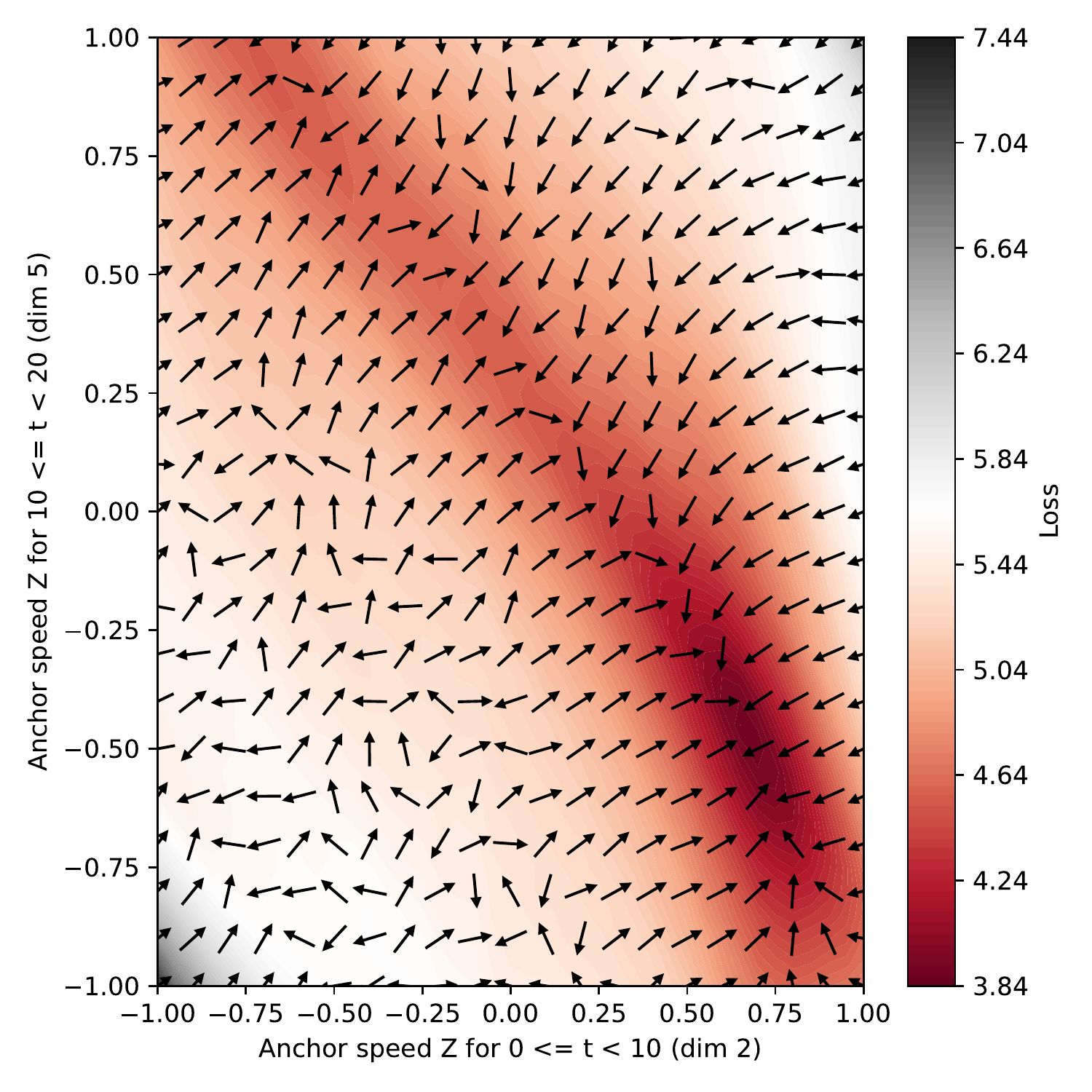}
    \caption{Landscape (top) and gradient (bottom) plots for Writer environment with 15D parameters. \textit{Left column} -- dimensions 0 and 3. \textit{Middle column} -- dimensions 1 and 4. \textit{Right column} -- dimensions 2 and 5.}
\end{figure}

\textbf{Flip}\quad
In the paper, we showed that the \textit{Flip} environment has rugged landscapes with suboptimal gradient quality. Below we show additional landscape and gradient plots of the \textit{Flip} environment to confirm this. As seen in the plots, the landscape is very rugged, and gradients are pointing to rather random directions.

\begin{figure}[H]
    \centering
    \includegraphics[width=0.30\textwidth]{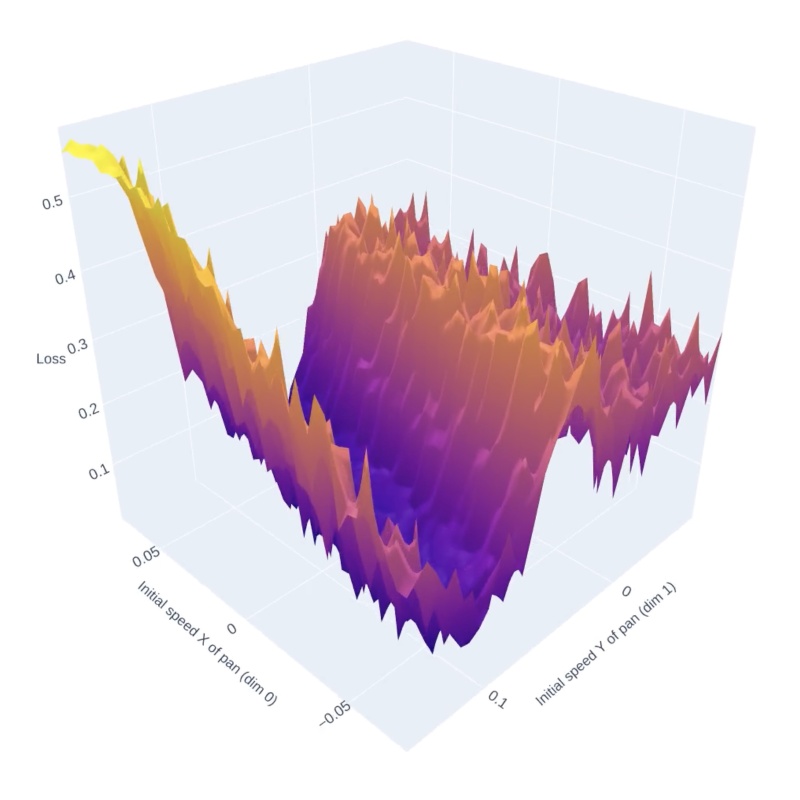}
    \includegraphics[width=0.30\textwidth]{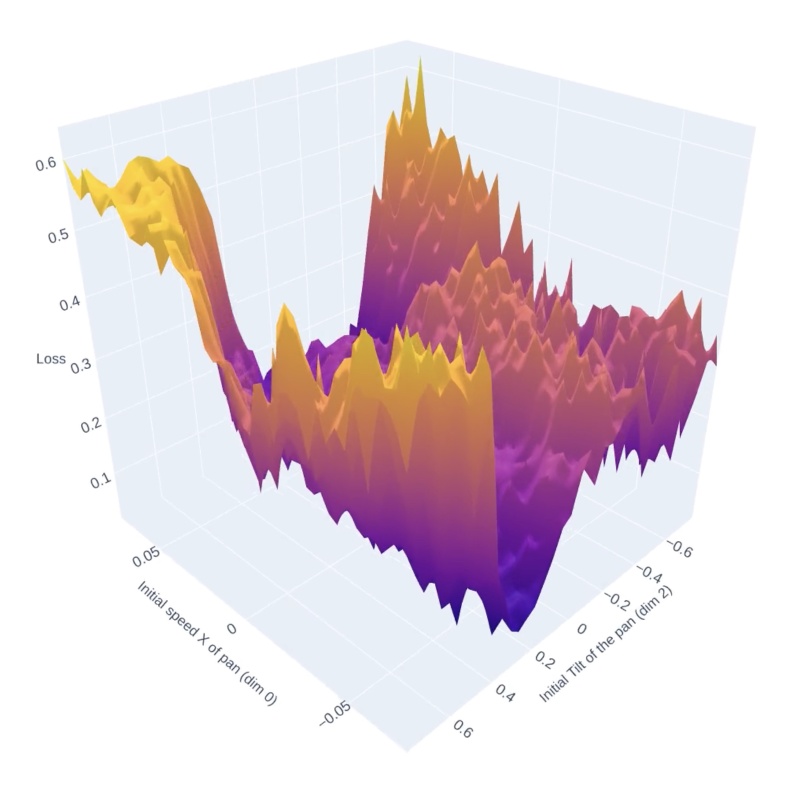}
    \includegraphics[width=0.30\textwidth]{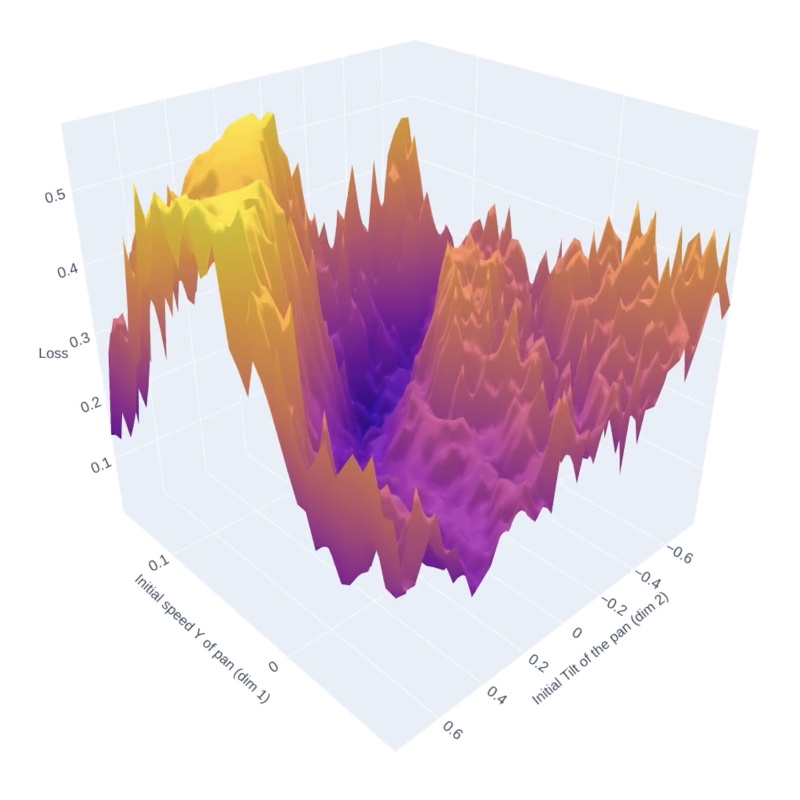}
    \includegraphics[width=0.30\textwidth]{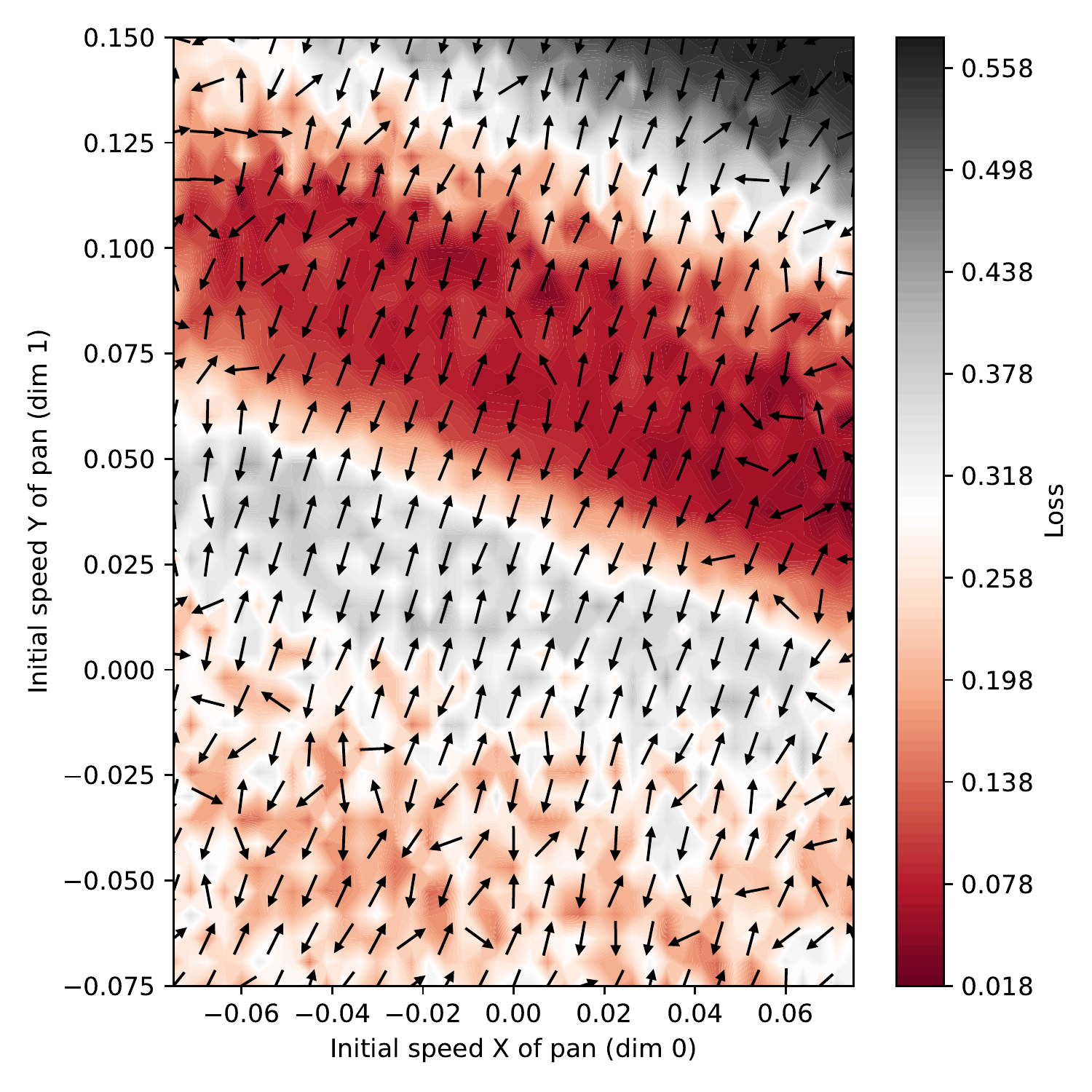}
    \includegraphics[width=0.30\textwidth]{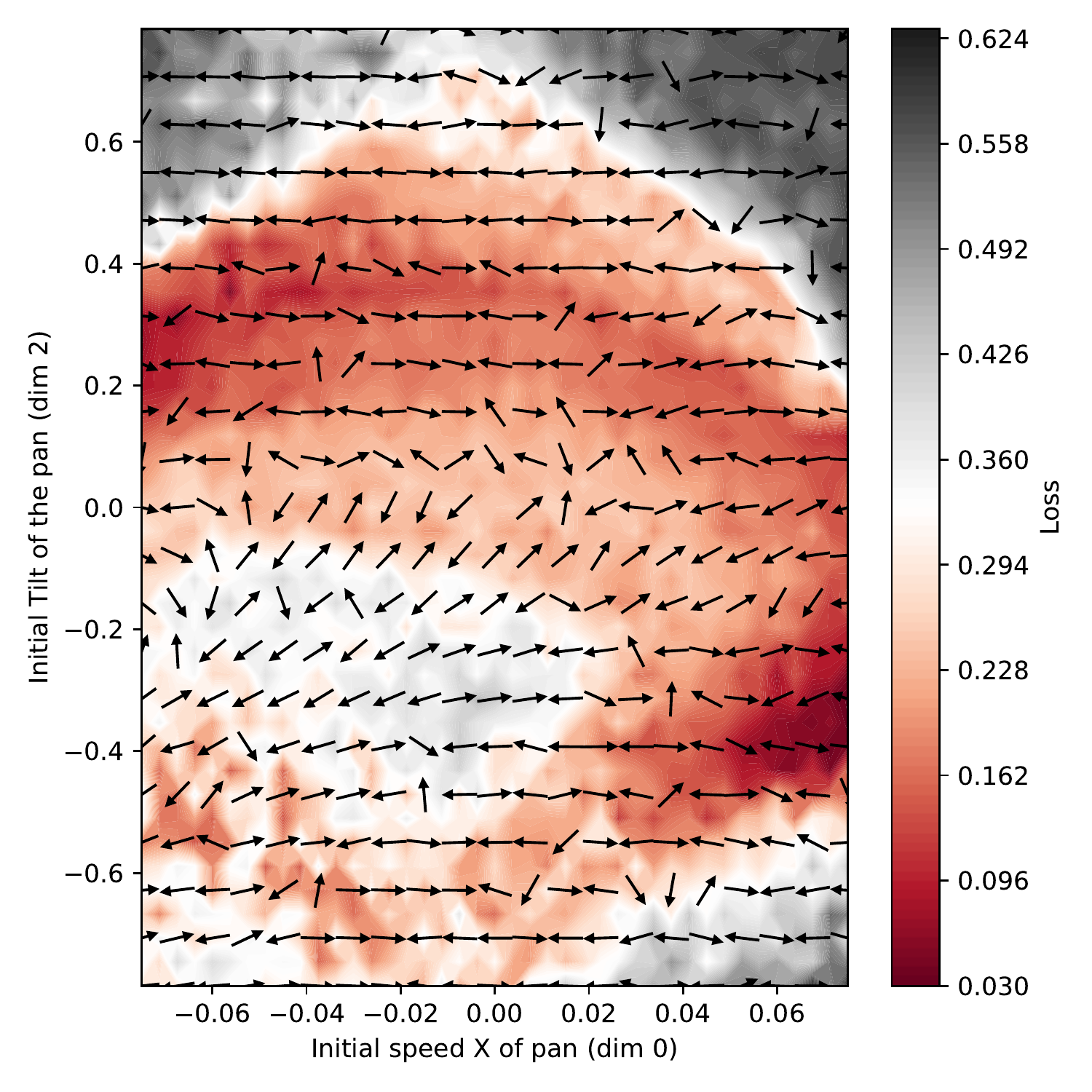}
    \includegraphics[width=0.30\textwidth]{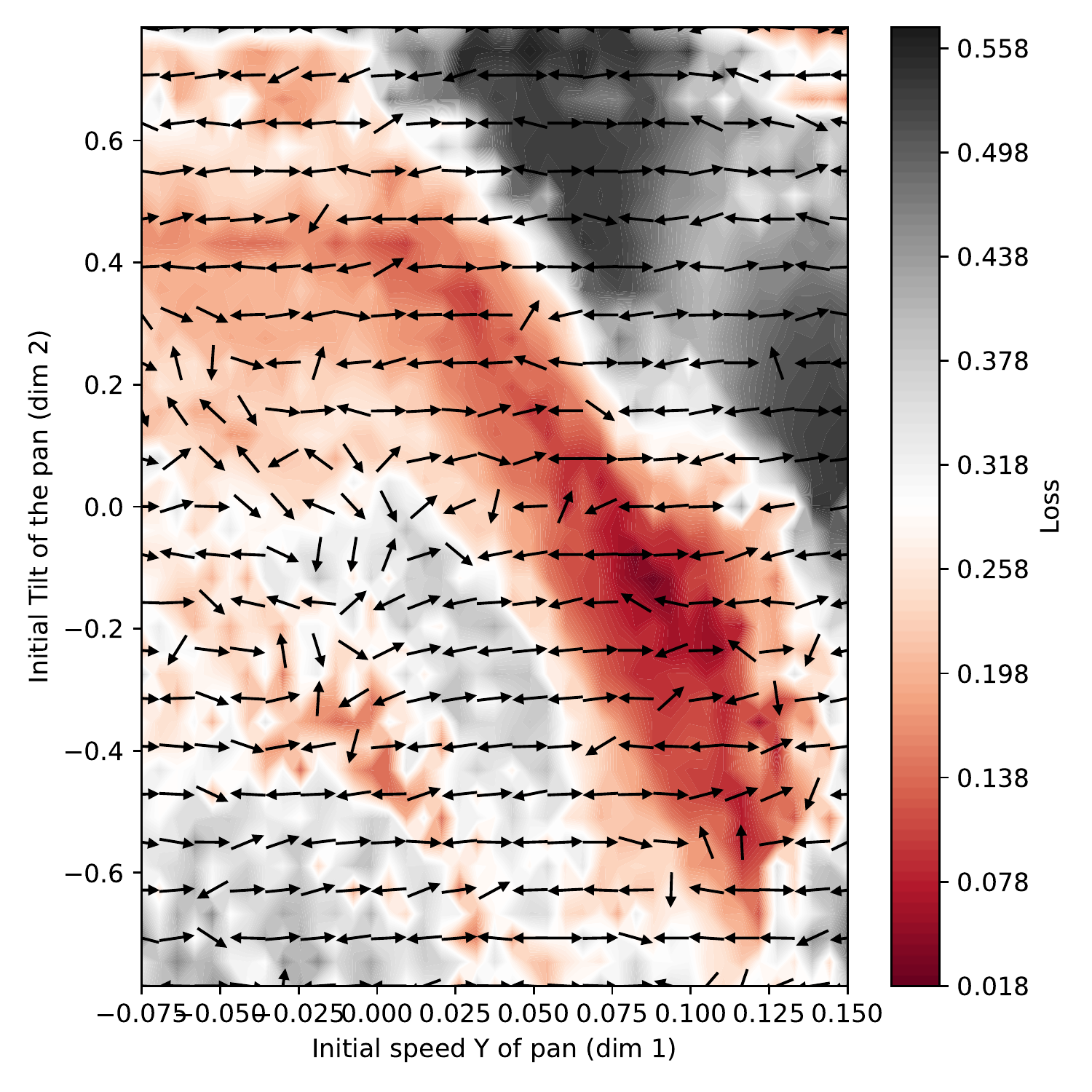}
    \caption{Landscape (top) and gradient (bottom) plots for Flip environment. \textit{Left column} -- dimensions 0 and 1. \textit{Middle column} -- dimensions 0 and 2. \textit{Right column} -- dimensions 1 and 2.}
\end{figure}
\section{Details for Method Implementation and Compute Resources}

For implementing CMA-ES, we used a fast and lightweight library provided by~\cite{codecmaes}.

For implementing Bayesian optimization (BO), we used the BOTorch library~\cite{balandat2020botorch}, which supports various versions of Gaussian processes (exact \& approximate) and various BO acquisition functions. 
For all experiments described in the main paper, we used the Lower Confidence Bound (LCB) acquisition function, with default parameters (i.e. exploration coefficient $\alpha\!=\!1.0$). In our previous experience, LCB had a similar performance as other commonly used functions (such as the Expected Improvement acquisition function), and LCB has the advantage of being very easy to implement and interpret. See Section IV in~\cite{BOtutorial2016} for more information. BOTorch implements automatic hyperparameter optimization based on maximizing the marginal likelihood (see~\cite{BOtutorial2016}, Section V-A). We used this for all our BO-based experiments.

We experimented with various versions of Gaussian process (GP) models, including exact GP and sparse variational GP versions that are provided in BOTorch. BOTorch uses the lower-level \mbox{GPyTorch} library~\cite{gardner2018gpytorch} for GP implementations. We found that exact GPs performed best, and used these for all BO experiments reported in the main paper. In future work, it would be interesting to experiment with other GP implementations that could support posteriors with a much larger number of points.

We used NVIDIA Tesla T4 GPUs and 32 cores of an Intel Xeon 2.3GHz CPU for our experiments. The computational requirements of each simulation environment differ widely. For example, our environments based on Warp, Nimble and mesh-based DiffTaichi were fastest, requiring only a few minutes for 1,000 episodes (including gradient computations). 
Particle-based DiffTaichi environments (\textit{Fluid} and the PlasticineLab environments) required significantly more time (e.g. \textit{Fluid} took $\approx\! 2$ hours for 1,000 episodes including gradient computations).

\bibliography{main} 

\end{document}